\documentclass[runningheads]{llncs}
\usepackage{graphicx}

\usepackage{tikz}
\usepackage{comment}
\usepackage{amsmath,amssymb} 
\usepackage{color}
\usepackage{framed}
\usepackage{multicol}
\usepackage[export]{adjustbox}
\usepackage[accsupp]{axessibility}  
\usepackage{latexsym}
\usepackage{multirow}
\usepackage{amssymb}
\usepackage{siunitx}
\usepackage{caption}
\usepackage{subcaption}
\usepackage{makecell}
\usepackage[shortlabels]{enumitem}
\usepackage[utf8]{inputenc} 
\usepackage[T1]{fontenc}    
\usepackage[hidelinks]{hyperref}
\usepackage{url}            
\usepackage{booktabs}       
\usepackage{amsfonts}       
\usepackage{nicefrac}       
\usepackage{microtype}      
\usepackage{comment}        
\usepackage{bm}
\usepackage{hhline}
\usepackage{mathtools}
\usepackage{xcolor}
\usepackage{arydshln}
\usepackage[font=small]{caption}
\usepackage{wrapfig}
\usepackage{algorithm}
\usepackage{algorithmic}
\usepackage{lipsum}
\usepackage{sidecap}
\usepackage{pifont}
\usepackage{fixltx2e}
\usepackage[flushleft]{threeparttable}
\usepackage{diagbox}
\usepackage{listings}

\usepackage{numprint}
\npthousandsep{,}
\pdfmapline{-dummy LMRoman12-Regular}

\definecolor{gg}{RGB}{15,150,15}
\definecolor{rr}{RGB}{230,45,45}

\usepackage{subcaption}

\usepackage[T1]{fontenc}

\usepackage[utf8]{inputenc}

\usepackage{microtype}

\usepackage{amsmath}
\usepackage{arydshln}
\usepackage{graphicx}

\newcommand{\namel}{\textsc{Physical Audiovisual CommonSense}}
\newcommand{\names}{\textsc{PACS}}

\usepackage[width=122mm,left=12mm,paperwidth=146mm,height=193mm,top=12mm,paperheight=217mm]{geometry}

\newcolumntype{C}{>{\centering}p{55pt}}
\newcolumntype{D}{>{\centering}p{60pt}}

\begin{document}
\pagestyle{headings}
\mainmatter
\def\ECCVSubNumber{5310}  

\title{\names: A Dataset for Physical Audiovisual CommonSense Reasoning}

\titlerunning{Physical Audiovisual CommonSense Reasoning}
%
\author{Samuel Yu\inst{1}\and
Peter Wu\inst{2} \and
Paul Pu Liang\inst{1} \and
Ruslan Salakhutdinov\inst{1} \and
Louis-Philippe Morency\inst{1}}
\authorrunning{Yu et al.}
%

\institute{Carnegie Mellon University \\ \email{\{samuelyu,pliang,rsalakhu,morency\}@cs.cmu.edu} \and
University of California, Berkeley \\ \email{peterw1@berkeley.edu}}

\maketitle

\begin{abstract}
In order for AI to be safely deployed in real-world scenarios such as hospitals, schools, and the workplace, it must be able to robustly reason about the physical world. Fundamental to this reasoning is \textit{physical common sense}: understanding the physical properties and affordances of available objects, how they can be manipulated, and how they interact with other objects. Physical commonsense reasoning is fundamentally a multi-sensory task, since physical properties are manifested through multiple modalities - two of them being vision and acoustics. Our paper takes a step towards real-world physical commonsense reasoning by contributing \names: the first audiovisual benchmark annotated for physical commonsense attributes. \names\ contains \numprint{13400} question-answer pairs, involving \numprint{1377} unique physical commonsense questions and \numprint{1526} videos.
Our dataset provides new opportunities to advance the research field of physical reasoning by bringing audio as a core component of this multimodal problem.
Using \names, we evaluate multiple state-of-the-art models on our new challenging task. While some models show promising results ($70\%$ accuracy), they all fall short of human performance ($95\%$ accuracy). We conclude the paper by demonstrating the importance of multimodal reasoning and providing possible avenues for future research.


\end{abstract}

\section{Introduction}

\begin{figure}[t]
  \centering
  \includegraphics[width=0.88\textwidth]{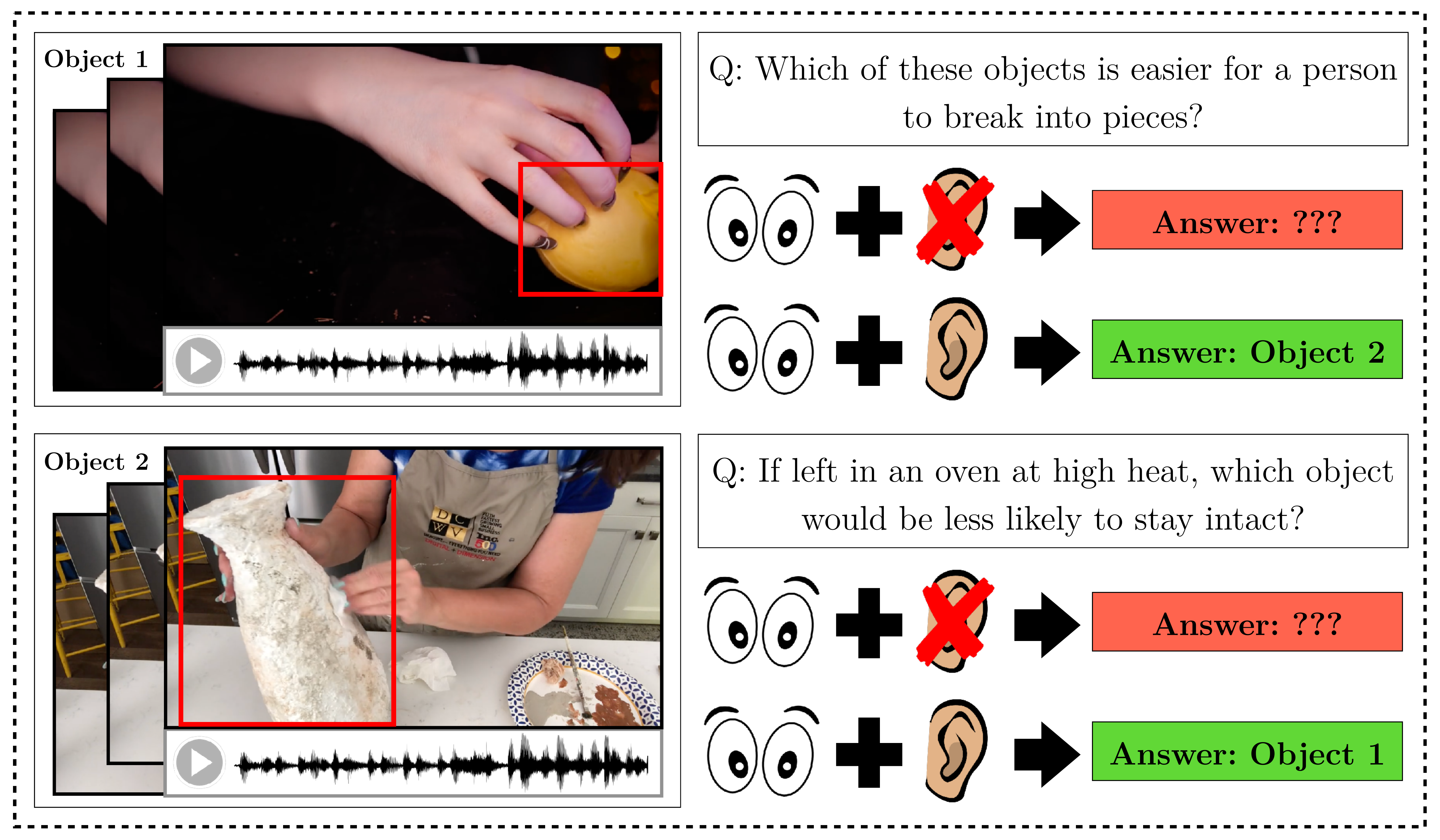}
  \caption{\names\ is the first audiovisual benchmark annotated for physical commonsense attributes, containing \numprint{13400} question-answer pairs, \numprint{1526} videos, and \numprint{1377} unique questions. By benchmarking state-of-the-art unimodal and multimodal models to highlight \textit{where} and \textit{why} current models fail, \names\ provides new opportunities to advance the research of physical reasoning through studying multimodal reasoning. This figure shows two example datapoints from \names, with each datapoint containing a question and a pair of objects (in this figure, object 1 is a plastic lemon and object 2 is a ceramic vase). To view the video clips, please see the supplementary material.}
  \label{fig:dataset_example}
\end{figure}

To safely interact with everyday objects in the real world, AI must utilize physical commonsense knowledge about everyday objects: including their physical properties, affordances, how they can be manipulated, and how they interact with other physical objects~\cite{bisk2020piqa,hespos2004conceptual}. Humans use \textit{physical commonsense reasoning} in all facets of day-to-day life, whether it is to infer properties of previously unseen objects (``the water bottle over there is made of plastic, not glass''), or to solve unique problems (``I can use a puffy jacket in place of my missing pillow'') \cite{bisk2020piqa}. This type of general understanding of object interactions is necessary in building robust and complete AI systems that can be safely deployed in the real world (e.g., a package delivery robot needs to treat heavier or lighter objects differently).

Physical commonsense reasoning is fundamentally a multi-sensory task, as physical properties are manifested through multiple modalities, including vision and acoustics~\cite{corlett2018conditioned,minsky2000commonsense,zhang2017shape}. If two objects appear similar visually, audio can provide valuable information to distinguish the physical properties between these objects. For example, in Figure~\ref{fig:dataset_example}, instead of plastic, object 1 could be mistaken for squishy foam, and instead of ceramic, object 2 could be mistaken for painted plastic, glass, or even paper. Without the necessary audio information, this could result in the erroneous answer that object 1 is easier to break than object 2. In the real world, this misunderstanding may lead to the damaging or mishandling of an object. Therefore, to enable physical commonsense reasoning in AI, it is essential for these models to reason across both audio and visual modalities.

Recent work has explored the use of vision and/or text to understand basic physical properties~\cite{hessel2018quantifying,jimenez2020learning,krishna2017visual,Storks2021TieredRF,yatskar2017commonly,zhao-etal-2020-learning}, or benchmark \textit{physical commonsense} in language~\cite{bisk2020piqa,Forbes2019DoNL}. Our work complements these previous settings by adding the acoustic modality as part of the problem formulation. Furthermore, we include not only static frames but also temporal information using videos. In these directions, our paper takes a step towards real-world physical commonsense reasoning by contributing \namel\ (\names): the first \textit{audiovisual} benchmark annotated for physical commonsense attributes. \names\ contains a total of \numprint{13400} question-answer pairs, involving \numprint{1526} object-oriented videos that cover a diverse set of objects, and \numprint{1377} unique physical commonsense questions involving a variety of physical properties. 

In our paper, we first detail the construction of our new audiovisual benchmark of physical commonsense and establish the need for both the audio modality and commonsense reasoning to succeed on our task. Using this benchmark, we evaluate the performance of multiple state-of-the-art unimodal and multimodal models in comparison with human performance. We also performed an analysis of \textit{where} and \textit{why} current models fail, highlighting the increased difficulty of reasoning about physical commonsense, the lack of fine-grained temporal information due to limitations in current models' video and audio processing, and the need for more advanced audiovisual models. We hope our work will elicit further research into building robust multimodal representations of the physical world. \footnote{For dataset download links, benchmarked models, and evaluation scripts, please visit \href{https://github.com/samuelyu2002/PACS}{\color{red}\underline{https://github.com/samuelyu2002/PACS}}.}


\section{Related Work}
\label{sec:related_work}

We cover related work in commonsense reasoning, particularly on physical understanding, which has been studied in domains spanning psychology, language, vision, robotics, and multimodal machine learning.

\noindent
\textbf{Psychology:} Physical commonsense was first studied in humans, with psychology experiments based on naive and intuitive physics~\cite{bobrow1984qualitative,forbus1984qualitative,hayes1987knowledge,hespos2016objects,mccloskey1983intuitive}. In these experiments, humans are asked to predict object motion or the result of multi-object interactions. Further research has also been conducted in general physical modeling~\cite{bliss2008commonsense} and the multisensory perception of physical properties~\cite{corlett2018conditioned,minsky2000commonsense,zhang2017shape}. In particular, studies on human behavior indicate that the audio modality contains valuable information about the physical properties of objects~\cite{handel1995timbre,minsky2000commonsense,morrongiello1998crossmodal,wilcox2006shake}.

\noindent
\textbf{Language:} Related work has studied physical commonsense within the text modality~\cite{bisk2020piqa,Forbes2019DoNL,jimenez2020learning,Storks2021TieredRF,zhao-etal-2020-learning}. To our knowledge, the generalizability of their findings to other modalities is still understudied. Our dataset extends these text-based knowledge graphs and language models to multimodal settings.

\noindent
\textbf{Vision:} Methods utilizing physical commonsense have been applied to several visual commonsense tasks, including scene understanding~\cite{Chen_2019_ICCV,wu2017deanimation}, activity recognition~\cite{li2022hake}, and cause-effect prediction~\cite{mottaghi2016whatif}. We note that these methods focus solely on the visual modality, which may bring challenges in tasks with unknown or occluded objects. Including information from other modalities such as audio and language could help mitigate these challenges.

\noindent
\textbf{Audio} provides valuable information for one's understanding of the world~\cite{handel1995timbre,wilcox2006shake}. Currently, AI tasks studying physical properties through the lens of the audio modality include navigation~\cite{chen2020soundspaces}, perception~\cite{zhang2017shape}, and generative modeling~\cite{Zhang_2017_ICCV,gao2021ObjectFolder}. We extend this research direction to higher-order reasoning through \names.

\noindent
\textbf{Robotics:} Comprehension of physical properties has been shown to be valuable for tool usage and object manipulation tasks~\cite{agrawal2016learning,coumans2021,nair2019tool,toussaint2018differentiable,tuli2021tango}. Our paper provides a direction for generalizing physical commonsense reasoning utilizing both audio and visual modalities.

\noindent
\textbf{Multimodal:} Recent work has introduced question-answering datasets with image and text inputs (e.g., VQA~\cite{antol2015vqa}, NLVR~\cite{suhr2017nlvr}, NLVR2~\cite{suhr2019nlvr2}), with some annotated for commonsense reasoning tasks (e.g., VCR~\cite{zellers2019vcr,zellers2022merlotreserve}, VisualCOMET~\cite{park2020visualcomet}). There has also been the use of multimodal answer choices, such as a combination of text and image regions in VCR~\cite{zellers2019vcr} and VisualCOMET~\cite{park2020visualcomet}. Other works have also introduced datasets with video and text inputs to test for temporal reasoning (e.g., MovieQA~\cite{tapaswi2016movieqa}, MovieFIB~\cite{maharaj2017dataset}, TVQA~\cite{lei2018tvqa,zellers2022merlotreserve}). To our knowledge, none of these approaches have explored audio and video together for physical commonsense reasoning.

\section{\names\ Dataset}

We introduce \names, a benchmark dataset designed to help create and evaluate a new generation of AI algorithms able to reason about physical commonsense using both audio and visual modalities. The underlying task is binary question answering, where given a question $\boldsymbol{q}$ and objects $\boldsymbol{o_1}, \boldsymbol{o_2}$, the model must pick the more appropriate object to answer the question. Each object is represented by a video $\boldsymbol{v}$ showing a human interacting with the object, the corresponding audio $\boldsymbol{a}$, and a bounding-box $\boldsymbol{b}$ drawn around the object in the middlemost frame of $\boldsymbol{v}$.\footnote{In our experiments, we usually represent the bounding box $\boldsymbol{b}$ as a red bounding box drawn directly on the middlemost frame of the video. Thus, we also interchangeably notate the bounding box as an image $\boldsymbol{i}$.} Thus, each datapoint in \names\ is a tuple of values $(\boldsymbol{q}, (\boldsymbol{b_1}, \boldsymbol{v_1}, \boldsymbol{a_1}), (\boldsymbol{b_2}, \boldsymbol{v_2}, \boldsymbol{a_2}), \boldsymbol{l})$, representing the question, two objects, and a binary label of which object is the correct answer (see Figure \ref{fig:dataset_example} for an example datapoint in our dataset).

In this section, we first outline various design principles used in the creation of our dataset. Then, we give an overview of \names\ statistics (see Figure \ref{fig:dataset_stats} for a complete overview), and finally discuss each component of our data collection and annotation process (see Figure \ref{fig:dataset_pipeline} for our complete annotation pipeline). For a more detailed overview of our data collection pipeline, please refer to section~\ref{appendix:data}.

\subsection{Design Principles}
\label{subsec:design}

Through synthesis of previous work, we divide physical commonsense into two main categories based on which we designed \names. These categories were used as guidance for annotators when creating physical commonsense questions.
\begin{enumerate}
    \item \textbf{Intuitive physics, and a functional world model:} This category is inspired by previous psychology and AI experiments relating to physical commonsense, such as predicting object motion~\cite{kaiser1986intuitivephysics,kim2001intuitivephysics,smith2013physics,wu2015physics}, or how objects interact with each other~\cite{hespos2016objects}. Questions in this category focus on predicting the result of single or multi-object interactions. Easy questions involve a single object and action, such as: \textit{``Which object will break after being dropped on the ground?'' (\textcolor{gg}{a vase}, a ball of paper)}. Harder questions involve multiple objects or actions, including interactions between the two objects, such as: \textit{``Which object will become deformed if the other object is placed on top of it?'' (a vase, \textcolor{gg}{a ball of paper})}.
    \item \textbf{Common real-world knowledge:} This category is inspired by previous commonsense datasets, which test for more concrete understandings of how and why humans or objects function in the real world~\cite{bisk2020piqa,Forbes2019DoNL,zadeh2019socialiq,zellers2019vcr}. Questions in this category ask about possible uses of an object in real-life scenarios. Importantly, these scenarios focus on less prototypical uses of an object, therefore reducing the possibility of abusing learned knowledge \cite{bisk2020piqa}, such as \textit{``Which object is better suited to clean up a watery mess'' (\textcolor{gg}{an old t-shirt}, a plastic box)}. Harder questions can introduce more complicated or uncommon scenarios involving multiple objects: \textit{``If I were to stack the two objects, which would logically go on the bottom?'' (an old t-shirt, \textcolor{gg}{a plastic box})}.
\end{enumerate}

\subsection{Dataset Statistics}

\begin{figure}[t!]
    \centering
    \begin{subfigure}[t]{0.37\textwidth}
    \centering
      \includegraphics[width=\linewidth]{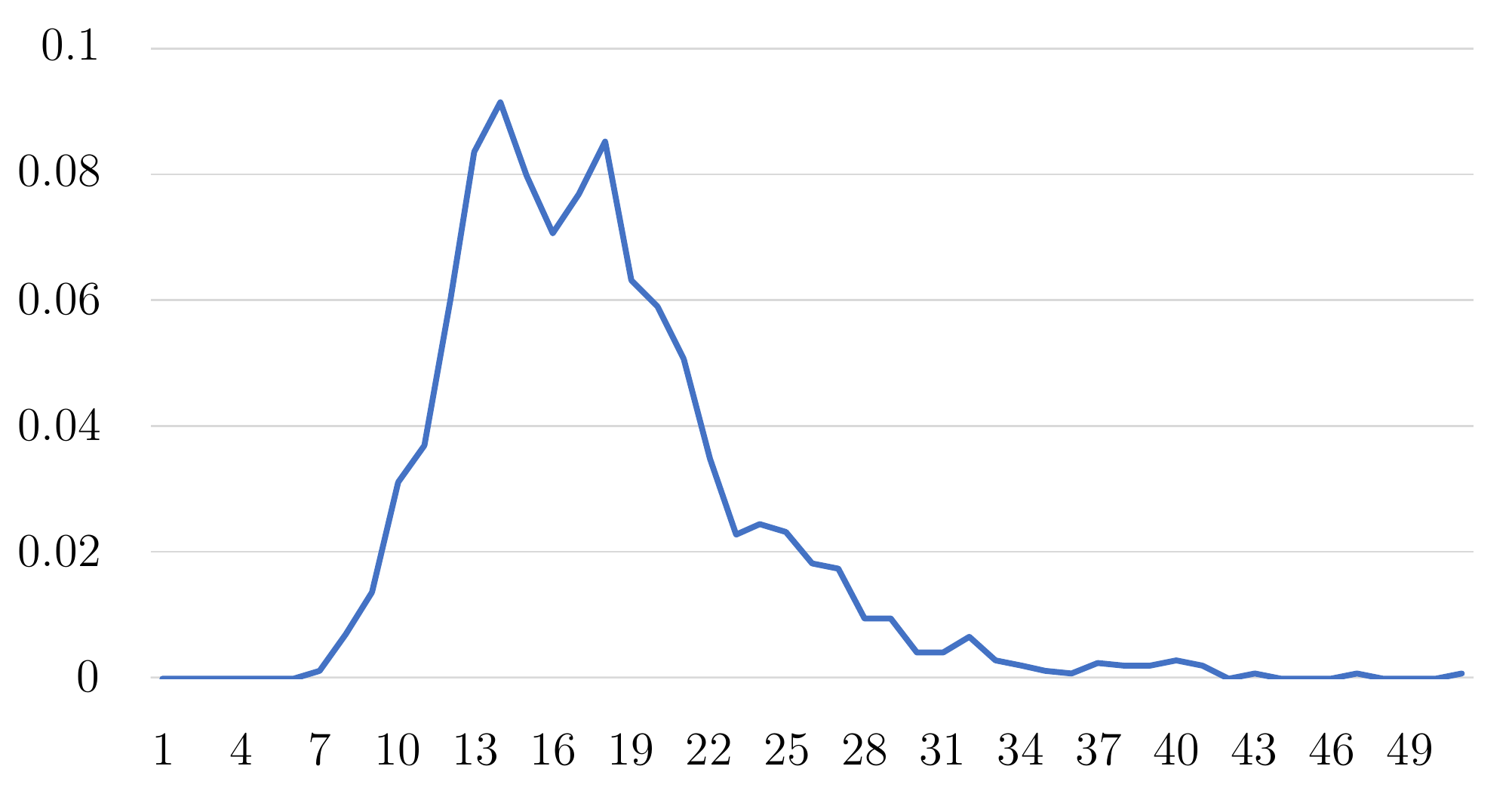}
      \subcaption[]{Question length distribution (as tokenized by spaCy).}
      \label{fig:qlengths}
    \end{subfigure}
    \hfill
    \begin{subfigure}[t]{0.28\textwidth}
        \centering
      \includegraphics[width=\textwidth]{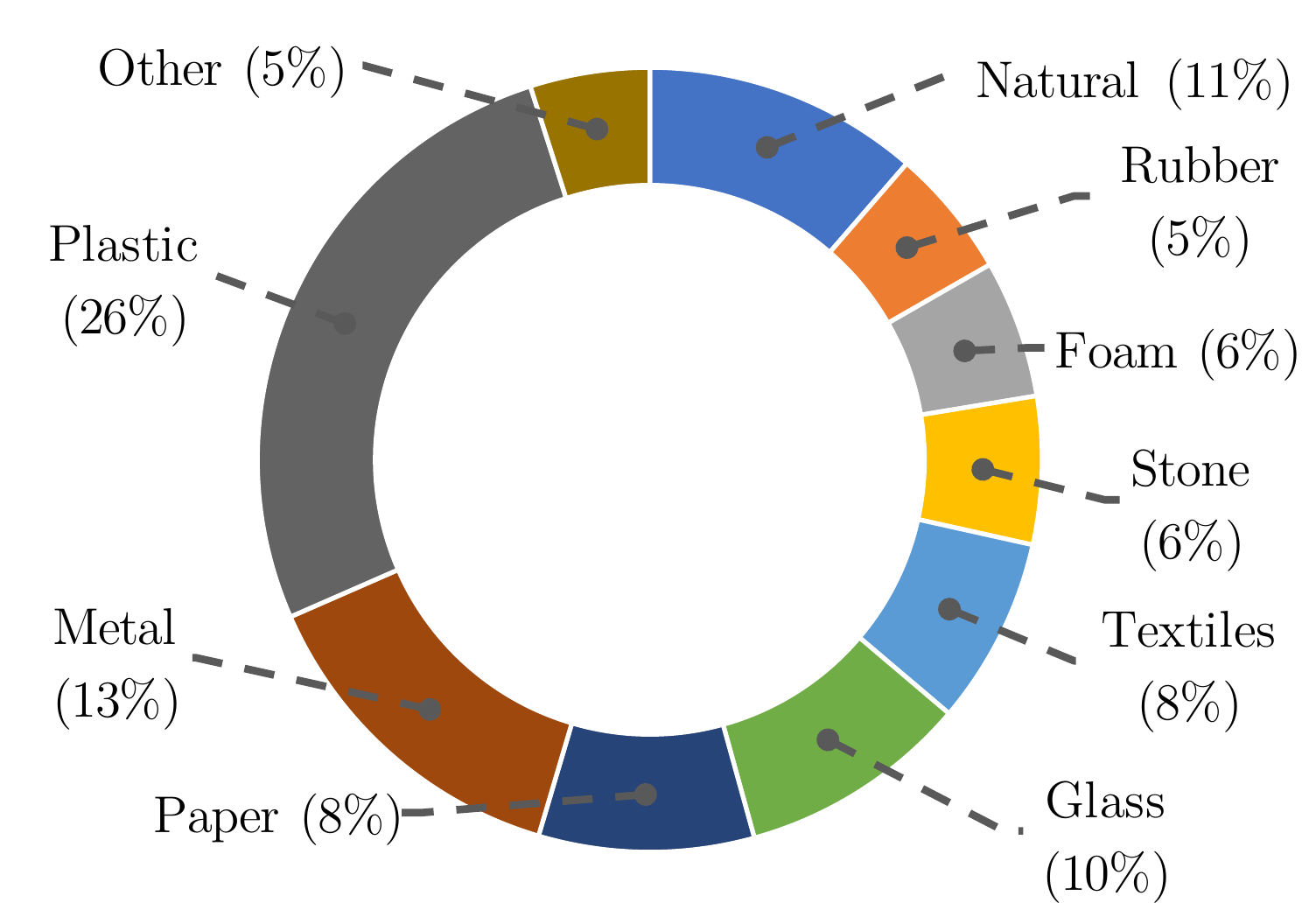}
      \subcaption{Distribution of materials making up each object.}
      \label{fig:materials}
    \end{subfigure}
    \hfill
    \begin{subfigure}[t]{0.28\textwidth}
        \centering
      \includegraphics[width=\textwidth]{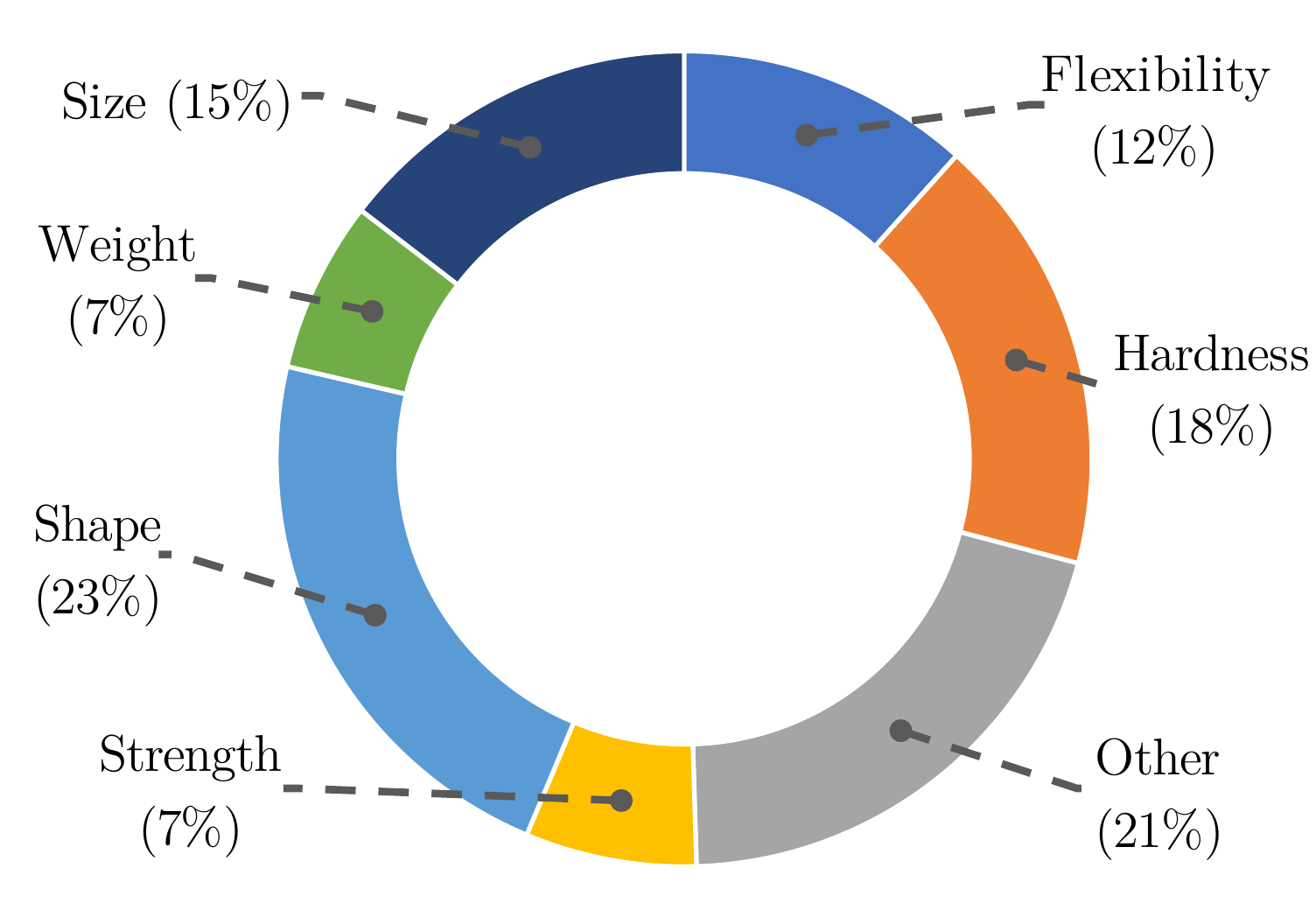}
      \subcaption{Distribution of relevant physical properties.}
      \label{fig:properties}
    \end{subfigure}
    \begin{subfigure}[t]{0.37\textwidth}
        \centering
      \includegraphics[width=\textwidth]{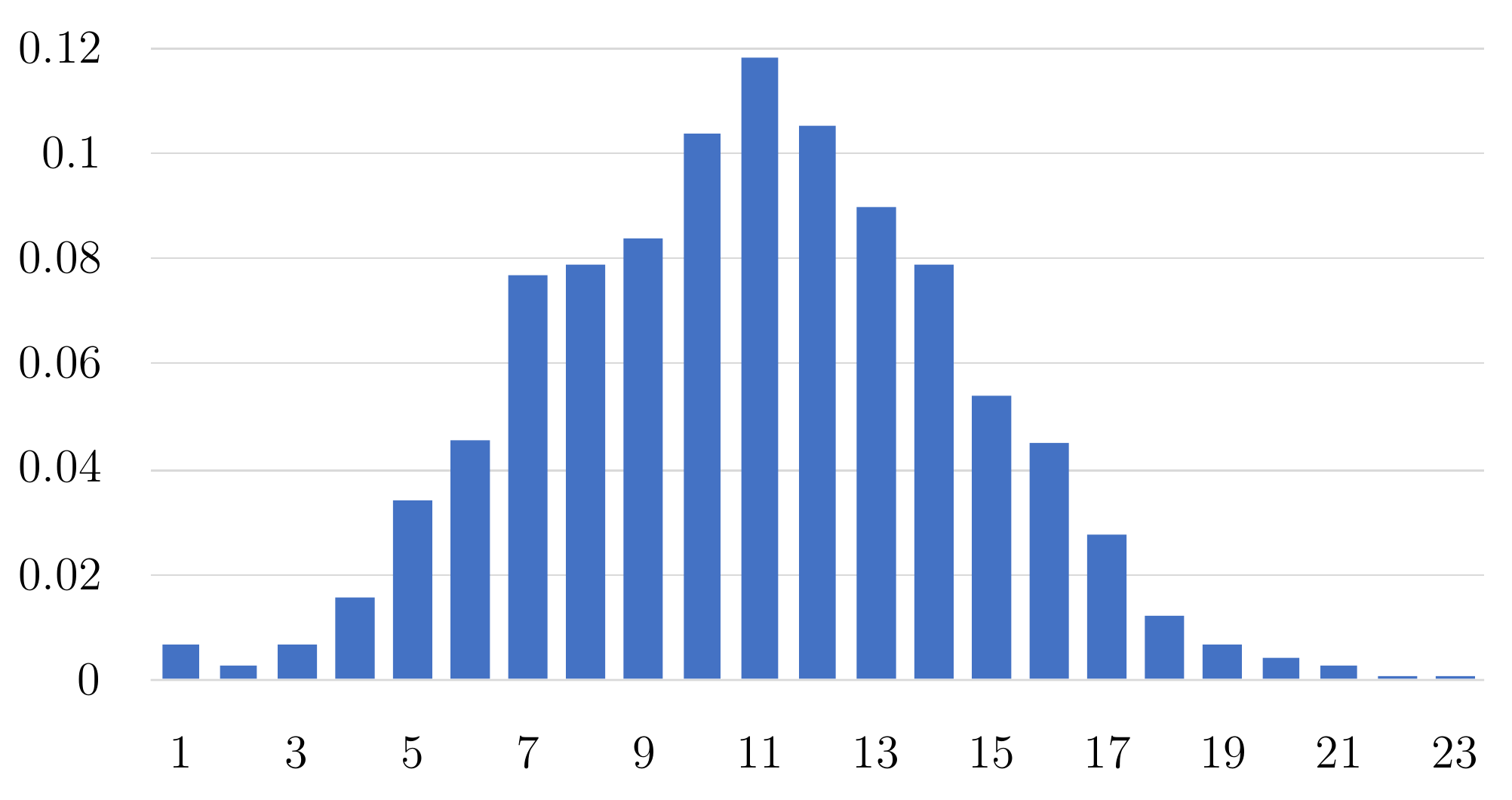}
      \subcaption{Number of times each question was reassigned to new object pairs.}
      \label{fig:qfreq}
    \end{subfigure}
    \hfill
    \begin{subfigure}[t]{0.28\textwidth}
      \includegraphics[width=\textwidth]{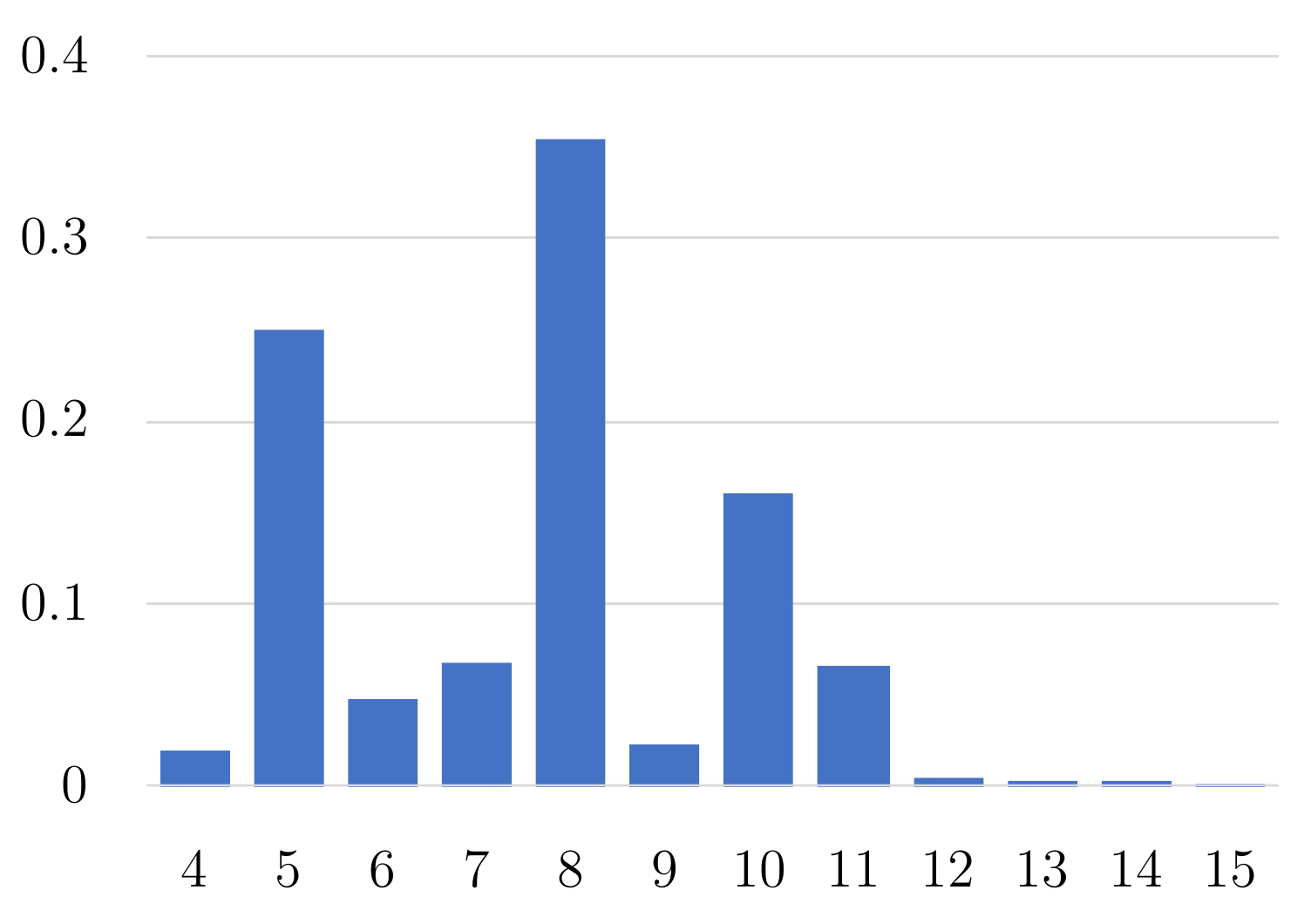}
      \subcaption{ Duration of videos (rounded to nearest second).}
      \label{fig:vid_durations}
    \end{subfigure}
    \hfill
    \begin{subfigure}[t]{0.28\textwidth}
    \centering
        \includegraphics[width=\textwidth]{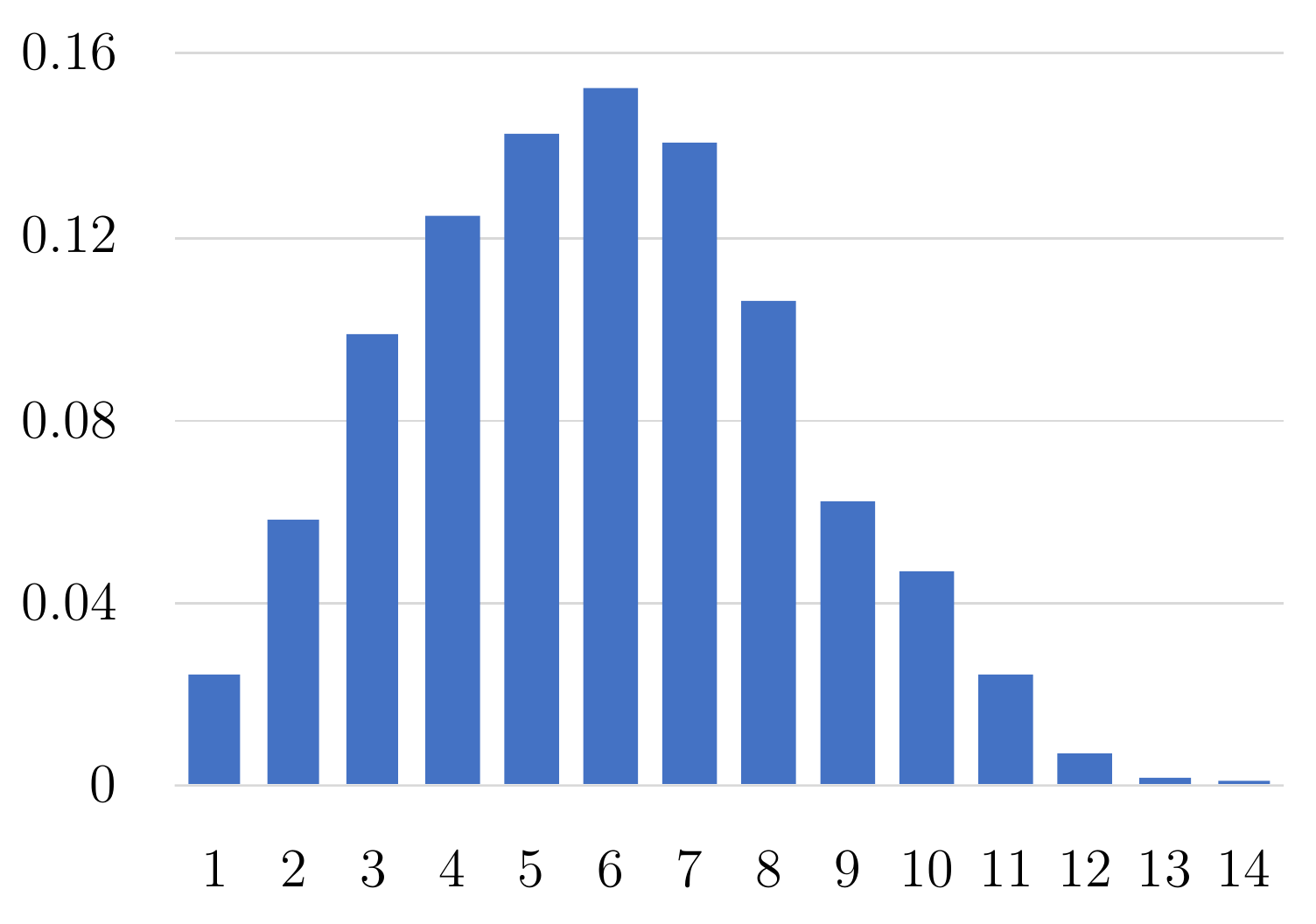}
         \subcaption{Number of questions per pair of objects.}
         \label{fig:qpairs}
    \end{subfigure}
    \begin{subfigure}{\textwidth}
        \centering
      \includegraphics[width=\textwidth]{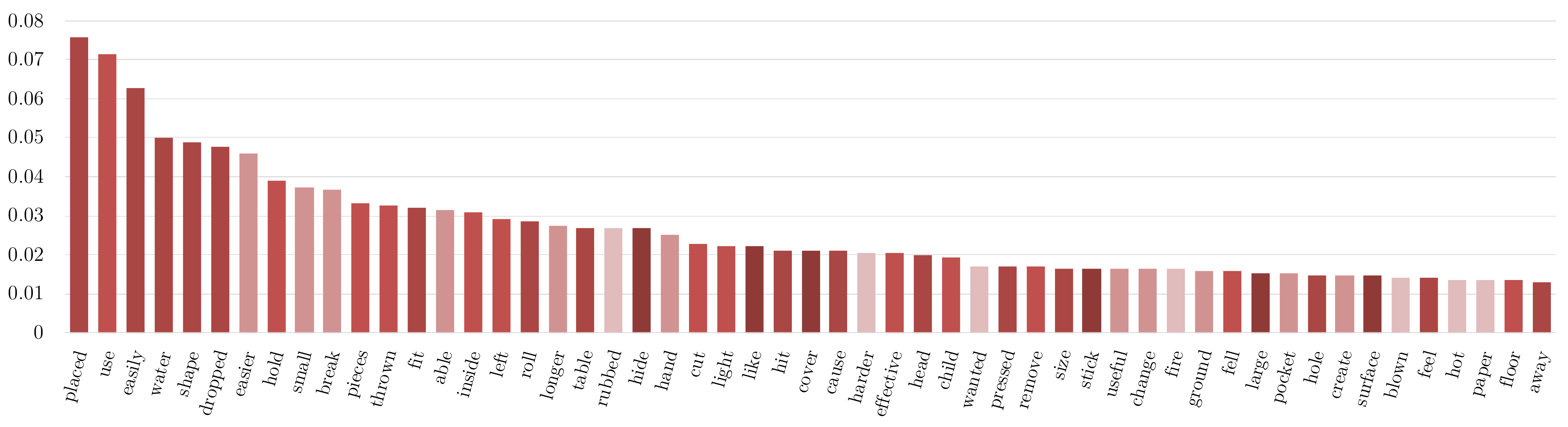}
      \subcaption{Frequency of most common words in \names. The top 4 words (object, item, likely, better) are excluded due to their high frequency being a result of our problem formulation. The bars are colored based on the accuracy that AudioCLIP~\cite{guzhov2021audioclip} achieves on them, with darker being higher accuracy.}
      \label{fig:common_words}
    \end{subfigure}
    \caption{Dataset statistics for \names. Best viewed zoomed-in and with color. Figure~\ref{fig:materials} and Figure~\ref{fig:properties} show that the questions and objects in our dataset are diverse, involving different physical properties and materials. Figure~\ref{fig:common_words} shows a variety of actions (e.g., placed, dropped, thrown, roll, rubbed, pressed, blown) covered in our diverse questions.}
    \label{fig:dataset_stats}
\end{figure}

This subsection presents the main object and question statistics of \names. Each datapoint is the combination of a question, two objects, and the correct answer. Figure~\ref{fig:qpairs} shows the distribution of the number of questions relating to each object pair, with an average of $5.86$ questions per pair. 

\noindent \textbf{Object statistics:} \names\ contains a total of \numprint{1526} objects, each represented by a unique video clip, with included audio and a bounding box in the middlemost frame of the video. Figure \ref{fig:materials} shows a rough distribution of materials that the objects in our dataset are made of, as annotated in our video filtering step. Materials such as ``Wax'' or ``Foam'' occur more commonly in our dataset than in real life, due to our focus on creating a diverse set of objects. Figure \ref{fig:vid_durations} shows the length of each video. On average, videos in our dataset are $7.6$ seconds long.

\noindent \textbf{Question statistics:} \names\ contains a total of \numprint{1377} unique questions each used multiple times across various pairs of objects. Figure \ref{fig:qfreq} shows how many times each question was used, where on average, a question was distributed to $10.8$ pairs of videos. Figure \ref{fig:qlengths} shows the distribution of question length in terms of the number of words. On average, a question was $16.6$ words long. Figure \ref{fig:properties} shows the distribution of physical properties that our questions relate to. Figure \ref{fig:common_words} shows the most commonly occurring words in our dataset and is also color-coded by CLIP's accuracy on datapoints conditioned on the occurrence of a specific word. We can see a variety of action words (e.g., placed, dropped, thrown, roll, rubbed, pressed, blown), each associated with different physical properties. Furthermore, we see that AudioCLIP struggles with certain physical concepts, such as having low accuracy on heat-related words (e.g., hot, fire).

\subsection{Dataset Creation}

\begin{figure}[t!]
\begin{framed}
    \centering
    \begin{subfigure}{\textwidth}
        \centering
        \caption{\textbf{Video collection}: We first downloaded YouTube videos and split them into 5-10 second long clips. In this example, the first 5 clips came from a video with the query ``ASMR slime no talking'', and the last 5 came from a video with the query ``ASMR plastic no talking''.}
        \includegraphics[width=\textwidth]{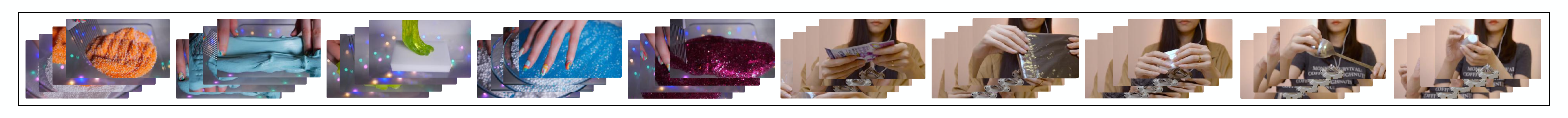}
    \end{subfigure} 
    \begin{subfigure}{\textwidth}
        \centering
        \caption{\textbf{Video clip annotation and filtering}: Clips were filtered with an audio classifier and sparsely sampled to be sent for human filtering. Clips that passed human filtering were annotated with a bounding box (denoting the object) and added to the final dataset.} 
        \includegraphics[width=\textwidth]{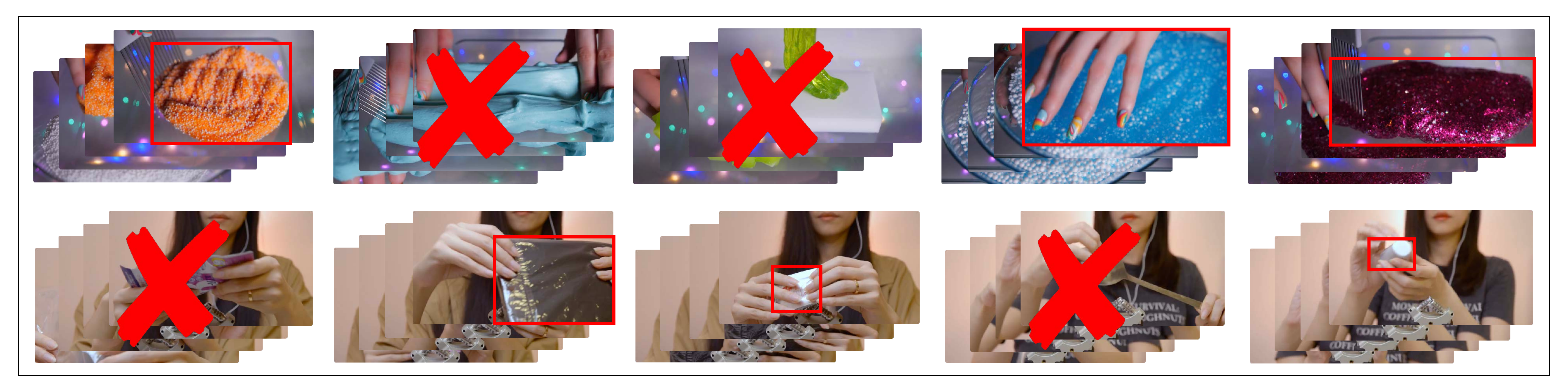}
    \end{subfigure}
    \begin{subfigure}{\textwidth}
        \centering
        \caption{\textbf{Question creation}: Each object was randomly paired with three other objects, and a subset of the object pairs was given to annotators to create physical commonsense questions.} 
        \includegraphics[width=\textwidth]{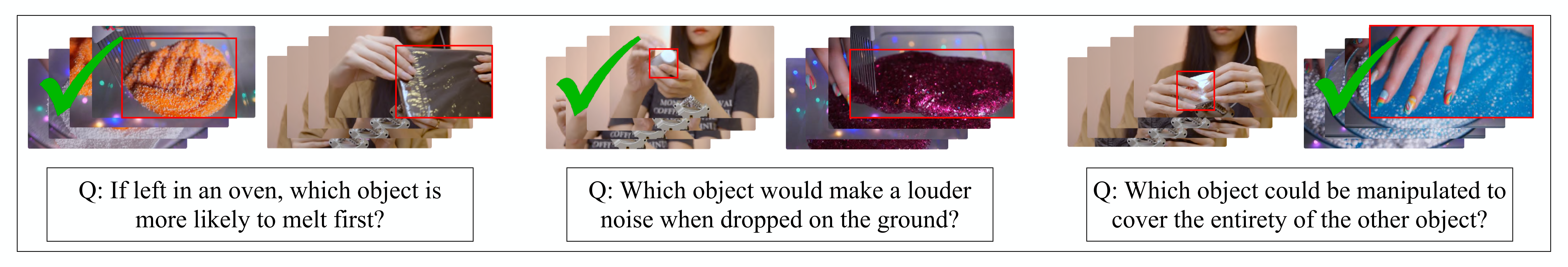}
    \end{subfigure}
    \begin{subfigure}[b]{0.48\textwidth}
        \centering
        \caption{\textbf{Question reassignment}: Questions created in the previous step were randomly distributed to unannotated object pairs. Annotators removed irrelevant questions.}
        \includegraphics[width=\textwidth]{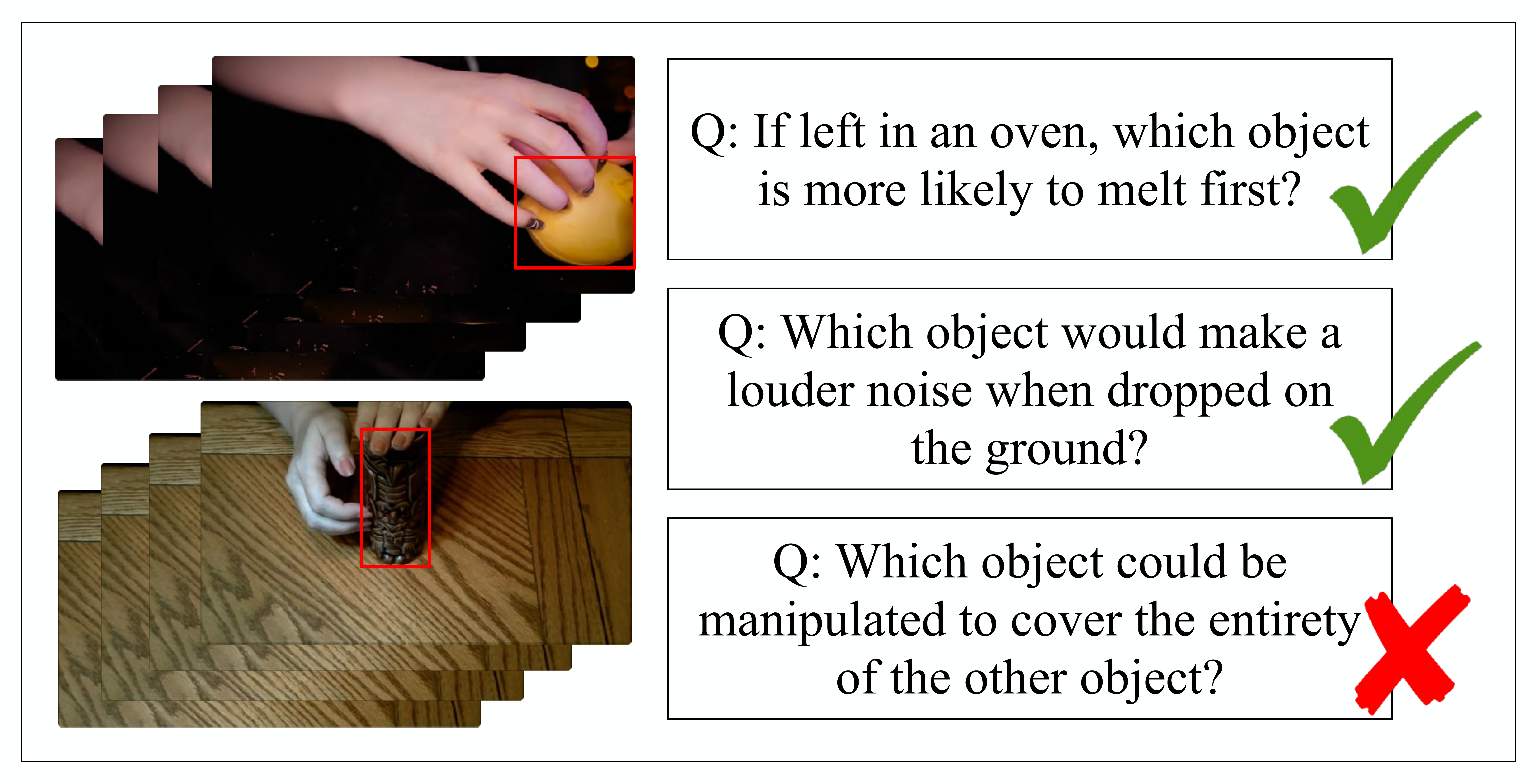}
    \end{subfigure}
    \hfill
    \begin{subfigure}[b]{0.48\textwidth}
        \centering
        \caption{\textbf{Quality checking}: The remaining datapoints were answered by additional annotators, and datapoints without unanimous agreement were removed. } 
        \includegraphics[width=\textwidth]{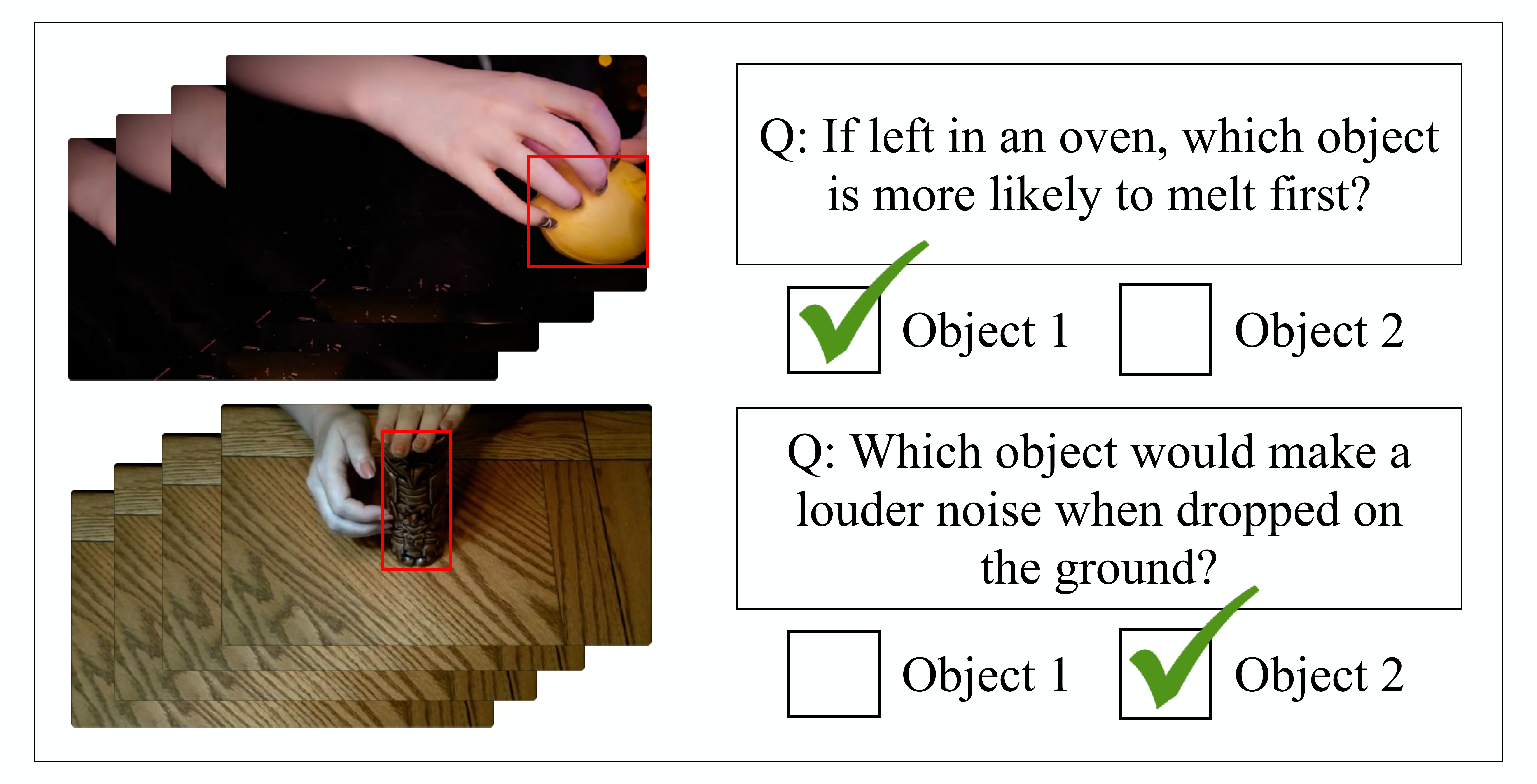}
    \end{subfigure}
    \end{framed}
\caption{Diagram of our data collection process, showing steps starting from gathering objects, to creating and checking datapoints. Best viewed zoomed-in and with color.}
\label{fig:dataset_pipeline}
\end{figure}

In this subsection, we outline the steps used to gather and label datapoints in \names\ (see Figure~\ref{fig:dataset_pipeline} for a complete overview).

\noindent \textbf{(a) Video collection}: A broad set of ASMR videos were downloaded from YouTube. Specifically, we chose to use object-oriented ASMR videos
, as they provide high-quality audio, and often incorporate objects that people less commonly interact with.
We used a list of materials \cite{materiallist} to seed the search queries, which was later updated with more materials as we iterated through the first two data collection steps.
For each video, we use a shot boundary detector~\cite{souvcek2020transnet} to split each video into separate scenes, and then further split each scene into roughly 5-10 second long clips. Finally, an audio classifier~\cite{giannakopoulos2015pyaudioanalysis} was used to remove videos with background music, talking, or silence. The remaining clips were sparsely sampled to create the candidate set of clips.

\noindent \textbf{(b) Video clip annotation and filtering}: When analyzing the candidate set of clips, we noticed that a large number of objects that appeared in these clips were common household objects, resulting in many repeated objects. Furthermore, common objects do not require as much multimodal understanding, as a single image and a decent knowledge base may be enough to identify the object and extract necessary physical properties. Thus, as a heuristic for how common or obvious an object is, we test to see if annotators are able to classify the materials each object is made of. If annotators are able to correctly identify an object's materials using just a single image, then this suggests that the object is likely common, and has physical properties that are easily distinguishable. 

In this task, annotators were first given a single image from a candidate video clip and asked to draw a bounding box around the ``object of focus'', which we define as the object the person is touching in the video (if the guess is wrong, the clip is thrown away). Then, they were were asked to select the materials that make up the object from a list, and to provide a confidence score from $1-5$. Once they submitted their initial answer, annotators were then given access to the whole video and audio and asked to redo the task. If their confidence did not increase and their answers did not change, then the clip was removed. Otherwise, the clip and the bounding box were added to the dataset as an object, with each clip containing exactly one bounding box annotation (one object). 


The final set of \numprint{1526} objects was partitioned into train, test, and validation of \numprint{1224}, $152$, and $150$ videos respectively. Then, each object was paired with three other objects in the same subset, resulting in \numprint{2289} pairs of objects. 

\noindent \textbf{(c) Question creation}: From the \numprint{2289} object pairs gathered, $242$ were randomly selected to be used in this step, while the other \numprint{2047} pairs were used in the next step. In this step, annotators were asked to write questions that require physical commonsense knowledge to answer. Annotators were given two videos, and a frame from each video containing a bounding box that specified the object. The had the option to write one or two commonsense questions related to the pair of objects, and answer with ``Object 1'' or ``Object 2''. In total, \numprint{1377} questions were created, with each pair of videos given to $5$ separate annotators. 

To facilitate the process of creating high-quality questions, we provided annotators with a more detailed version of the categorization developed in section~\ref{subsec:design} as guidance for what constitutes physical commonsense as instructions. They were also required to provide at least one relevant physical property
for each question to encourage topical questions. Finally, questions were required to have a certain level of complexity, and were all quality-checked (e.g., questions that directly asked about a physical property such as \textit{``Which object is more sticky''}, or \textit{``Which object is larger?''} were forbidden).

\noindent \textbf{(d) Question reassignment}: We evenly redistribute the \numprint{1377} questions created in the previous step to the remaining \numprint{2047} object pairs. Reusing questions on new pairs of objects can create interesting scenarios, as it matches object pairs with questions that human annotators may not normally come up with~\cite{bisk2020piqa}. The goal is to create matchings such as: \textit{``If you absolutely needed to tie your hair up, which item would you use?'' (\textcolor{gg}{a plastic straw}, a piece of paper)}. In this example, the question and object pair are not normally associated with each other, but are still answerable by humans, who have the ability to draw new connections. This puts more of the challenge on drawing relationships between physical properties, rather than directly applying past knowledge.

Specifically, in this task, each unused object pair is assigned a list of 13 questions, which is then given to annotators.
Then, annotators can either mark each object-question matching as ``completely irrelevant'', or choose to answer the question, thus creating a new datapoint. 


\noindent \textbf{(e) Quality checking}: To ensure the quality of final datapoints, each candidate datapoint gathered from the Question Creation and Question Reassignment stages was given to additional annotators to double-check. Every candidate was answered three times between the question annotation stages and only kept in our dataset if there was unanimous agreement.

\section{Experimental Setup}

In this section, we first outline the setup for testing human performance. We then list the models for checking dataset biases, and several state-of-the-art models that we tested. Finally, we outline the creation of \names-material, a material classification subtask on our dataset.\footnote{For more details on experimental setups, refer to section~\ref{appendix:experimental}.} 

Our experiments were designed to answer the following research questions:
\begin{enumerate}
    \item How difficult is our task, as measured by the performance of human annotators and state-of-the-art models? We evaluate open-source state-of-the-art models that have high performance on comparable datasets such as VCR~\cite{zellers2019vcr}, TVQA~\cite{lei2018tvqa}, and NLVR2~\cite{chen2020uniter} (section~\ref{sec:models}), and compare these results to human performance on \names\ (section~\ref{sec:human}).
    \item Are there potential biases in our dataset? While the paired binary question answering format is designed to limit bias in the language modality (correlations between questions and correct vs incorrect answers) as opposed to standard QA datasets~\cite{agrawal2016analyzing,anand2018blindfold,jabri2016revisiting,zadeh2019socialiq}, we explore other sources of biases in language, video, and audio in \names\ (section~\ref{sec:bias}). 
    \item What is the importance of audio in our task, and what are the specific areas where audio is beneficial? We compare human and model performance with and without audio (with otherwise the same configurations) and analyze specific qualitative examples where including audio leads to better results (see section~\ref{sec:human} and section~\ref{sec:models} for how we set up human and model benchmarks).
    \item How challenging is the level of reasoning required to capture physical commonsense? To establish this difficulty, we create an additional material classification task to compare with our physical commonsense task (section~\ref{sec:material}).
\end{enumerate}

\subsection{Human Performance} \label{sec:human}

To test human performance with and without audio, we randomly sampled $243$ datapoints from the dataset, and give them to $10$ annotators to answer. The annotators were given half of the datapoints with audio and half without, such that each datapoint would be annotated with five answers with audio, and five answers without. Consistent with other works, we compute human accuracy as a majority vote \cite{bisk2020piqa,zellers2019vcr}, and also report $90\%$ confidence intervals for the results.

\subsection{Detecting Biases} \label{sec:bias}

We construct four different combinations of late-fusion models by combining state-of-the-art pre-trained image, audio, video, and text models. We used ViT~\cite{dosovitskiy2020vit} as the image model, AST~\cite{gong2021ast} as the audio model, TDN~\cite{wang2021tdn} as the video model, and DeBERTa-V3~\cite{he2021debertav3,he2021deberta} as the text model. The specific configurations chosen for bias detection were inspired by past work studying bias on Visual Question Answering datasets \cite{cadene2019rubi,yang2020gives,zellers2019vcr}. We test for two main types of bias: \textit{answer choice bias} (are there systematic biases in the answer choices that give away the correct answer without even seeing the question?), and \textit{unimodal question-answerability} (is information from one modality enough to correctly answer the question?).

\noindent\textbf{I + A + V}: We study the predictability of our task given only information about the objects (no question is provided). This test demonstrates whether there is a pattern between the objects and the correct answer.

\noindent\textbf{Q + I}: Evaluates the usefulness of images (I) in predicting correct answers.

\noindent\textbf{Q + V}: Evaluates the usefulness of videos (V) in predicting correct answers.

\noindent\textbf{Q + A}: Evaluates the usefulness of audio (A) in predicting correct answers.

\subsection{Baseline Models} \label{sec:models}

\textbf{Late Fusion \cite{pandeya2021fusion}}: We train a model using late fusion of all four input modalities as a simple baseline. We use SOTA image~\cite{dosovitskiy2020vit}, audio~\cite{gong2021ast}, and video~\cite{wang2021tdn} models pretrained on large-scale classification datasets such as ImageNet21k~\cite{ridnik2021imagenet21k}, AudioSet~\cite{gemmeke2017audioset}, and Something-Something V2~\cite{goyal2017something}, and the text~\cite{he2021debertav3} model is pretrained using replaced token detection. We concatenate the unimodal embeddings and use a linear layer to create multimodal embeddings for prediction. 

\noindent \textbf{CLIP \cite{radford2021clip}} is a powerful image-text model pre-trained on a large set of images and text captions and can be used for a variety of zero-shot and finetuning tasks. CLIP embeds image and text into a shared vector space, where we can use \textit{cosine similarity} to measure the similarity between image and text embeddings.
We use CLIP to separately embed images of both objects and the question. The predicted object is the object with more similar embedding to the question embedding.

\noindent \textbf{AudioCLIP \cite{guzhov2021audioclip}} extends CLIP for audio inputs by training on AudioSet~\cite{gemmeke2017audioset}, which enables the embedding of audio inputs into the same vector space. Using this model, we extend the CLIP model mentioned above to include audio by concatenating the image and audio embedding, and using a linear layer to project them onto the same vector space as the text embedding. 

\noindent \textbf{UNITER \cite{chen2020uniter}} is an image and text model that is pre-trained using four different image-text tasks and achieves strong results on tasks such as NLVR2~\cite{suhr2019nlvr2}. We largely follow the procedure used to prepare and finetune UNITER on the NLVR2 dataset \cite{suhr2019nlvr2}. We split up both objects and generate two object-question embeddings, and finally concatenate them and use an MLP to classify the answer.

\noindent \textbf{Merlot Reserve \cite{zellers2022merlotreserve}} uses image, audio, video, and text, achieving state-of-the-art results on VCR~\cite{zellers2019vcr} and TVQA~\cite{lei2018tvqa}. We follow the methods used to train Merlot Reserve on VCR and TVQA by constructing two multimodal sequences using all input modalities. Then, we separately generate confidence scores for both sequences and compare the two values as a classification output. 

\subsection{Material Classification} \label{sec:material}

By comparing with the simpler task of classification, we can gain an understanding of the level of higher-order reasoning required in our task. In our main question-answering task, errors can come from multiple sources, either from misidentifying the properties of an object, or correctly identifying the objects, but failing to reason about the properties. Results from a material classification task using the same objects can give us an estimate on how much error stems from misidentifed objects, and how much comes from the failure to exhibit higher-order reasoning.

We create a material classification task (\names-material) formulated identically to our dataset, where a pair of objects is accompanied by a comparison question (e.g., \textit{``Which object is more likely to be made out of glass''}). The materials used are gathered from our data-collection stage (Figure~\ref{fig:materials} shows a distribution of material categories). We use the exact same object pairs as in the main task, and accompany each pair with comparison questions based on each object's material.
In total, we created \numprint{3460} training datapoints, \numprint{444} validation datapoints, and \numprint{445} testing datapoints. Each datapoint is a quadruplet $(\boldsymbol{o}^{(1)}, \boldsymbol{o}^{(2)}, \boldsymbol{q}, \boldsymbol{l})$, representing the two objects, the question, and the label.

\section{Results and Discussion}
\label{sec:results}

In this section, we assess the whether audiovisual understanding and physical commonsense reasoning are required to succeed on our dataset, and look at where current models fail. For additional results, refer to section~\ref{appendix:results}. 

\subsection{Human and Model Performance}

\begin{table}[t!]
\setlength{\tabcolsep}{4pt}
\renewcommand{\arraystretch}{1.0}
\centering
\begin{tabular}{l|c:c|c}
\Xhline{3\arrayrulewidth}
\multirow{2}{*}{\textbf{Baseline Model}} & \multicolumn{3}{c}{\textbf{Accuracy (\%)}} \\ \cline{2-4}
    & With audio & Without audio & $\Delta$ \\ \hline
    I + A + V~\cite{yang2020gives,pandeya2021fusion} &    $51.9 \pm 1.1$      &  - &-  \\ 
    Q + I~\cite{zadeh2019socialiq,pandeya2021fusion}     & -          &  $51.2 \pm 0.8$  &-        \\
    Q + A~\cite{zadeh2019socialiq,pandeya2021fusion}& $50.9 \pm 0.6$ &  - & - \\
    Q + V~\cite{zadeh2019socialiq,pandeya2021fusion}& - & $51.5 \pm 0.9$ & - \\ \hline
    Late Fusion~\cite{pandeya2021fusion} & $55.0 \pm 1.1$ & $52.5 \pm 1.6$ & $2.5$ \\
    CLIP/AudioCLIP~\cite{guzhov2021audioclip,radford2021clip} & $60.0 \pm 0.9$ & $56.3 \pm 0.7$ & $3.7$ \\
    UNITER (Large)~\cite{chen2020uniter} & - & $60.6 \pm 2.2$ & - \\
    Merlot Reserve (Base)~\cite{zellers2022merlotreserve} & $66.5 \pm 1.4$  & $64.0 \pm 0.9$ & $2.6$ \\
    Merlot Reserve (Large)~\cite{zellers2022merlotreserve} & $70.1 \pm 1.0$ & $68.4 \pm 0.7$ & $1.8$ \\ \hline
Majority & $50.4$ & $50.4$ & - \\
Human    & $96.3 \pm 2.1$ & $90.5 \pm 3.1$  & $5.9$ \\
\Xhline{3\arrayrulewidth}
\end{tabular}
\caption{Results on \names\ test set: baseline models are reported with the mean and standard deviation of 5 runs, while human accuracy is reported with a $90\%$ confidence interval. There is a large gap between model and human performance, with the best performing model (Merlot Reserve) lagging behind by over $25\%$. Models with audio also consistently outperform the corresponding models without audio, demonstrating the need for information from all modalities to succeed in our task.}
\label{table:all_results}
\end{table}

A summary of all model performances is shown in Table~\ref{table:all_results}. Notably, all methods struggle to achieve results close to human performance, with the gap in accuracy between the best model (Merlot Reserve) and human performance being over $25\%$. This gap is much larger than the gap between SOTA and human performance on other datasets such as TVQA ($3\%$) and VCR ($14\%$)~\cite{zellers2022merlotreserve}
, demonstrating the challenging nature of our dataset.


We believe that the gap in performance comes from (1) the inherent challenge of developing physical commonsense (section~\ref{sec:reasoning}), and (2) the loss of information in each model. This includes the lack of video information in CLIP and UNITER, and the sparse sampling of video frames in the Merlot Reserve and Late Fusion models.
Some physical information may require clear alignment between the actions displayed in the video and the audio signal to accurately understand the object, and thus require more fine-grained temporal information.

\subsection{Checking for Biases in \names}

Table~\ref{table:all_results} shows the performance of our bias testing models, where we see that there is low performance among all configurations of models used. The I+A+V configuration tests for bias among the answer choices (objects), which achieves a low accuracy of $52\%$, demonstrating that the answer choices alone do not give away the answer. Furthermore, solely providing image, audio, or video information alongside the question yields poor performance, and it is only when all three modalities are combined that results solidly deviate from randomly guessing ($55\%$ accuracy).
We believe the low results when provided with unimodal information are because all modalities play an important role. Only the image input specifies the object via a bounding box, thus making it difficult to succeed without the image. Additionally, since our dataset was curated to consist of complex objects that require video and/or audio to understand, removing such modalities also result in low performance.

\subsection{Importance of Audio}
\label{sec:audio_analysis}

In Table \ref{table:all_results}, we can see the benefit of including audio. Perhaps the most important experiment is how much audio helps humans, as the error rate decreases by more than half, with no overlap between the confidence intervals for the two values.
When provided with audio, the models don't seem to improve as much. We theorize a few reasons for this: (1) for Merlot Reserve, the pretraining data is from a very different distribution, mostly consisting of human speech, and the input spectrograms may not be fine-grained enough to capture higher-pitched, sharper noises, such as tapping.
(2) In contrast, AudioCLIP uses raw audio as an input, but the method of fusing audio and video through concatenation may be too simple.

\begin{figure}[t]
    \begin{subfigure}{0.48\textwidth}
        \centering
        \includegraphics[width=0.9\textwidth]{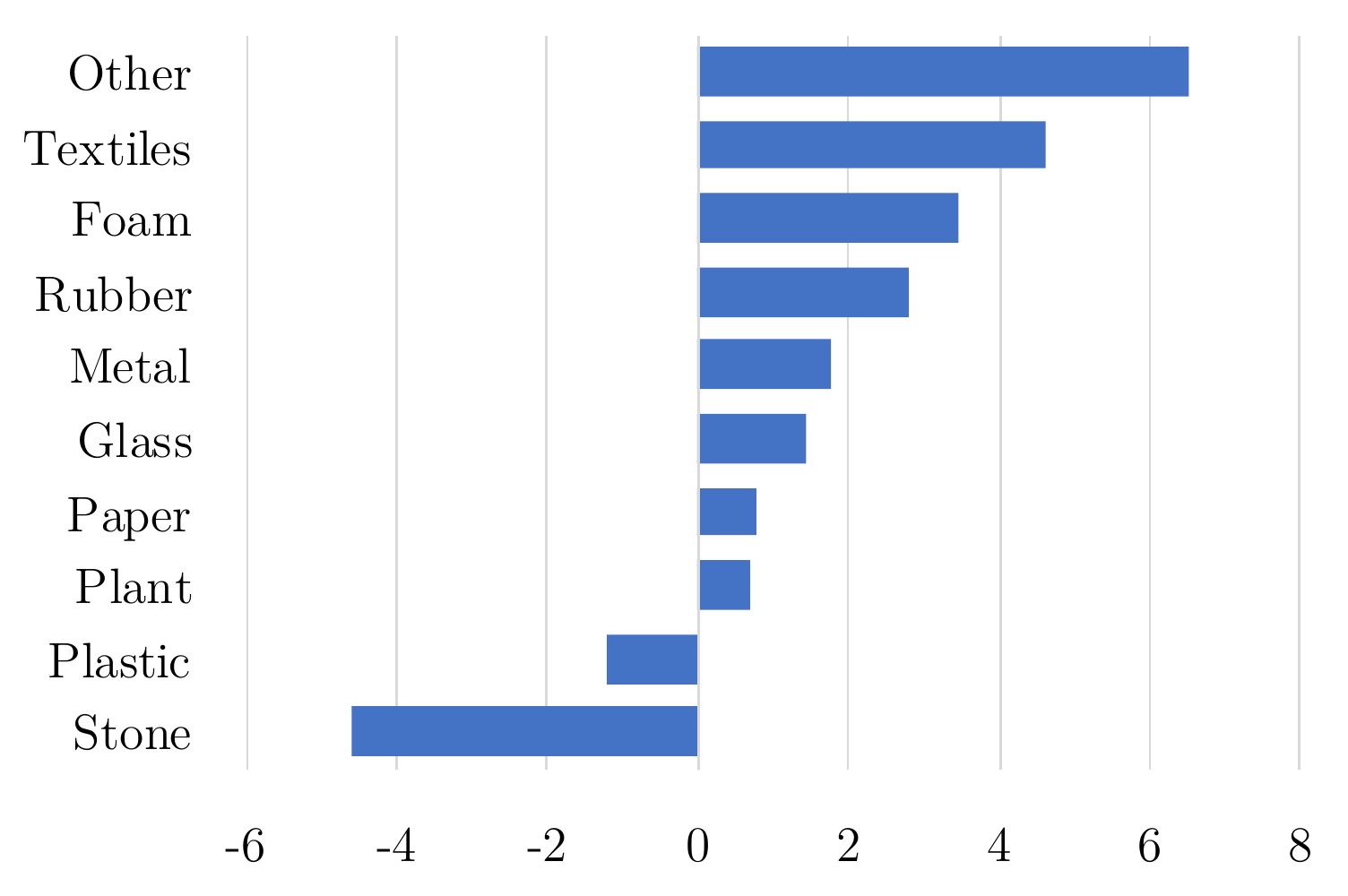}
        \caption{Difference in accuracy with and without audio, conditioned on object materials. }
        \label{fig:acc_merlot_mats}
    \end{subfigure}
    \hfill
    \begin{subfigure}{0.48\textwidth}
        \centering
        \includegraphics[width=0.9\textwidth]{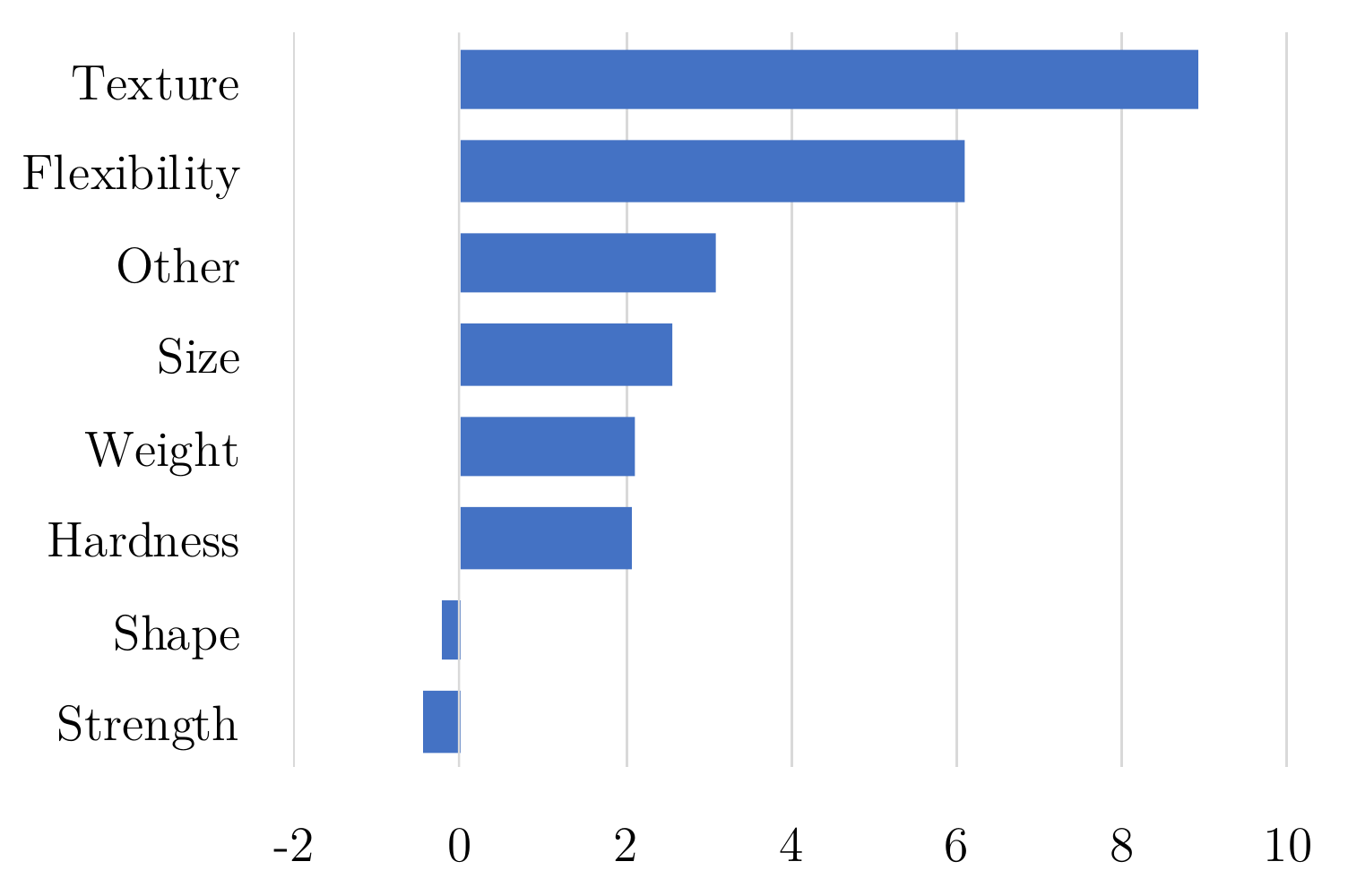}
        \caption{Difference in accuracy with and without audio, conditioned on physical properties. }
        \label{fig:acc_merlot_props}
    \end{subfigure}
    \label{fig:merlot_comp}
    \caption{Comparison of results on Merlot Reserve when trained with and without audio. These results are conditioned on the material of the objects in the object pair, and on the physical properties relevant to the question (see section~\ref{sec:audio_analysis}).}
\end{figure}

\noindent \textbf{Performance on the most ``unique'' objects:} Using the material and physical property labels gathered in the annotation steps, we can also compare results conditioned on specific materials and properties. We calculate performance with respect to a specific material (e.g., metal) by only counting datapoints where at least one of the objects is made of metal. Similarly, we calculate performance with respect to a physical property (e.g., hardness) by only counting datapoints where the question is related to the property. In Figure \ref{fig:acc_merlot_mats}, we see that the biggest improvement in accuracy is on datapoints containing objects made of ``Other'' materials. Since our material labels cover the most common materials appearing in the dataset, this suggests that audio is especially important when reasoning about uncommon objects. From Figure \ref{fig:acc_merlot_props}, we see that properties such as texture and flexibility show the most improvement, and no category's results suffer greatly with the addition of audio.

\begin{figure}[t]
    \centering
    \includegraphics[width=0.9\textwidth]{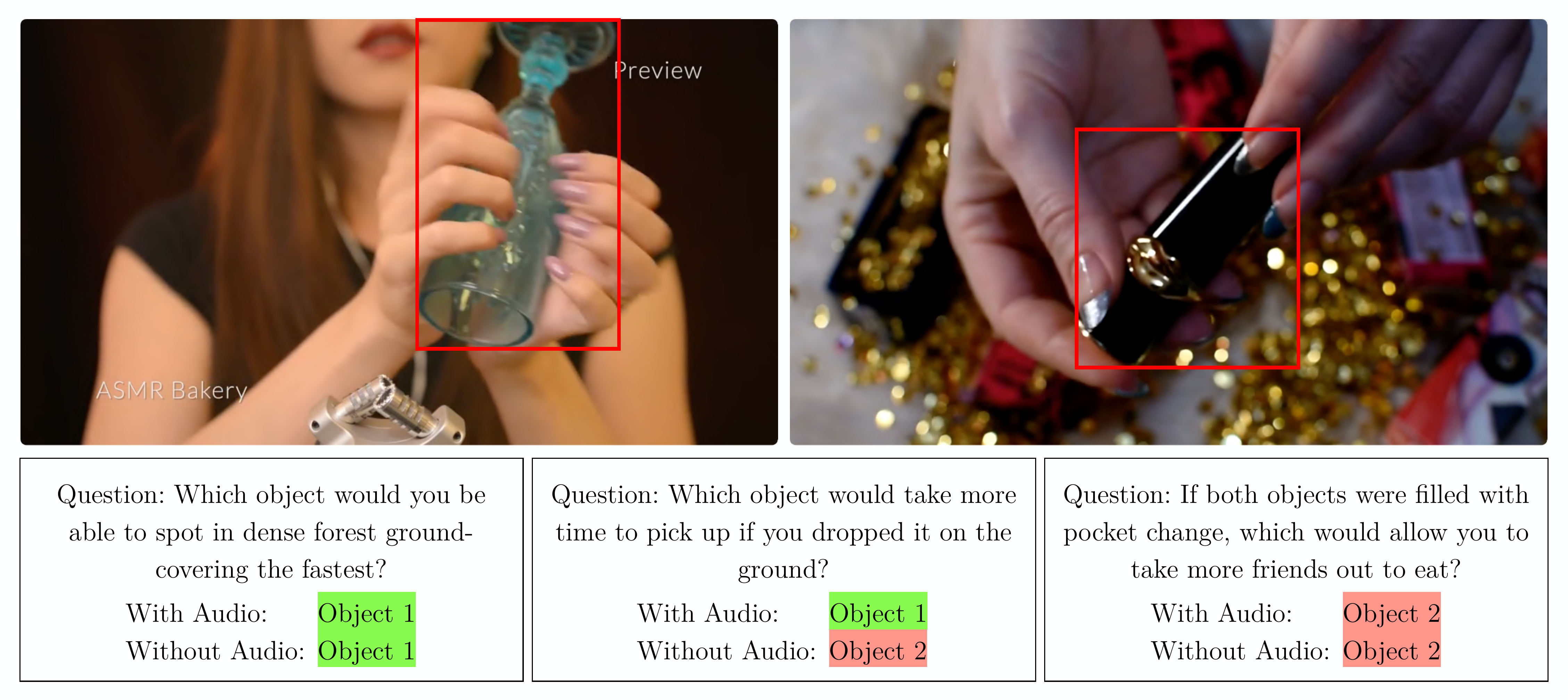}
    \caption{Qualitative results showing predictions from Merlot Reserve models trained with and without audio. In this example, the first object could be mistaken as plastic and the second object could be made of plastic or metal. Thus, the model without audio doesn't realize that the glass object will shatter and takes longer to pick up off the ground. Furthermore, both models fail to answer the third question, which indirectly asks about the size and shape of both objects. This shows that models struggle on questions that are more complex, or require more implicit knowledge.}
    \label{fig:qualitative}
\end{figure}

\subsection{Difficulty of Reasoning} \label{sec:reasoning}

\begin{table}[t!]
\setlength{\tabcolsep}{6pt}
\renewcommand{\arraystretch}{1.0}
\centering
\begin{tabular}{l|c|c:c|c}
\Xhline{3\arrayrulewidth}
\multirow{2}{*}{\textbf{Baseline Model}} &\multirow{2}{*}{Subset}   & \multicolumn{3}{c}{\textbf{Accuracy (\%)}} \\ \cline{3-5}
    &   &  \names-material   & \names & $\Delta$ \\ \hline
    \multirow{2}{*}{Late Fusion~\cite{pandeya2021fusion}} &  Val  & $67.8 \pm 0.8$ & $55.5 \pm 0.3$ & $12.3$ \\
    & Test & $67.4 \pm 1.5$ & $55.0 \pm 1.1$ & $12.4$ \\ \hline
    \multirow{2}{*}{AudioCLIP~\cite{guzhov2021audioclip}} & Val & $81.9 \pm 1.2$ & $61.6 \pm 0.9$ & $18.8$ \\
    & Test & $75.9 \pm 1.1$ & $60.0 \pm 0.9$ & $15.0$ \\
\Xhline{3\arrayrulewidth}
\end{tabular}
\caption{Comparison of \names-material and \names. Despite \names-material being created from relatively noisy labels, we observe that it is a far easier task, with models performing $10-20\%$ better on it than on \names. This suggests that our dataset requires a level of reasoning that goes beyond what is required in classification tasks.} 
\label{table:classification_results}
\end{table}

As seen in Table~\ref{table:classification_results}, the material classification task on our dataset is much easier than our main task, with models achieving $10$-$20\%$ higher accuracy, despite being trained using fewer datapoints (\numprint{11044} vs \numprint{3460}). Since the only other difference between \names\ and \names-material lies in the content of the questions, we believe that this gap in performance is due to the added difficulty of physical commonsense reasoning. The remaining $20$-$30\%$ of misclassified datapoints on \names-material can be attributed to both noisy labels resulting in imperfect training and evaluation, and a true failure in understanding the objects' material makeup.

\subsection{Example Predictions} 

Finally, we analyze some specific examples to see where audio is helpful, and where both models fail. Generally, audio is helpful when models are presented with visually ambiguous or uncommon objects. In these situations, audio is necessary to clarify the physical properties of the objects (e.g., question 2 in Figure~\ref{fig:qualitative}). Furthermore, despite the presence of audio, both models may still fail when asked complex and/or uncommon questions that require the understanding of implicit information (e.g., question 3 in Figure~\ref{fig:qualitative}).

\section{Conclusion}

We introduced \names, a large-scale audiovisual dataset for physical commonsense reasoning. We find that the best models still struggle to (1) fully leverage multimodal information, and (2) develop a strong understanding physical commonsense. 
Through experiments, we evince the multimodal nature of \names\ and its usefulness in benchmarking future work in multimodal commonsense reasoning. We also provide multiple promising directions for bridging the gap between human and AI performance, which we hope provides insight in progressing towards safe and robust multimodal representations of the physical world.


\paragraph{\textbf{Acknowledgements}} This material is based upon work partially supported by the National Science Foundation (Awards \#1722822 and \#1750439) and National Institutes of Health (Awards \#R01MH125740, \#R01MH096951, and \#U01MH116925). Additionally, we would also like to acknowledge NVIDIA's GPU support and Google's TPU support.
Paul Pu Liang is partially supported by a Facebook PhD Fellowship and a Carnegie Mellon University's Center for Machine Learning and Health Fellowship. Ruslan Salakhutdinov is partially supported by NSF IIS1763562 and ONR Grant N000141812861.
Any opinions, findings, conclusions, or recommendations expressed in this material are those of the author(s) and do not necessarily reflect the views of the National Science Foundation, National Institutes of Health, Facebook, Carnegie Mellon University's Center for Machine Learning and Health, or Office of Naval Research, and no official endorsement should be inferred. 
Finally, we are extremely grateful to Martin Ma, Chaitanya Ahuja, Volkan Cirik, Amir Zadeh, Alex Wilf, Victoria Lin, Dong Won Lee, and Torsten W\"{o}rtwein for helpful discussions and feedback on initial versions of this paper.

\clearpage

\bibliographystyle{splncs04}
\bibliography{egbib}

\clearpage

\section*{Appendix}

\appendix
In our work, we presented the creation of \names\ and benchmarked and analyzed the performance of current state-of-the-art models on our dataset. To promote reproducibility and additional understanding of our contributions, we provide details, insights, and possible limitations in our appendix, focusing on the following areas:
\begin{enumerate}
    \item More information about dataset creation (section~\ref{appendix:data}).
    \item Details about model configurations and hyperparameters (section~\ref{appendix:experimental}).
    \item Additional qualitative results from Merlot Reserve and CLIP (section~\ref{appendix:results}).
    \item A datasheet for \names\ (section~\ref{appendix:datasheet}).
\end{enumerate}

\section{Data Collection Details}
\label{appendix:data}

In this section, we elaborate more on all $5$ steps of data collection as outlined in our main paper. Table~\ref{table:pacs_stats} shows the broad statistics of both \names\ and \names-material, such as the number of datapoints in each subset of our dataset. 

\subsection{Video Collection}
\label{appendix:data1}

We split this subsection into two main parts: firstly, we detail how we gathered the initial raw YouTube videos, and secondly, we detail the automatic procedures used to convert the videos into the candidate set of 5-10 second long clips, which we sent to the next step. 

\subsubsection{Creating a Set of Search Queries}

\begin{figure}
    \centering
    \includegraphics[width=\textwidth]{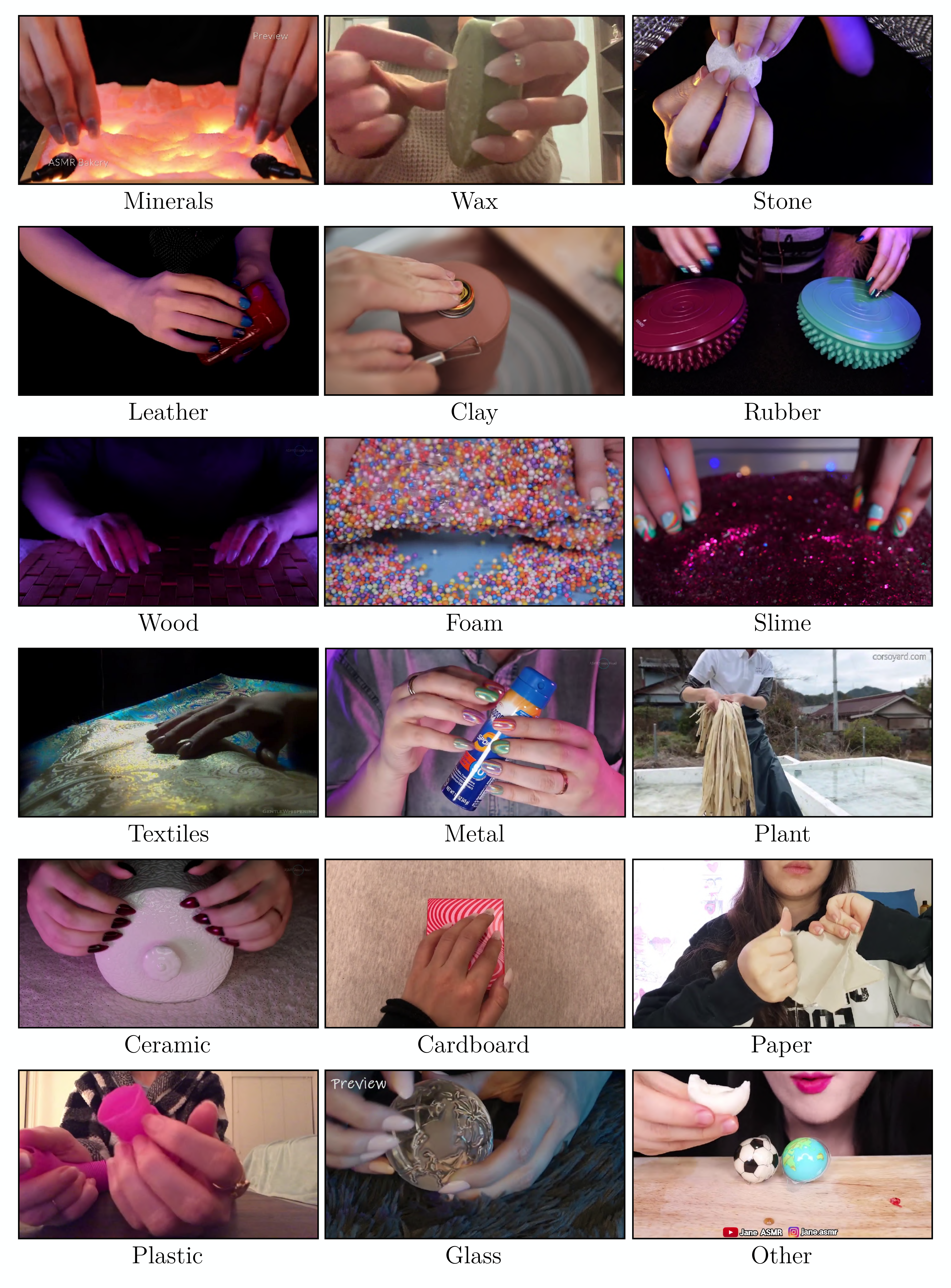}
    \caption{Example objects from each of the most common material types in our dataset. Our dataset contains a diverse set of materials and a number of uncommon or visually ambiguous objects within each material category.}
    \label{fig:exemplars}
\end{figure}

We began with a list of 10 materials, using a standard list of recycling materials~\cite{materiallist} as an initial guide, which was then expanded into a list of 18 materials as we iterated through rounds of video collection, and video clip annotation and filtering. Namely, 8 of the most common materials that were annotated in the ``video clip annotation and filtering'' round (that weren't on the initial list) were added. For each material (e.g., ``paper''), we harvest the top 200 results from YouTube using the search query ``ASMR \{material\} no talking -{}-mukbang'',\footnote{Mukbang videos are videos depicting people eating a large amount of food. We included ``-{}-mukbang'' as part of the query to filter out the high amount of mukbang videos that were initially being scraped, and these videos were intentionally omitted as they were undesirable for our dataset and contained many human faces.} and also enforce a maximum video length of 180 minutes. The final list of materials used are as follows:
\begin{multicols}{3}
\centering
\begin{enumerate}
    \item Wood
    \item Plant
    \item Foam
    \item Paper
    \item Cardboard
    \item Leather
    \item Textiles
    \item Metal
    \item Plastic
    \item Wax
    \item Clay
    \item Rubber
    \item Stone
    \item Concrete
    \item Minerals
    \item Ceramic
    \item Glass
    \item Slime
\end{enumerate}
\end{multicols}

When analyzing our dataset, we noticed that videos scraped from these queries contained a much more diverse set of objects and materials than what was reflected from our list of 18 materials. For example objects from each of the 18 material classes, see Figure~\ref{fig:exemplars}.

\begin{table}[t!]
\setlength{\tabcolsep}{6pt}
\renewcommand{\arraystretch}{1.0}
\centering
\begin{tabular}{c:c|c|c|c}
\Xhline{3\arrayrulewidth}
Dataset & Subset & Datapoints & Withheld Questions & \#Videos \\ \hline
\multirow{3}{*}{\names} & Train & 11044  & 0 & 1224 \\
& Val & 1192 & 146 & 150 \\ 
& Test & 1164 & 119 & 152 \\ \hline
\multirow{3}{*}{\names-material} & Train &  3460 & - & 1224 \\
& Val & 444 & - & 150 \\ 
& Test & 445 & - & 152 \\
\Xhline{3\arrayrulewidth}
\end{tabular}
\vspace{1mm}
\caption{The number of datapoints in each split in \names\ and \names-material. ``Withheld Questions'' refer to the number of questions that only appear in the specific subset of our dataset, to promote model generalization. }
\label{table:pacs_stats}
\end{table}

\subsubsection{Splitting the Videos and Sampling Video Clips}

After we gathered the candidate set of videos, each video was split into segments using a shot boundary detector~\cite{souvcek2020transnet}. Video segments less than 4 seconds long were removed, and the remaining segments were randomly split into 5-10 second long clips. As noted in Figure~\ref{fig:vid_durations} in our main paper, the distribution is not perfectly even, and this is because the videos downloaded were encoded with keyframes that were a set distance apart. We split the videos at keyframes instead of re-encoding the videos, which would be very resource-intensive, but necessary if we wanted clips that were exactly $7.5$ seconds long. 

We then use a simple audio classifier~ which outputs class probabilities for [``talking'', ``silence'', ``music'', ``other'']. We remove any clip that has a probability higher than 0.2 for the ``talking'', ``silence'', and ``music'' classes, or has a probability lower than 0.7 for the ``other'' class. From the remaining set of video clips, we use a set of four rules to avoid sampling consecutive clips from the same video, to avoid having a large number of repeated objects:

\begin{enumerate}
    \item Only one clip can be chosen from each segment (as determined by the shot boundary detector) 
    \item For videos under 10 minutes long, clips that were chosen must be at least 5 minutes apart
    \item For videos under 35 minutes long, clips that were chosen must be at least 7 minutes apart
    \item For all other videos, clips that were chosen must be at least 10 minutes apart
\end{enumerate}

We randomly sampled clips while following these rules, and generated a candidate set of over \numprint{10000} clips. However, it is important to note that only a subset of these clips was used in the next step due to resource limitations. 

\subsection{Video Clip Annotation and Filtering} \label{appendix:data2}

We used \numprint{6700} clips from the candidate set in this stage and ended up keeping \numprint{1526}. In this subsection and beyond, we refer to clips and videos interchangeably. Furthermore, we also refer to clips and objects interchangeably, because we only annotate a single ``object of focus'' from each clip. The guidelines for determining which object is the ``object of focus'' were simple - the clip should show a human interacting with an object, and thus the object of focus was defined as whatever object the human in the video was touching or interacting with.

This ``video clip annotation and filtering'' task contains three successive subtasks, which we detail below.

\begin{figure}[t]
    \centering
  \includegraphics[width=\textwidth]{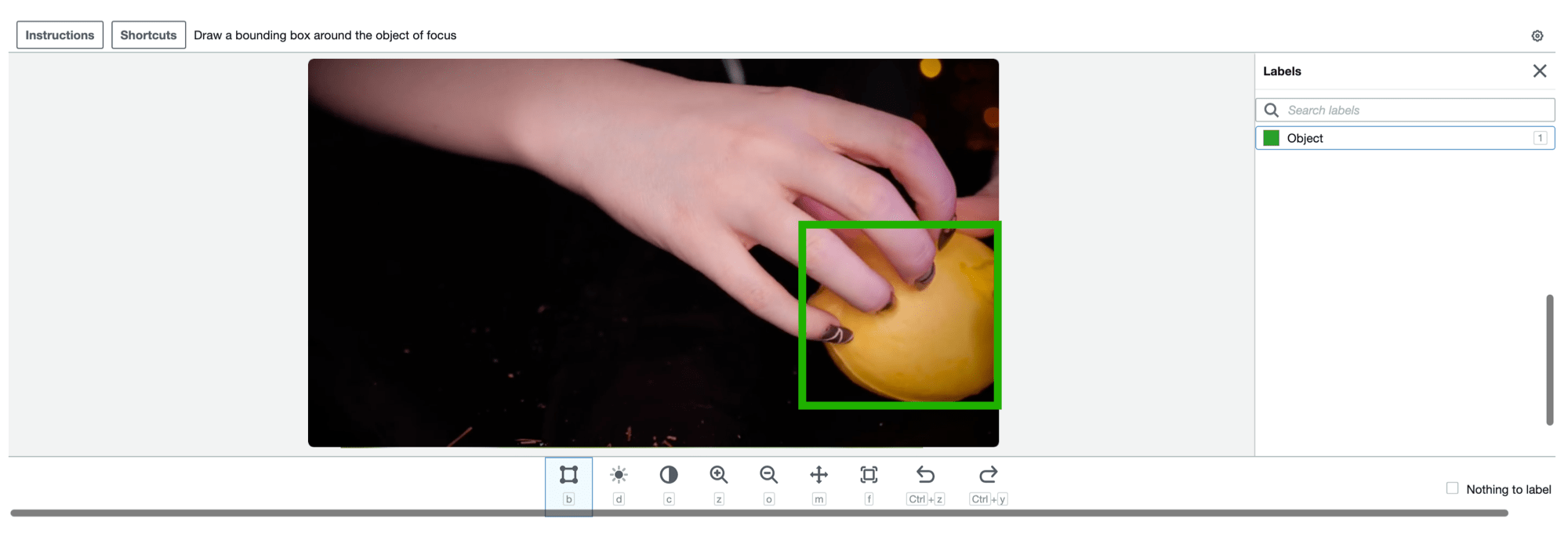}
  \caption{Annotation interface to draw a bounding box around the object of focus.}
  \label{fig:bb_interface}
\end{figure}

\begin{figure}[t]
    \centering
  \includegraphics[width=\textwidth]{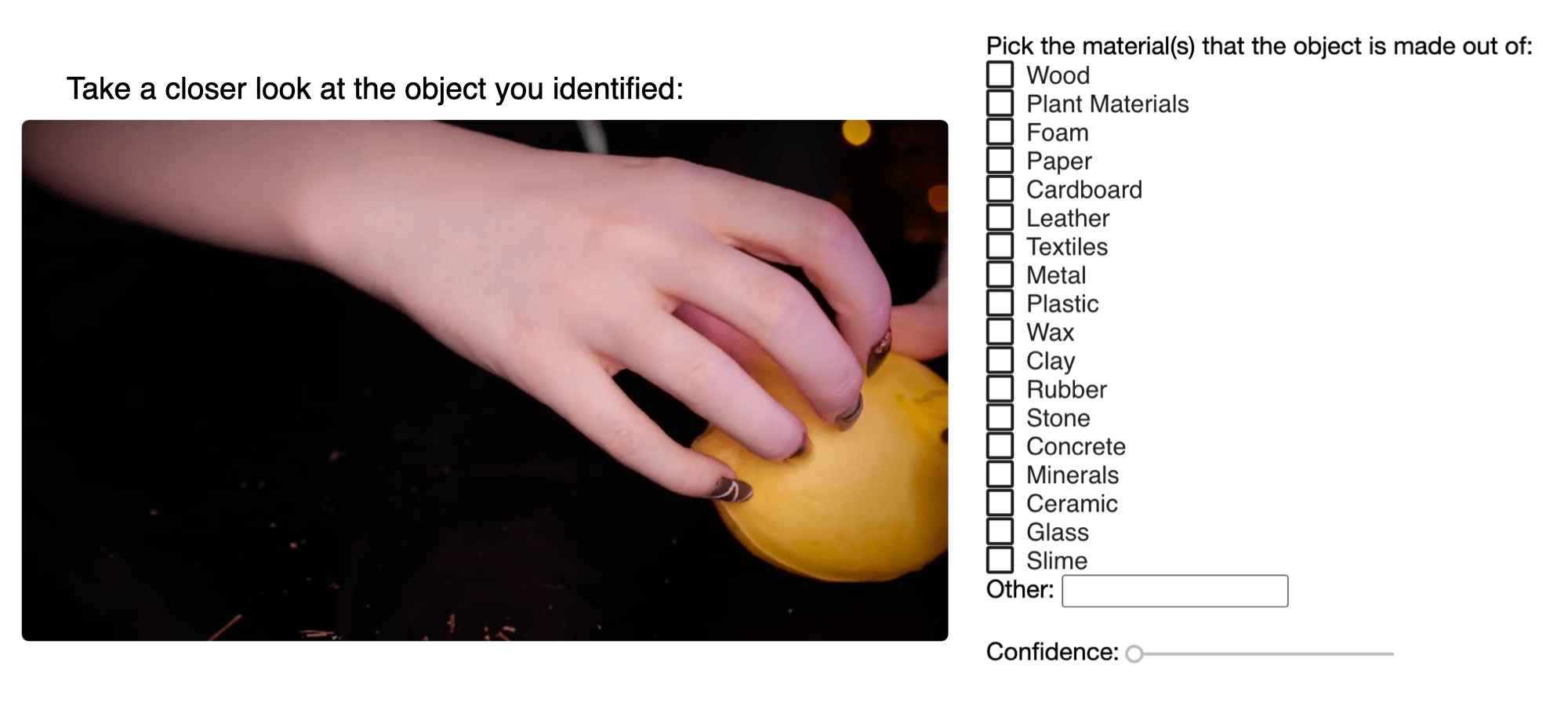}
  \caption{Annotation interface for annotator's first attempt at material classification.}
  \label{fig:guess1_interface}
\end{figure}

\begin{figure}[t]
    \centering
  \includegraphics[width=\textwidth]{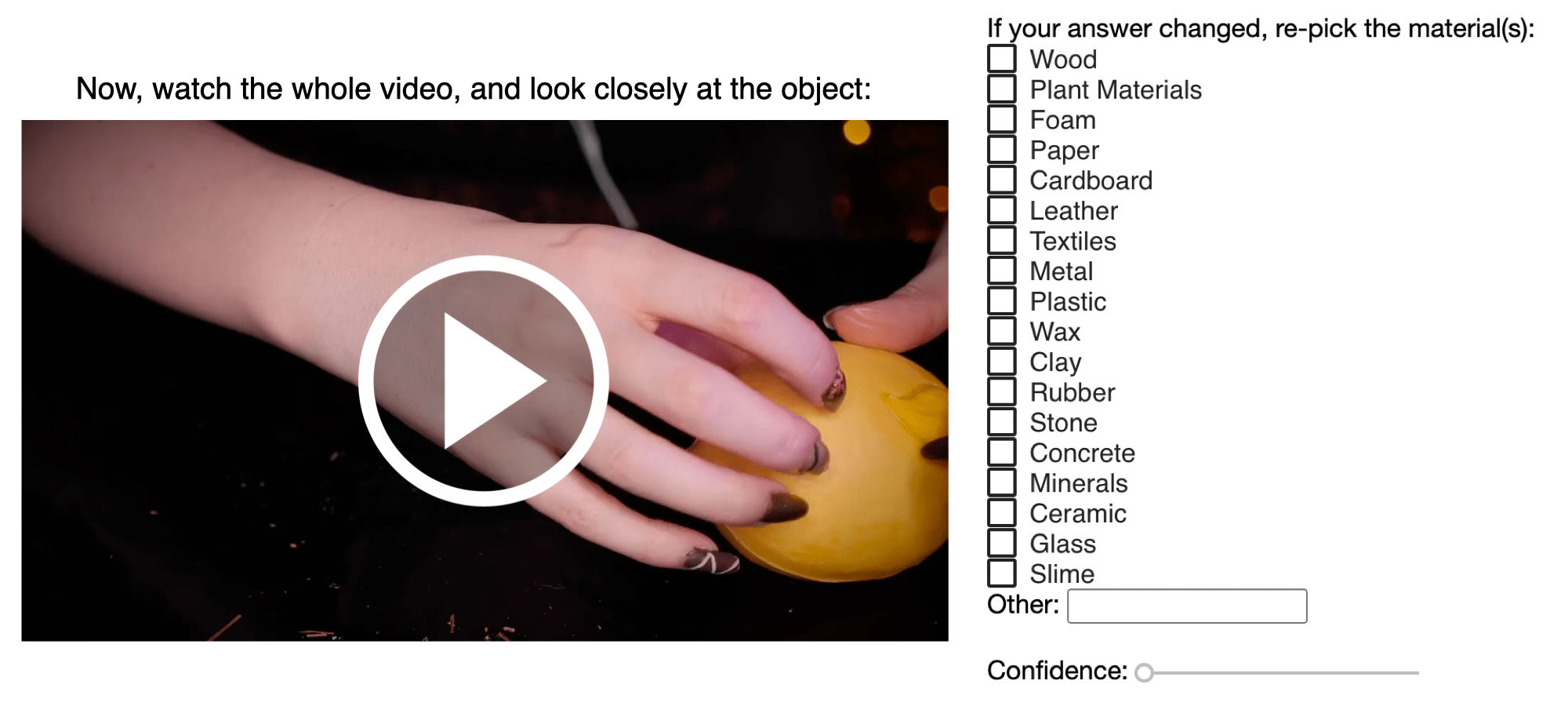}
  \caption{Annotation interface for annotator's second attempt at material classification.}
  \label{fig:guess2_interface}
\end{figure}

\begin{enumerate}
    \item \textbf{Annotating the ``Object of Focus'':} In this subtask, annotators were given the middlemost frame from a video clip and asked to draw a bounding box around what they believed to be the object of focus. Importantly, this was a guess that was sometimes incorrect. If annotators realized that was the case in  the latter steps, they were asked to simply flag the clip with a message (e.g., ``wrong guess of object''), and move on to the next clip. See Figure~\ref{fig:bb_interface} for an example annotation interface of this step.
    \item \textbf{Classifying Materials (First Attempt):} Once annotators had annotated the object of focus, they were then given a list of materials (these are the same 18 materials as shown in section~\ref{appendix:data1}), and alternatively, the option to select ``Other'' and type in a material that was not listed. From this list, they picked the materials they thought the object was made out of. Finally, they were asked to provide a confidence score from 1-5. See Figure~\ref{fig:guess1_interface} for an example annotation interface of this step.
    \item \textbf{Classifying Materials (Second Attempt):} After answering the material classification task for the first time, annotators were then given the whole video and audio to watch, and the chance to redo the task from the previous step. They were also asked to provide a new confidence score from 1-5. See Figure~\ref{fig:guess2_interface} for an example annotation interface of this step.
\end{enumerate}

\subsubsection{Other Annotation Options} Annotators were also given the ability to flag the clip, and all flagged clips were automatically removed from the dataset. The first reason is if they incorrectly guessed the object of focus. The second reason is if the clip was ``bad'', where a bad clip is defined with the following criteria:
\begin{enumerate}
    \item It is impossible to identify an object of focus in the frame given.
    \item The clip dramatically cuts halfway and shows different content.
    \item The audio in the clip contains loud background music or talking.
    \item The clip is almost silent or has no audio at all.
    \item The audio in the clip does not match with the video.
    \item The clip is sped up or slowed down.
    \item The clip is completely off-topic.
    \item The clip is overly blurry or has other weird artifacts.
    \item The clip contains inappropriate or sensitive content.
    \item The clip is completely off-topic.
\end{enumerate}

\subsubsection{Distribution of Materials}
Note that the distribution of materials shown in Figure~\ref{fig:materials} of the main paper is taken from labels in this task. However, we group some materials together to form larger categories, where the materials in each category share roughly similar properties. Specifically, the categories are defined as:
\begin{enumerate}
    \item Natural: materials annotated with ``wood'' or ``plant''
    \item Rubber: materials annotated with ``rubber''
    \item Foam: materials annotated with ``foam''
    \item Paper: materials annotated with ``paper'' or ``cardboard''
    \item Glass: materials annotated with ``glass'' or ``ceramic''
    \item Textiles: Objects made of ``leather'' or ``textiles''
    \item Stone: Objects made of ``stone'', ``concrete'', or ``minerals''
    \item Metal: Objects made of ``metal''
    \item Plastic: Objects made of ``plastic''
    \item Other: Objects made of materials otherwise not listed above
\end{enumerate}

\subsubsection{Picking Clips to Use}
As mentioned previously, not every clip from the candidate set was used in this step. To try and encourage a more balanced set of objects, we biased the selection of clips to send through the filtering stage. Clips from search queries of less commonly appearing materials (e.g., Foam) would be selected more often than clips from search queries of common materials (e.g., Plastic). Specifically, if $(m_1, m_2, \dots, m_{18})$ represent the number of objects of each of the 18 main materials we currently have, and let $n = \sum_i m_i$, the probability that we would pick a clip from the search query $q_i$ would be $\frac{n-2m_i}{n^2 - 2n}$. Doing so had a small effect on the final distribution of objects, but we note that some materials such as plastic still show up much more often because the search queries did not directly correlate with the actual materials present in the videos downloaded. 

\subsubsection{Annotator Management}

Due to the importance of having high-quality objects, we used two in-house annotators in this annotation step. The annotators were trained through multiple rounds. First, they were given specific instructions and were asked to annotate 20 hand-picked videos. Then, the annotations were reviewed by the first author of this paper, and necessary feedback was given. Afterward, annotators were asked to annotate a new set of 20 videos, and they received final pieces of feedback from the first author. After this round, the two annotators were allowed to proceed with annotating all the candidate clips. From then on, the first author would spot-check annotations every 3 days and provide any pieces of feedback necessary, and the annotators were also able to directly communicate with the first author if there were any pressing issues or questions. 

\subsection{Question Creation}\label{appendix:data3}

\begin{figure}[t]
    \centering
  \includegraphics[width=\textwidth]{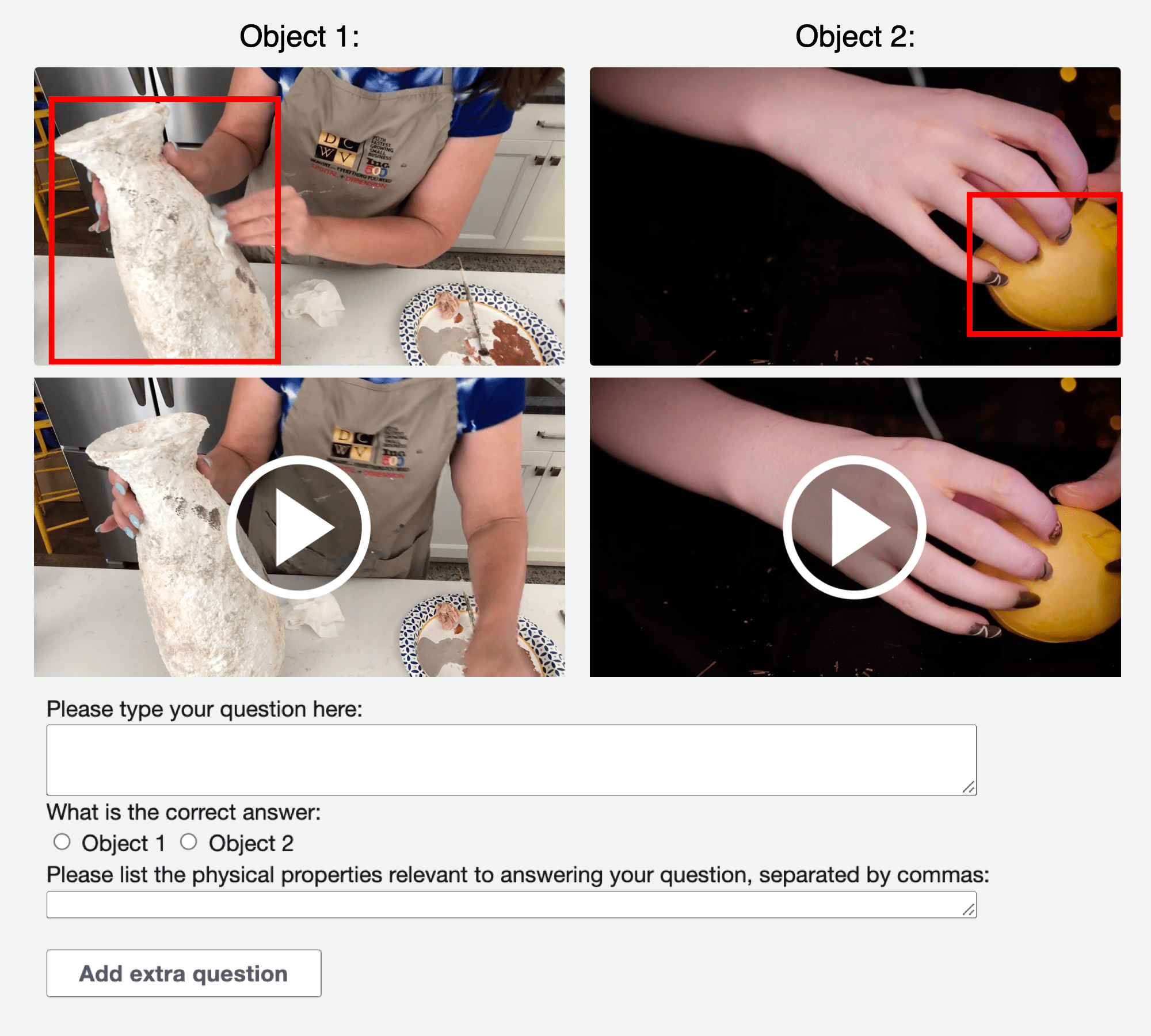}
  \caption{Annotation interface for question creation.}
  \label{fig:qm_interface}
\end{figure}

Out of the 242 pairs selected to be used, 24 were from the validation set and 25 were from the test set. We used Amazon Mechanical Turk (MTurk) as the annotation platform for this step. We divide this subsection into two parts: firstly, we will detail the instructions provided to the annotators, and secondly, we detail the specifics for annotator qualification and management. See Figure~\ref{fig:qm_interface} for the annotation interface used in this step. 

\subsubsection{Annotator Instructions} As mentioned in the main paper, annotators were provided with the two categories of physical commonsense (common real-world knowledge and intuitive physics), along with example questions created from pairs of objects. Alongside this categorization, they were also given a list of heuristics to help evaluate the quality of their question:
\begin{enumerate}
    \item Does the question incorporate interactions between multiple objects? In this case, multiple objects could mean an interaction \textit{between} the two objects we provide, such as ``If I were to stack the two objects, which would logically go on the bottom?''. However, you can also introduce a new object in the question, and ask about an interaction involving the new object, such as ``Which object would make a higher-pitched noise when hit with a metal hammer?''. 
    \item Does the question involve at least one relevant physical property? To determine what a relevant physical property is, imagine which physical property(s) would need to be swapped between the two objects to make the answer to the question change. For example, consider the question ``Which object is more likely to break into multiple pieces if dropped from head height onto the pavement?'', where the two objects are a plastic container and a glass bottle. Then, size is \textbf{not} a relevant property, because even if the two objects swapped sizes, the answer would not change. However, if the two objects swapped strengths, then the answer would likely swap, and thus strength \textbf{is} a relevant property.\footnote{As part of the instructions, we also provide 6 examples and definitions of physical properties (size, shape, weight, hardness, flexibility, and strength) and questions that relate to each of them. Annotators were also encouraged to visit the Wikipedia article detailing physical properties to get a better idea of what physical properties exist~\cite{wikipediaphysicalprops}. Thus, as we see in Figure~\ref{fig:properties} of the main paper, the majority of the relevant physical properties in our questions involved these 6 physical properties, and as such, may be a source of bias or a limitation of our dataset.}
    \item How does the question frame the relevant physical property(s)? Does the question simply restate or provide a reworded definition of the physical property(s), or is there a level of reasoning required to understand what physical property(s) are relevant? For example, questions such as ``Which object is larger?'' or ``Which object is easier to break?'' are too simple, since they present the relevant physical properties in an obvious manner.
    \item Does your question have proper grammar? Also, try and make the question concise and easy to understand.
\end{enumerate}

\subsubsection{Annotator Management} To ensure high-quality questions, annotators were selected using a qualification round. In this round, annotators were given three pairs of videos and had to write one question for each pair. Furthermore, the three questions written were required to incorporate at least three different physical properties in total, to show that the annotators could make a diverse set of questions. The questions were then graded on a scale of 1-4 by the first author, with 3 being good and 4 being exceptional. Annotators with an average score of 3 or greater were qualified for the task. Otherwise, the first author provided personalized feedback to annotators who requested it, and they were then allowed a second attempt. In total, we qualified 23 annotators for this step. 

After the initial qualification, we also continuously monitored the quality of the annotations. Every three days, the first author of this paper would review new annotations, and provide feedback to specific annotators. If a common pattern was observed, aggregate feedback was given to all the annotators. If low quality from a specific annotator was observed through consecutive checks, they were de-qualified. In total, roughly 200 questions were specifically reviewed in this manner. 

In this step, annotators were paid \$0.22 for each question they made and took an average of 1-1.5 minutes to create a question. This resulted in an average wage of \$9-\$13 per hour. 

\subsection{Question Reassignment} \label{appendix:q_reassign}

\begin{figure}[t]
    \centering
  \includegraphics[width=\textwidth]{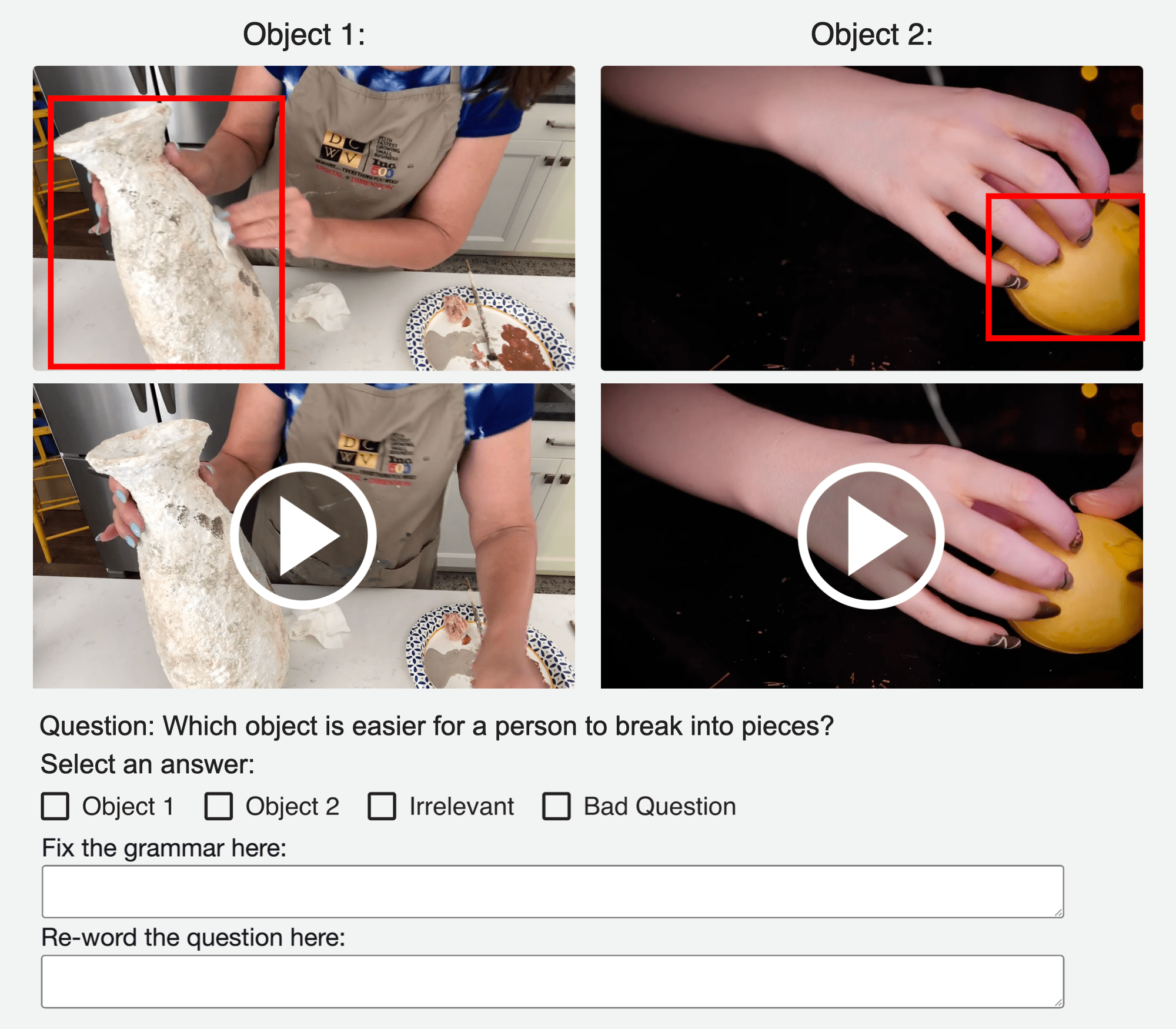}
  \caption{Annotation interface for question reassignment. Note that the last two text boxes were only available for in-house annotators and not MTurk workers.}
  \label{fig:assign_interface}
\end{figure}

For this step, we used two in-house annotators, and also a group of MTurk annotators. The in-house annotators were used to help iterate through versions of this task. Then, they were given 110 instances of this task, with each instance containing 13 questions and a new pair of objects. This way, every question created during the previous phase would appear in a single instance of the new task. In doing so, we add a small subtask specifically for our in-house annotators to quality-check each question. In this subtask, they were given guidelines similar to those outlined in section~\ref{appendix:data3} and asked to flag bad questions. Furthermore, they were asked to fix any small grammar issues.\footnote{We also tested a question augmentation subtask where the in-house annotators were able to edit the question to fit the new object pair better. This was scrapped before the task was given to MTurk workers due to the difficulty in training MTurk workers on this subtask, and to maintain the intention of creating uncommon question-object matchings. However, the ~300 datapoints created from this subtask with our in-house annotators were kept in the dataset.}

Once the 110 instances were annotated by our in-house annotators, and bad questions were removed, the remaining set of \numprint{1377} questions was then distributed to all the remaining \numprint{2047} object pairs that were not used in the Question Creation step. Each object pair was randomly given a list of 13 questions. To promote model generalization, we also ensured that questions made for object pairs in the validation and test sets were not distributed outside the corresponding subset. Specifically, given three sets of questions $q_{train}, q_{test}, q_{val}$ created in the previous step, pairs in the validation set were randomly given questions from $q_{train} \cup q_{val}$, pairs in the test set were randomly given questions from $q_{train} \cup q_{test}$, but pairs in the train set were only given questions in $q_{train}$. Then, the MTurk annotators were given the object pairs and list of questions, and asked to either answer the question, or mark the question as irrelevant. Figure~\ref{fig:assign_interface} shows the general interface used in this step. However, the last two text-box options were only available when we gave the task to our in-house annotators. 

\subsubsection{Annotator Instructions} We provided annotators with an explanation for when to mark a question-object matching as ``Irrelevant''. Specifically, matchings that were \textit{irrelevant} constitute:
\begin{enumerate}
    \item Questions that ask for a hypothetical scenario, but one of the objects cannot be placed in such a scenario. For example, if one of the objects is a laptop, and the question is ``If I placed both objects in a sock and rapidly swung the sock around, which would be more likely to hurt a bystander?'', the matching is \textit{irrelevant} because a computer cannot fit in a sock, so the hypothetical situation is invalid.
    \item Objects that are equally bad or equally good answers to the question \textbf{to the point where it is virtually impossible to differentiate}. For example, if the two objects are a Rubik's cube and a roll of scotch tape, and the question was ``In a pinch, which object would be a more suitable replacement for a pillow?'', the matching would be irrelevant because both objects are very bad answers. However, annotators were specifically warned about the difference between both objects being bad answers, versus the objects being \textit{uncommon} answers, that could still possibly work.
\end{enumerate}

\subsubsection{Annotator Management}

To ensure annotation quality, two rounds of qualification were used. Firstly, annotators were asked to annotate two sets of handpicked objects and 13 questions. In this stage, they were required to have an accuracy of above $80\%$. In the second round, they were given another pair of objects and 10 questions, but were also required to write a sentence or two of justification for each answer. Then, the first author reviewed each explanation, and if the explanations were deemed as reasonable, the annotators were given the qualification. We qualified 27 annotators for this step. Once qualified, we used a simple attention check to monitor annotators. For every 1 in 10 pairs of objects, one of the questions would be replaced with an attention check, such as: ``Please select both `Object 1' and `Bad Question' as your answer''. If workers failed an attention check, they were given a warning, and if they failed another, they were de-qualified. We paid $\$0.32$ for each set of 13 questions, and annotators took roughly 1-2 minutes per set, resulting in a pay rate of \$9.6-\$19.2/hr. 

\subsection{Quality Checking}

\begin{figure}[t]
    \centering
  \includegraphics[width=\textwidth]{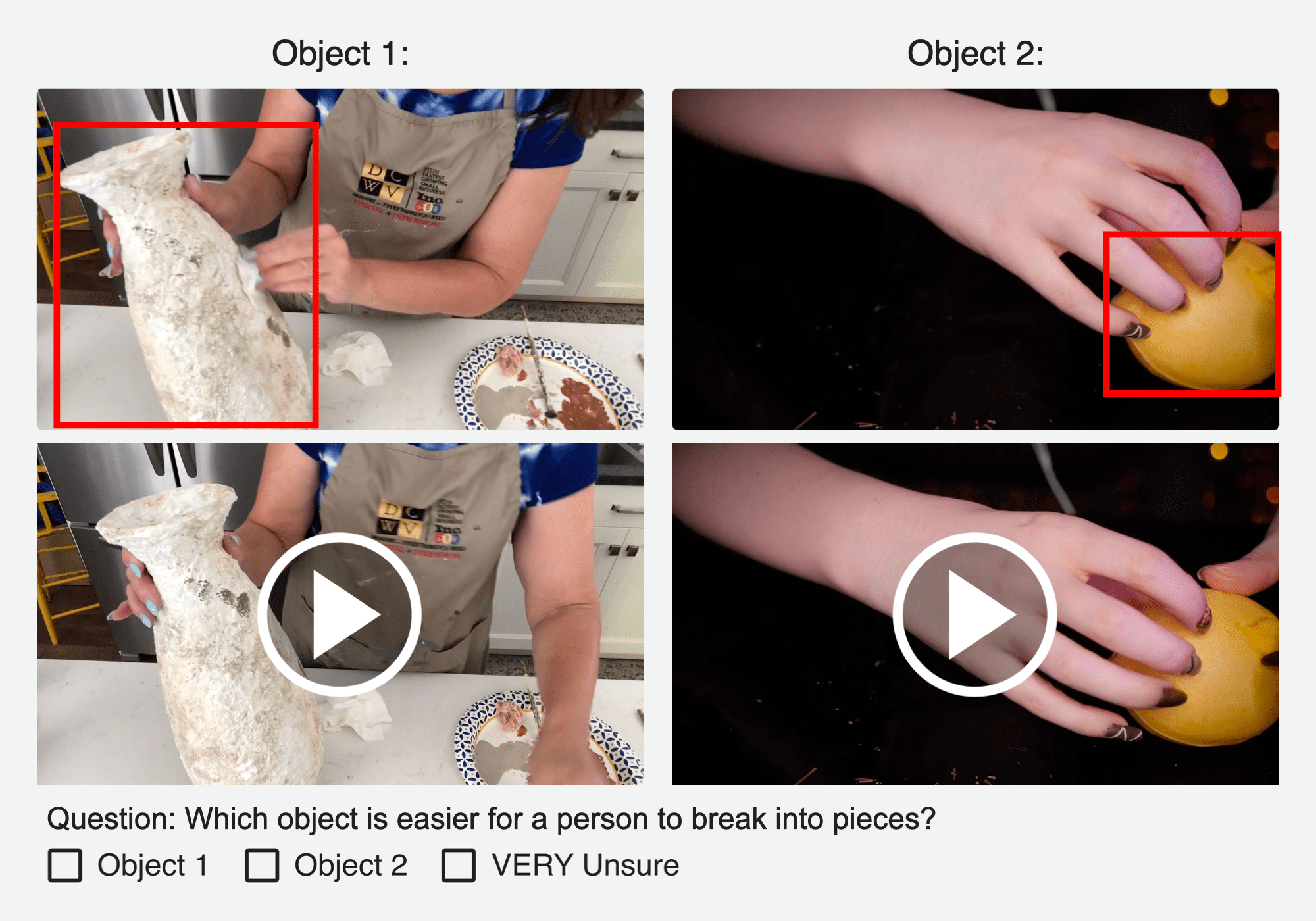}
  \caption{Annotation interface for question checking.}
  \label{fig:check_interface}
\end{figure}

From the previous annotation steps, we were able to create a set of candidate datapoints. To ensure the quality of the datapoints, we created a final question-answering task that we gave to 2 extra annotators and removed any datapoint without unanimous agreement. Annotators were given an object pair, and all the candidate datapoints relevant to the object pair. They were also given the ability to flag a datapoint if there were any quality issues, or if it was unanswerable, but were instructed to use the flag very sparingly. Annotators in this step were a combined set of the annotators from the previous two MTurk steps and were paid \$0.22 for each object pair they checked, resulting in an average pay rate of roughly \$12 an hour. Figure~\ref{fig:check_interface} shows the annotation interface used in this step.

\section{Experimental Setup Details}\label{appendix:experimental}

In this section, we provide details about our experiments, including model backbones, hyperparameters, data augmentation, and other procedures. First, we will explain some vocabulary that will be consistently used throughout the section, and then provide details about each specific experiment. 

\subsection{Relevant Vocabulary}

\subsubsection{Bounding-Box-Centered Images} In many of our experiments, we use an image backbone, which takes a single image input. For these single-image inputs, we commonly use \textbf{Bounding-Box-Centered Images}. These images are sourced from the middlemost frame of the video, which were each annotated with a bounding-box (representing the object). We crop each image such that center of the crop is matched with the center of the bounding-box. For a bounding-box of size $(w_b, h_b)$, the corresponding size of the crop is a square with side-length $1.25 \cdot max(w_b, h_b)$.\footnote{If the boundary of the crop extends beyond the image, we resize the crop accordingly. Thus, some crops will not be a perfect square (e.g., if the bounding box covers the entire image).}

\subsubsection{Drawn-on Bounding Box} We follow past work in ``drawing-on'' bounding boxes directly onto the image as a simple way to represent the object (without needing an additional bounding-box input)~\cite{zellers2022merlotreserve,zellers2021merlot}. Thus, we ``draw'' a bright-red bounding box with a width of 2 pixels onto the middlemost frame of each video, and use this as the image input in various experiments. During finetuning, the model(s) learn the connection between the bounding box and the object that the question is asking about. 

\subsection{Human Performance}

We tested human performance using 243 random datapoints from the \names\ validation and test sets. The 243 datapoints were divided into two subsets $\mathbf{A}$ and $\mathbf{B}$, and we made sure that no videos used in subset $\mathbf{A}$ appeared in $\mathbf{B}$, and vice-versa. The 10 annotators chosen to test human performance were annotators qualified for the Question Reassignment state (section~\ref{appendix:q_reassign}), but did not annotate any of the videos appearing in this subset. Thus, these annotators did not require additional training and were simply asked to answer each question to the best of their ability. 

Similarly to the 243 datapoints, we divide the 10 annotators into two sets of 5. Then, the first 5 annotators were given subset $\mathbf{A}$ \textit{with audio}, and subset $\mathbf{B}$ \textit{without audio}, and the other 5 annotators were given the opposite. Thus, for each datapoint we received 10 annotations, 5 with audio and 5 without. 

\subsection{Late Fusion Model~\cite{pandeya2021fusion}}

In this subsection, we detail the data augmentation methods, model hyperparameters, and other implementation details related to the general late fusion method. Then, we go into specific details for each configuration of the late fusion model, such as the four bias-testing models.

\subsubsection{Specific Backbones Used} 
\begin{enumerate}
    \item \textbf{Image model:} We used the ViT/B-16 model pre-trained on ImageNet-21k from HuggingFace~\cite{wolf2020huggingface}. The model can be downloaded from \href{https://huggingface.co/google/vit-base-patch16-224-in21k}{\email{here}}.
    \item \textbf{Audio model:} We use the AST (Audio Spectogram Transformer)~\cite{gong2021ast} model with a time and frequency stride of 10 and weight averaging, pre-trained on the full AudioSet~\cite{gemmeke2017audioset} as the audio model. The model can be downloaded from \href{https://github.com/YuanGongND/ast}{\email{here}}. 
    \item \textbf{Video model:} We use TDN (Temporal Difference Network)~\cite{wang2021tdn} pre-trained on Something-Something-V2 with a ResNet101 backbone and 8 frame inputs as the video model. The model can be downloaded from \href{https://github.com/MCG-NJU/TDN}{\email{here}}. 
    \item \textbf{Text model:} We use a pre-trained DeBERTa-V3-Large model~\cite{he2021debertav3} from HuggingFace~\cite{wolf2020huggingface}. The model can be downloaded from \href{https://huggingface.co/microsoft/deberta-v3-large}{\email{here}}. Since we do not do any text augmentations, we also pre-extract the text embeddings from the <CLS> token of the output layer (pre-pooler) of the text model to save time during training. 
\end{enumerate}

\subsubsection{Hyperparameter Details} For each late fusion model, we used a learning rate of \numprint{5e-4}, a weight decay of \numprint{5e-5}, and a batch size of 64. This value was determined by training the Fusion model using all four modalities, and using a simple grid search to compare the validation accuracy achieved with learning rates of \numprint{5e-3}, \numprint{5e-4}, \numprint{5e-5}, and \numprint{5e-6}, where weight decay was set to $\frac{1}{10}$ of the learning rate. We trained each configuration for 40 epochs, and decrease the learning rate by a factor of 10 after 20 and 30 epochs. When finetuning, we freeze all backbone layers, and thus the only trainable layers are in the MLPs we use to fuse multimodal information. 

\subsubsection{Data Augmentation}
\begin{itemize}
    \item \textbf{Image Augmentations:} We use bounding-box centered images, each containing a drawn-on bounding box. Images were resized with the short side between $[224, 264)$. Next, we randomly adjust the brightness, contrast, and saturation between a range of $(0.9, 1.1)$, and hue between a range of $[-0.05, 0.05]$, where these parameters follow the ColorJitter function in PyTorch~\cite{pytorch2019}. Finally, we randomly flip the image horizontally with a probability of 0.5, and normalize the image with the values $\mu = (0.5, 0.5, 0.5)$, $\sigma = (0.5,0.5,0.5)$.
    \item \textbf{Video Augmentations:} We follow the video augmentation steps used to pre-train TDN on the Something-Something V2 dataset~\cite{wang2021tdn}. We begin with video frames of size $252 \times 252$, and randomly pick 8 evenly spaced frames. Then, we crop the same $224 \times 224$ from each image, and randomly flip the images horizontally with a probability of 0.5. 
    \item \textbf{Audio Augmentations:} We follow the audio augmentation steps used to pre-train AST on AudioSet, using frequency and time masking~\cite{gong2021ast}. We use 128 mel bins, with a target length of 1024. Then, we mask a band of size 48 in the frequency domain and a band of size 144 in the time domain. Finally, we normalize the spectrogram such that $\textsc{spec} = (\textsc{spec}  + 4.26) / (4.57 * 2)$, and add random noise.
    \item \textbf{Text Augmentations:} As mentioned previously, we pre-compute text embeddings, and thus do not use any text augmentations.
    \item \textbf{Other Augmentations:} Since we treat the task as a binary classification task, we flip the order of the two objects in the input tuple with a 50\% probability, and flip the binary label accordingly. 
\end{itemize}

\subsubsection{Configuration Details}

We denote the image, video, audio, and question embeddings for object $n$ as $(\boldsymbol{e}_i^{(n)},\boldsymbol{e}_v^{(n)},\boldsymbol{e}_a^{(n)},\boldsymbol{e}_q^{(n)})$. 

\begin{enumerate}
    \item \textbf{I + A + V:} For this configuration, we generate image, audio, and video embeddings for both objects. Then, we separately concatenate the unimodal embeddings for both objects, and use an MLP to generate two object embeddings $\boldsymbol{e}_o^{(1)}, \boldsymbol{e}_o^{(2)}$. Finally, we concatenate the two object embeddings and use an MLP to generate a binary classification output. 
    \item \textbf{Q + A:} For this configuration, we generate a question embedding, and audio embeddings for both objects. Then, we separately concatenate both audio embeddings with the question embedding, and use an MLP to generate two question-object embeddings $\boldsymbol{e}_{qo}^{(1)}, \boldsymbol{e}_{qo}^{(2)}$. Finally, we concatenate the two question-object embeddings and use an MLP to generate a binary classification output.  
    \item \textbf{Q + I:} For this configuration, we generate a question embedding, and image embeddings for both objects. Then, we separately concatenate both image embeddings with the question embedding, and use an MLP to generate two question-object embeddings $\boldsymbol{e}_{qo}^{(1)}, \boldsymbol{e}_{qo}^{(2)}$. Finally, we concatenate the two question-object embeddings and use an MLP to generate a binary classification output.  
    \item \textbf{Q + V:} For this configuration, we generate a question embedding, and video embeddings for both objects. Then, we separately concatenate both video embeddings with the question embedding, and use an MLP to generate two question-object embeddings $\boldsymbol{e}_{qo}^{(1)}, \boldsymbol{e}_{qo}^{(2)}$. Finally, we concatenate the two question-object embeddings and use an MLP to generate a binary classification output.  
    \item \textbf{Q + I + V:} For this configuration, we generate a question embedding, and image and video embeddings for both objects. Then, we separately concatenate the image and video embeddings for both objects and use an MLP to generate two object embeddings $\boldsymbol{e}_o^{(1)}, \boldsymbol{e}_o^{(2)}$. Next, we concatenate each object embedding with the question embedding and use an MLP to create two question-object embeddings $\boldsymbol{e}_{qo}^{(1)}, \boldsymbol{e}_{qo}^{(2)}$. Finally, we concatenate the two question-object embeddings and use an MLP to generate a binary classification output. 
    \item \textbf{Q + I + V: + A} For this configuration, we generate a question embedding, and image, audio, and video embeddings for both objects. Then, we separately concatenate the image, audio, and video embeddings for both objects and use an MLP to generate two object embeddings $\boldsymbol{e}_o^{(1)}, \boldsymbol{e}_o^{(2)}$. Next, we concatenate each object embedding with the question embedding and use an MLP to create two question-object embeddings $\boldsymbol{e}_{qo}^{(1)}, \boldsymbol{e}_{qo}^{(2)}$. Finally, we concatenate the two question-object embeddings and use an MLP to generate a binary classification output. 
\end{enumerate}

\subsection{CLIP~\cite{radford2021clip}}

For the CLIP model, our input datapoints are a tuple $\boldsymbol{d} = (\boldsymbol{o}^{(1)}, \boldsymbol{o}^{(2)}, \boldsymbol{q}, \boldsymbol{l})$, where $\boldsymbol{o}^{(n)} = \boldsymbol{i}^{(n)}$, since CLIP only takes single image inputs. As mentioned in the main paper, we generate three vector embeddings $\boldsymbol{e}^{(1)}$, $\boldsymbol{e}^{(2)}$, $\boldsymbol{e}^{(q)}$, representing the embeddigns for the two images, and the question embedding. Then, we compute two cosine similarity values $\textsc{sim} (\boldsymbol{e}^{(1)}, \boldsymbol{e}_{q})$ and $\textsc{sim} (\boldsymbol{e}^{(2)}, \boldsymbol{e}_{q})$, where the predicted object is the object with a higher similarity.

In our experiments with CLIP, we use the provided ViT/B-16 model as the image backbone, and the provided transformer model for the text backbone. During finetuning, we freeze all layers except for a single text projection layer. Additional tests were conducted to determine whether unfreezing additional layers would benefit performance (e.g., unfreezing later transformer layers, or the image projection layer) but these did not improve results.

\subsubsection{Loss Function} When finetuning CLIP, we took inspiration from CLIP's contrastive pre-training~\cite{radford2021clip}. Among the two object embeddings $\boldsymbol{e}^{(1)}$ and $\boldsymbol{e}^{(2)}$, one of them will be the correct embedding (let's call it $\boldsymbol{e}^{(+)}$), and the other will be the incorrect embedding ($\boldsymbol{e}^{(-)}$). The training goal for our task is to maximize the cosine similarity between $\boldsymbol{e}^{(+)}$ and $\boldsymbol{e}^{(q)}$, and at the same time, minimize the cosine similarity between $\boldsymbol{e}^{(-)}$ and $\boldsymbol{e}^{(q)}$. Our loss function can thus be defined as:

\begin{equation}
    \mathbf{L}(\boldsymbol{e}^{(+)}, \boldsymbol{e}^{(-)},\boldsymbol{e}^{(q)}) = \text{max}\{\textsc{sim}(\boldsymbol{e}^{(-)}, \boldsymbol{e}^{(q)}) - \textsc{sim}(\boldsymbol{e}^{(+)}, \boldsymbol{e}^{(q)}) + m, 0\}
\end{equation}
where $m$ is a tunable hyperparameter. In our implementation, we use PyTorch's TripletMarginWithDistanceLoss, where we define the distance function to be the negative of the cosine similarity function, and let $m=1.0$.\footnote{We also tested the concatenation of CLIP embeddings to create a classification task (similar to the Late Fusion models), but the results when doing this were poor.}

\subsubsection{Hyperparameter Details} We do a simple grid search, comparing the validation accuracy achieved on the model using learning rates of \numprint{1e-3}, \numprint{1e-4}, and \numprint{1e-5}, with the weight decay being $\frac{1}{10}$ of the learning rate. The learning rate of \numprint{1e-4} with weight decay of \numprint{1e-5} was found to be the best, though the difference between the values tested was not very large. We finetuned CLIP for 40 epochs with a batch size of 64, and decrease the learning rate by a factor of 10 after epochs 20 and 30. 

\subsubsection{Data Augmentation}

We use Bounding-Box Centered Crop images, but we \textit{do not} use a drawn-on bounding box, as results did not improve with the addition of the box. Images were resized with the short side between $[224, 264)$ with a probability of 0.9, and otherwise resized to a shorter edge of $224$. Then a random square crop of size 224 is taken from the resized image. Next, we randomly adjust the brightness, contrast, and saturation between a range of $(0.9, 1.1)$, and hue between a range of $[-0.05, 0.05]$, with these parameters following the ColorJitter function in PyTorch. Finally, we randomly flip the image horizontally with a probability of 0.5, and normalize the image with the values $\mu = (0.48145466, 0.4578275, 0.40821073)$, $\sigma = (0.26862954, 0.26130258, 0.27577711)$.

Similar to the late fusion model, we do not use any text augmentations. Furthermore, we do not randomly swap the order of the input objects, as the order of the input objects does not matter when using CLIP. 

\subsection{AudioCLIP~\cite{guzhov2021audioclip}}

For the AudioCLIP model, our input datapoints are a tuple $\boldsymbol{d} = (\boldsymbol{o}^{(1)}, \boldsymbol{o}^{(2)}, \boldsymbol{q}, \boldsymbol{l})$, where $\boldsymbol{o}^{(n)} = (\boldsymbol{i}^{(n)}, \boldsymbol{a}^{(n)})$. We generate five vector embeddings $\boldsymbol{e}_a^{(1)}$, $\boldsymbol{e}_i^{(1)}$, $\boldsymbol{e}_a^{(2)}$, $\boldsymbol{e}_i^{(2)}$, and $\boldsymbol{e}^{(q)}$, representing the embeddings for the two images and audio segments, and the question embedding. To keep the comparison with CLIP as fair as possible, we use the same problem formulation, where the goal is to maximize the similarity between the correct object and the question, and minimize the similarity between the incorrect object and the question. To account for the multimodal inputs, we add an intermediate step by concatenating the image and audio embeddings for each object, and use a linear layer to generate a final object embedding. Then, these new object embeddings are compared with the question embedding to determine the correct answer. 

In our experiments with AudioCLIP, we use the same pre-trained image and text backbone models as with CLIP (with the same pre-trained weights) and use the provided ESResNe(X)t-fbsp model~\cite{guzhov2021esresnextfbsp} as the audio backbone. To keep things as fair as possible with CLIP, we freeze all layers except the text projection layer, and the linear layer used to re-project the concatenated image-audio embedding.  

\subsubsection{Hyperparameter Details} We use the same hyperparameters as in CLIP finetuning, with a learning rate of \numprint{1e-4} and a weight decay of \numprint{1e-5}. We trained AudioCLIP for 40 epochs, and decrease the learning rate by a factor of 10 after epochs 20 and 30. 

\subsubsection{Data Augmentation} We use the exact same image augmentation methods as CLIP, but we also utilize various audio augmentations, largely following the augmentations used to pre-train AudioCLIP on AudioSet~\cite{guzhov2021audioclip}:

\begin{enumerate}
    \item \textbf{Time Inversion:} We randomly invert the audio track along its time axis with a probability of 0.5.
    \item \textbf{Random Crop and Padding:} If the track is longer than 5 seconds, we randomly crop 5 seconds of audio. If the track is shorter than 5 seconds, we randomly add silence to the start and end of the track until it is 5 seconds long.
    \item \textbf{Random Noise:} We add random Gaussian white noise to the track with a probability of 0.8. 
\end{enumerate}

\subsection{UNITER~\cite{chen2020uniter}}

Since the NLVR2 dataset~\cite{suhr2019nlvr2} has a similar formulation to \names, we largely follow the procedures used to finetune UNITER on NLVR2. Since UNITER takes a single image input, we use the middlemost frame of the videos and use the same bottom-up attention model~\cite{anderson2017bottomup} to extract image features for each detected object region. This provides an input sequence of image features $\langle v_0, v_1, \dots \rangle$. However, since our dataset contains a predefined bounding-box, we also extract the pooled ROI features $v_{obj}$ from the region specifically defined by the bounding box, and insert it at the start of the input sequence: $\langle v_{obj}, v_0, v_1, \dots \rangle$. Thus, the object will always be at the start of the sequence and thus there is  zero ambiguity. 

Though UNITER has three setups available (\textit{Pair}, \textit{Pair-biattn}, \textit{Triplet})~\cite{chen2020uniter}, we mainly used the \textit{Pair} setup in our experiments, and also used the Large backbone. In the \textit{Pair} setup,  one input datapoint is treated as two
text-image pairs by repeating the text. Then, the two [CLS] outputs from UNITER are depth concatenated as the joint embedding for the example, and an MLP is used to generate a classification output. We also tested the other two setups but found that results did not improve.\footnote{Consistent with NLVR2, the \textit{Triplet} setup yielded lower results. We suspect that the \textit{Pair-biattn} setup did not improve results due to  overfitting in the bi-attention module. On NLVR2, there is a larger emphasis on comparing the two input images, and while this also exists on \names\ (e.g., \textit{Which object could fit in the other?}), sharing information may not be as essential to succeed on our dataset as on NLVR2.}

\subsubsection{Hyperparameter Details} We used a learning rate of \numprint{1e-5}, a weight decay of $0.01$, and a batch size of 32. We used a simple grid search to compare learning rates of \numprint{1e-4}, \numprint{1e-5}, and \numprint{1e-6}, and found \numprint{1e-5} to be the best. The weight decay was kept constant at 0.01 in this grid search. We trained UNITER for \numprint{10000} iterations using the AdamW optimizer, with a linear warmup for the first \numprint{1000} iterations. All other hyperparameters and model settings not mentioned were kept the same as what was used to train UNITER on NLVR2. 

\subsubsection{Data Augmentation}

We employ two data augmentations that are not present in finetuning UNITER on NLVR2. Unlike NLVR2, there is no ``left/right'' relationship between the input images in our dataset, so we randomly flip the order of the input images during training, and flip the label accordingly. Additionally, given a sequence of input image features of length $n$, we randomly choose to keep the first $m \in [1, n]$ features and remove the rest during training. We compared values of $i \in \{1, n//4 + 1, n//2 + 1, 3n//4 + 1,  n\}$ to decide the range of $m \in [i, n]$, but found that $i = 1$ worked the best, as it provides the most variation in training data. 

\subsection{Merlot Reserve~\cite{zellers2022merlotreserve}}

When finetuning the Merlot Reserve model, we follow the treatment of image inputs used to finetuning Merlot Reserve on the VCR dataset, and process video and audio similarly to that used when finetuning Merlot Reserve on the TVQA dataset. We also follow the same method of treating multiple-choice answers (in our case, two objects) done in both TVQA and VCR, where we generate separate scores for each individual question and then concatenate them as a classification output. While this formulation massively suffers from information loss (as the two objects are never compared to each other), in our testing, results were better when keeping the answers separate, rather than using a concatenation of embeddings as done when finetuning UNITER.

We trained four versions of Merlot Reserve, using both the Base and Large backbones, and training versions with and without audio. Below, we detail specific hyperparameters and data augmentation methods used to train Merlot Reserve. All hyperparameter searches were conducted using the models \textit{without} audio input, and we then used the same hyperparameters for the model \textit{with} audio. 

\subsubsection{Hyperparameter Details} Since we finetuned versions of Merlot Reserve using both the Base and Large backbones, we detail the hyperparameters used for both. Any hyperparameter not mentioned below was kept the same as what was used to finetune Merlot Reserve on the TVQA dataset. 

\begin{itemize}
    \item \textbf{Base:} We compared learning rates of \numprint{5e-5}, \numprint{5e-6}, and \numprint{5e-7} with no weight decay. We found \numprint{5e-6} to be the best performing learning rate. Then, we compared weight decays of \numprint{1e-5}, \numprint{1e-6}, and \numprint{1e-7} with a learning rate of \numprint{5e-6}, and found \numprint{1e-7} to be the best. We trained each base model for 40 epochs using the AdamW optimizer with a batch size of 64. We use a linear warmup for the first epoch, followed by a constant linear decay to 0. 
    \item \textbf{Large:} We compared learning rates of \numprint{5e-5}, \numprint{5e-6}, and \numprint{5e-7} with no weight decay. We found \numprint{5e-6} to be the best performing learning rate. Then, we compared weight decays of \numprint{1e-5}, \numprint{1e-6}, and \numprint{1e-7} with a learning rate of \numprint{5e-6}, and found \numprint{1e-6} to be the best. We trained each base model for 40 epochs using the AdamW optimizer with a batch size of 16. We use a linear warmup for the first epoch, followed by a constant linear decay to 0. 
\end{itemize}

\subsubsection{Data Pre-Processing and Augmentation} We define our input sequence with 5 segments. The first segment of our input sequence contains the image (bounding-box) input and the question (this loosely follows the method of including metadata and the question in the first segment when finetuning on TVQA~\cite{lei2018tvqa}). Then, the next four segments contain a video frame and the corresponding audio information (for the models without audio input, the segments only contain the video frame).   

\begin{itemize}
    \item \textbf{Image:} We use the whole middlemost frame of the video, with a drawn-on bounding box as the image input. Additional image transformations such as random resize and cropping follow the procedures used to finetune Merlot Reserve on VCR~\cite{zellers2022merlotreserve}. 
    \item \textbf{Video:} While Merlot Reserve uses 7 video frames spaced 5 seconds apart as the video input for TVQA, we instead randomly select 4 video frames spaced 0.8 seconds apart (as we are constrained by the length of our videos). We tested a range of $[3, 7]$ video frames to determine which amount was the best, and decided to use 4 frames, as increasing the number did not improve results but resulted in a longer training time. Additional video augmentations such as cropping follow the procedures used to finetune TVQA~\cite{zellers2022merlotreserve}. 
    \item \textbf{Audio:} Similar to the video input, Merlot Reserve takes the surrounding 5 seconds of audio corresponding to each video frame used as the video input. Since our video frames are spaced so closely together, we take the surrounding 1.6 seconds of audio surrounding each video frame. This results in an overlap of audio in the input sequence, but is necessary, as 1.6 seconds is the minimum length allowed for audio segments. Additional audio augmentation such as random audio scaling follow procedures used to finetune TVQA~\cite{zellers2022merlotreserve}.  
    \item \textbf{Text:} We use a tokenized text input and do not use other text augmentations.
\end{itemize}

\subsection{Material Classification Dataset}

To construct the material classification dataset, we iterated through each existing object pair in our current dataset. For each pair of objects $(o^{(1)}, o^{(2)})$, we have a corresponding set of materials $\mathbf{M_1} = \{m_1^{(1)}, m_2^{(1)}, \dots \}$ and $\mathbf{M_2} = \{m_1^{(2)}, m_2^{(2)}, \dots \}$. Each list contained at least one material gathered from our Video Annotation step (see section~\ref{appendix:data2}). Then, for each material in $\mathbf{M_1} \setminus \mathbf{M_2}$, we would create a question by inserting the material into a randomly-chosen template string. Each question would then be given a label of $0$, and become a datapoint. Likewise, we also create a question for each material in $\mathbf{M_2} \setminus \mathbf{M_1}$, and assign a label of $1$. Note that some pairings would have no questions attached (if they were made of the same materials), and other pairings could have more than 2 questions. 

The specific list of template questions used is shown here:
\begin{enumerate}
    \item Which object is more likely to be made out of \_\_?
    \item Which object is more likely to be a \_\_-like object?
    \item Which item is probably made of \_\_?
    \item If you had to pick, which object do you think is made of \_\_?
    \item Which object looks like it is made out of \_\_?
    \item Out of the two objects, which is more likely to contain \_\_?
\end{enumerate}

In total, 1619 of the 1839 object pairs in the training set were used, 204 of the 225 pairs in the validation set were used, and 209 of the 228 pairs in the test set were used. 

\section{Additional Results}\label{appendix:results}

In this section, we provide extra quantitative results that could not fit into our main paper. These results focus on the importance of audio, by analyzing the effects that changes to the input audio have on model performance. Additionally, we provide some more qualitative results from CLIP and Merlot Reserve.

\subsection{Removing and Flipping Audio}

\begin{table}[t]
\setlength{\tabcolsep}{6pt}
\renewcommand{\arraystretch}{1.2}
\centering
\begin{tabular}{c|c|c|c}
\Xhline{3\arrayrulewidth}
\multirow{2}{*}{Subset}   & \multicolumn{3}{c}{\textbf{Accuracy ($\%$)}} \\ \cline{2-4}
     &  Normal  & Silent & Flipped \\ \hline
     Val  & $61.2 \pm 1.1$ & $58.9 \pm 0.7$ & $55.3 \pm 0.6$ \\
    Test & $60.7 \pm 0.3$ & $57.9 \pm 0.6$ & $55.0 \pm 1.5$ \\
\Xhline{3\arrayrulewidth}
\end{tabular}
\vspace{1mm}
\caption{Comparison of results using the AudioCLIP model when perturbing the input audio. ``Silent'' refers to results when \textit{one} of the input audios is changed to silence, and ``Flipped'' refers to results where the two input audios are swapped. Both perturbations will cause a noticeable amount of datapoints to become mislabeled and thus negatively impact the prediction accuracy.}
\label{table:new_results}
\end{table}

Previously, our analysis focused on the impact of removing audio from \textit{both} inputs, but here, we look at additional results of perturbing the input audio. As shown in Table~\ref{table:new_results}, there is roughly a 2\% decrease in performance using the AudioCLIP model if one of the two input audios is changed to silence, and roughly a 5-6\% decrease in performance when we \textit{flip} the input audios. This reinforces the idea that models rely on audio to extract physical properties, with changes to the input audio resulting in markedly worse results.

\subsection{Confusing Combinations of Materials}

\begin{figure}[t]
    \begin{subfigure}{0.48\textwidth}
        \centering
        \includegraphics[width=0.96\textwidth]{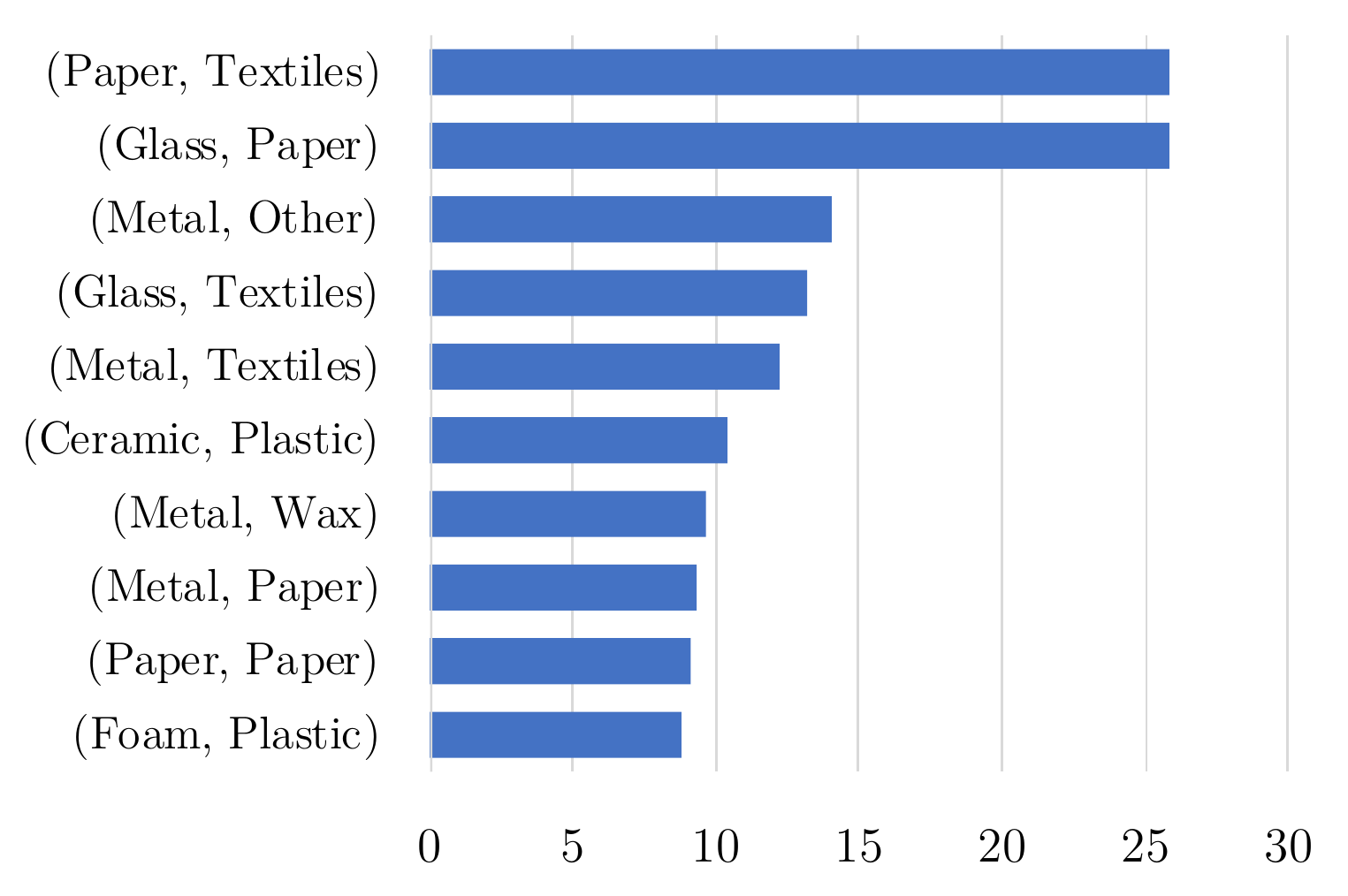}
        \caption{The difference in accuracy with and without audio on CLIP, conditioned on material pairs. }
        \label{fig:mat_pairs_clip}
    \end{subfigure}
    \hfill
    \begin{subfigure}{0.48\textwidth}
        \centering
        \includegraphics[width=0.96\textwidth]{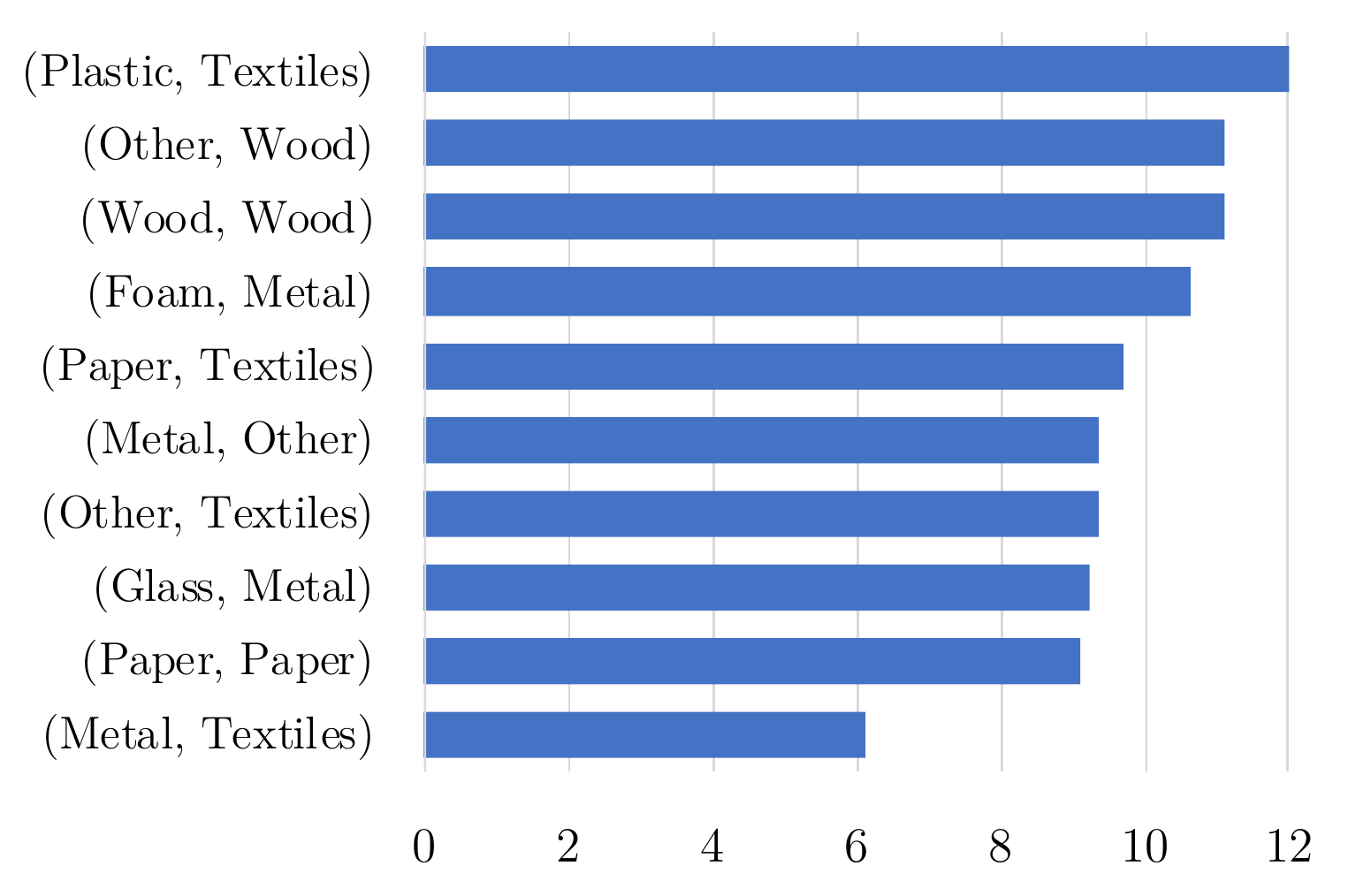}
        \caption{The difference in accuracy with and without audio on Merlot Reserve, conditioned on material pairs. }
        \label{fig:mat_pairs_merlot}
    \end{subfigure}
    \caption{Comparison of results on models with and without audio, when we condition on pairs of materials. For each pair of materials, we only consider datapoints where Object 1 is made from one of the two materials, and Object 2 is made from the other. We can see patterns such as pairs containing two of the same material (e.g., [Paper, Paper]), or possibly visually similar materials (e.g., [Ceramic, Plastic]).}
    \label{fig:mat_pairs_comp}
\end{figure}

Continuing with results when conditioning on certain variables, we can see results when conditioning on \textit{pairs} of materials in Figure~\ref{fig:mat_pairs_comp}. Specifically, when we condition on a given pair of materials (e.g., [glass, plastic]), we only look at results on datapoints where one of the objects is made of the first material (e.g., glass), and the other object is made of the second material (e.g., plastic). Here, we see that audio can be helpful when two objects are made of the same general material, and thus requires more fine-grained information to differentiate, or when materials may look somewhat similar, but sound very different (e.g., [Foam, Plastic]). 

\subsection{CLIP/AudioCLIP}
\begin{figure}[t]
    \centering
  \includegraphics[width=\textwidth]{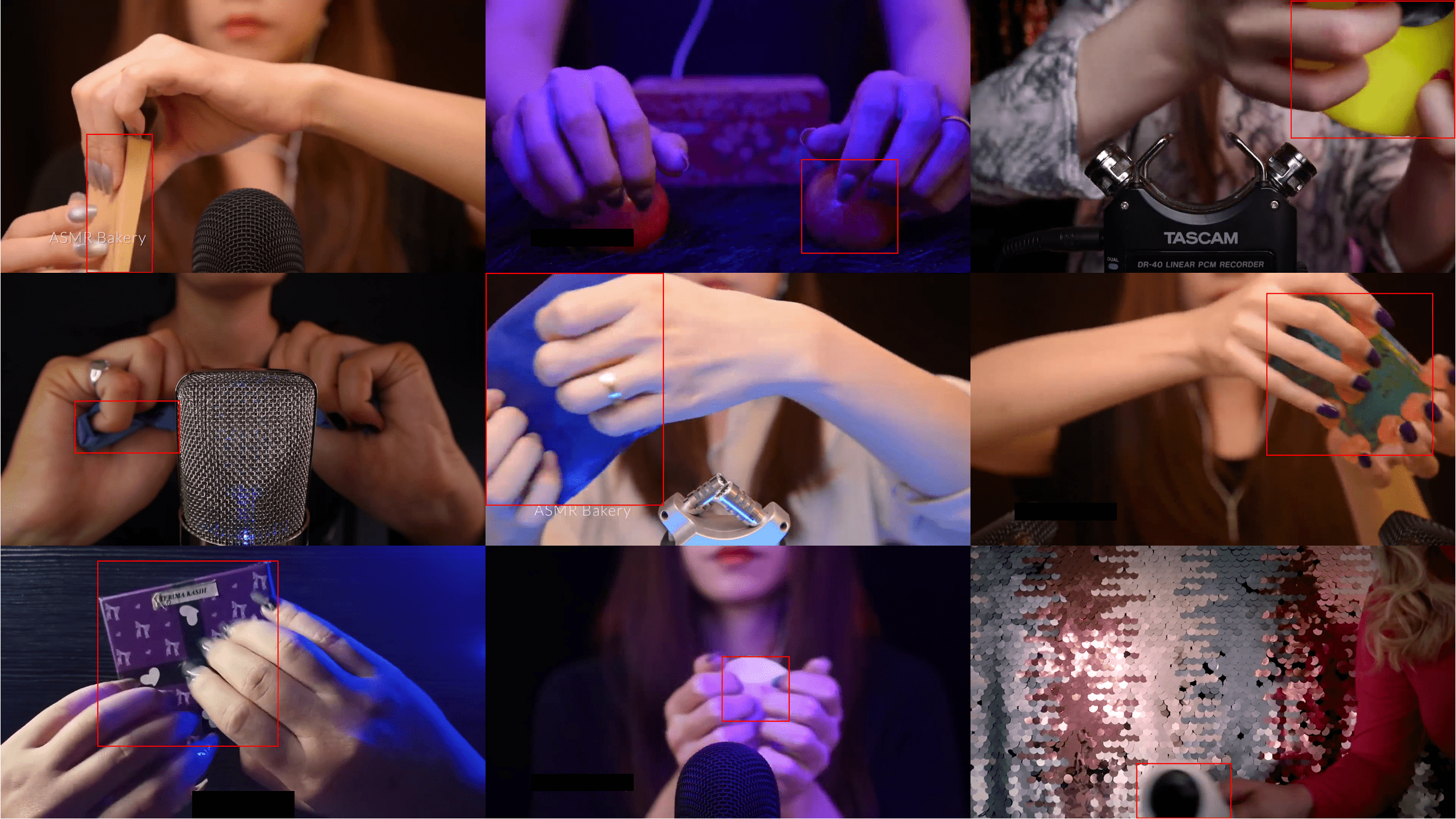}
  \caption{Example objects where CLIP without audio performed much worse than CLIP with audio. These objects are partially or mostly occluded, which makes audio especially important when given limited visual information. }
  \label{fig:clip_occluded}
\end{figure}

\begin{figure}[t]
    \centering
  \includegraphics[width=\textwidth]{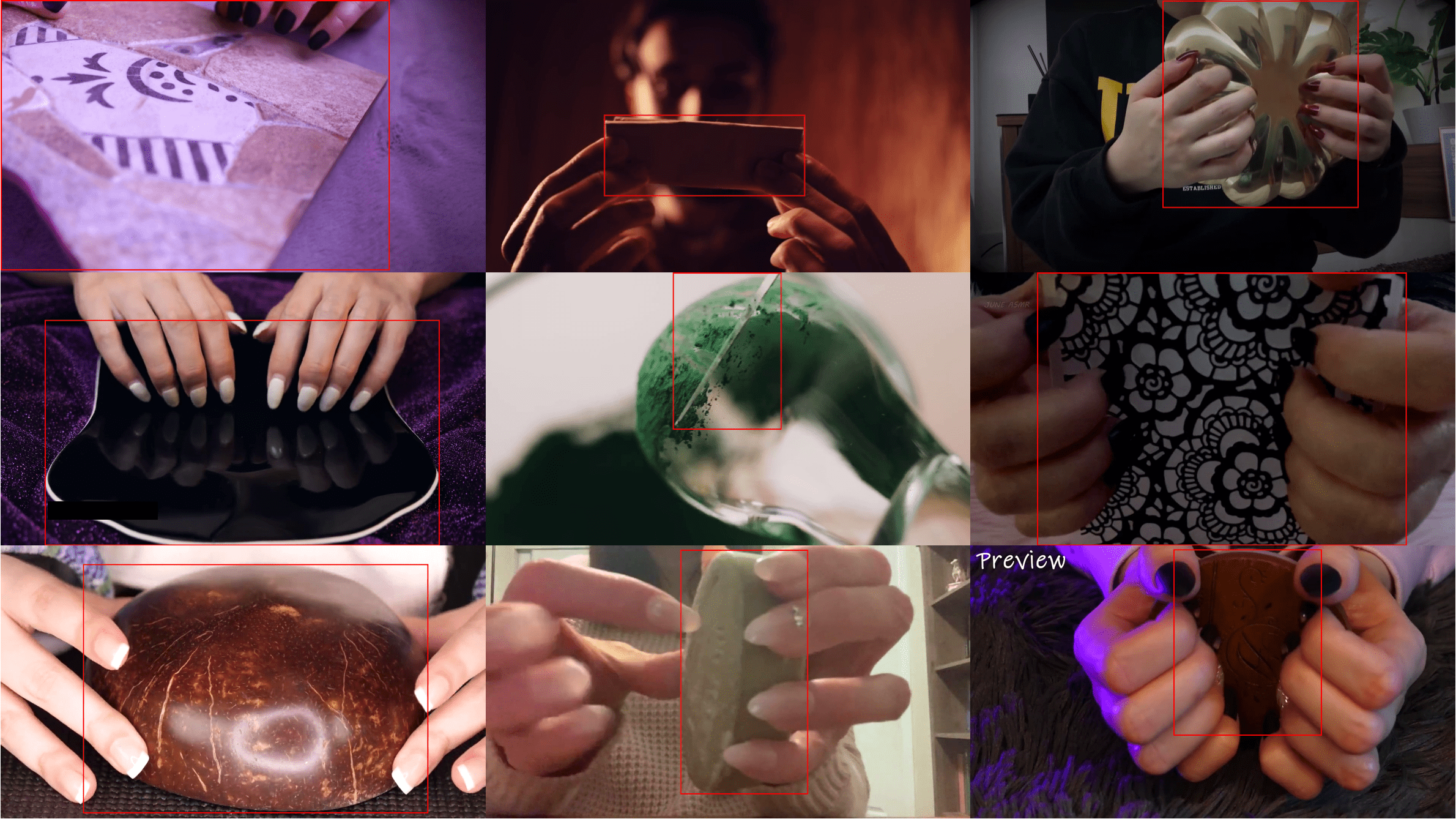}
  \caption{Example objects where CLIP without audio performed much worse than CLIP with audio. These show visually ambiguous objects, where it isn't obvious what the objects are made of, and thus audio is important in understanding various physical properties.}
  \label{fig:clip_ambiguous}
\end{figure}

Since each video in our dataset is paired three times, and each pair has multiple questions attached, we can calculate each individual model's accuracy on a specific object, by only looking at datapoints containing that object. When qualitatively analyzing the videos that AudioCLIP does well on, but CLIP struggles on, we see three major patterns:
\begin{enumerate}

    \item \textbf{Partially or fully occluded objects:} Since CLIP only accepts a single image input, if the object is occluded in the image, then it is almost impossible for CLIP to extract any information. In those cases, it is natural for audio to play a huge role in providing information about the object, and thus allows AudioCLIP to perform much better than CLIP (see Figure~\ref{fig:clip_occluded} for examples).
    \item \textbf{Visually ambiguous objects:} Due to our video filtering steps, our dataset contains a large amount of visually ambiguous objects, such that humans need the entire video or audio input to truly understand the objects' properties. Figure~\ref{fig:clip_ambiguous} shows examples of visually ambiguous objects where CLIP performed poorly, but AudioCLIP did not. Thus, audio may be helpful to provide the extra information needed to differentiate between various materials or pinpoint specific physical properties of objects (eg. whether the object is soft or hard).
    \item \textbf{Uncommon objects:} Even though CLIP was trained on a large set of images depicting a wide variety of objects and scenes~\cite{radford2021clip}, we find numerous examples of ``uncommon'' objects where audio is helpful. Due to the somewhat strange nature of the objects, CLIP may not have developed a solid understanding of these objects in the pre-training steps. Thus, in such cases, audio may be important for inferring the physical properties of such objects.
\end{enumerate}

\begin{figure}[t]
    \centering
  \includegraphics[width=\textwidth]{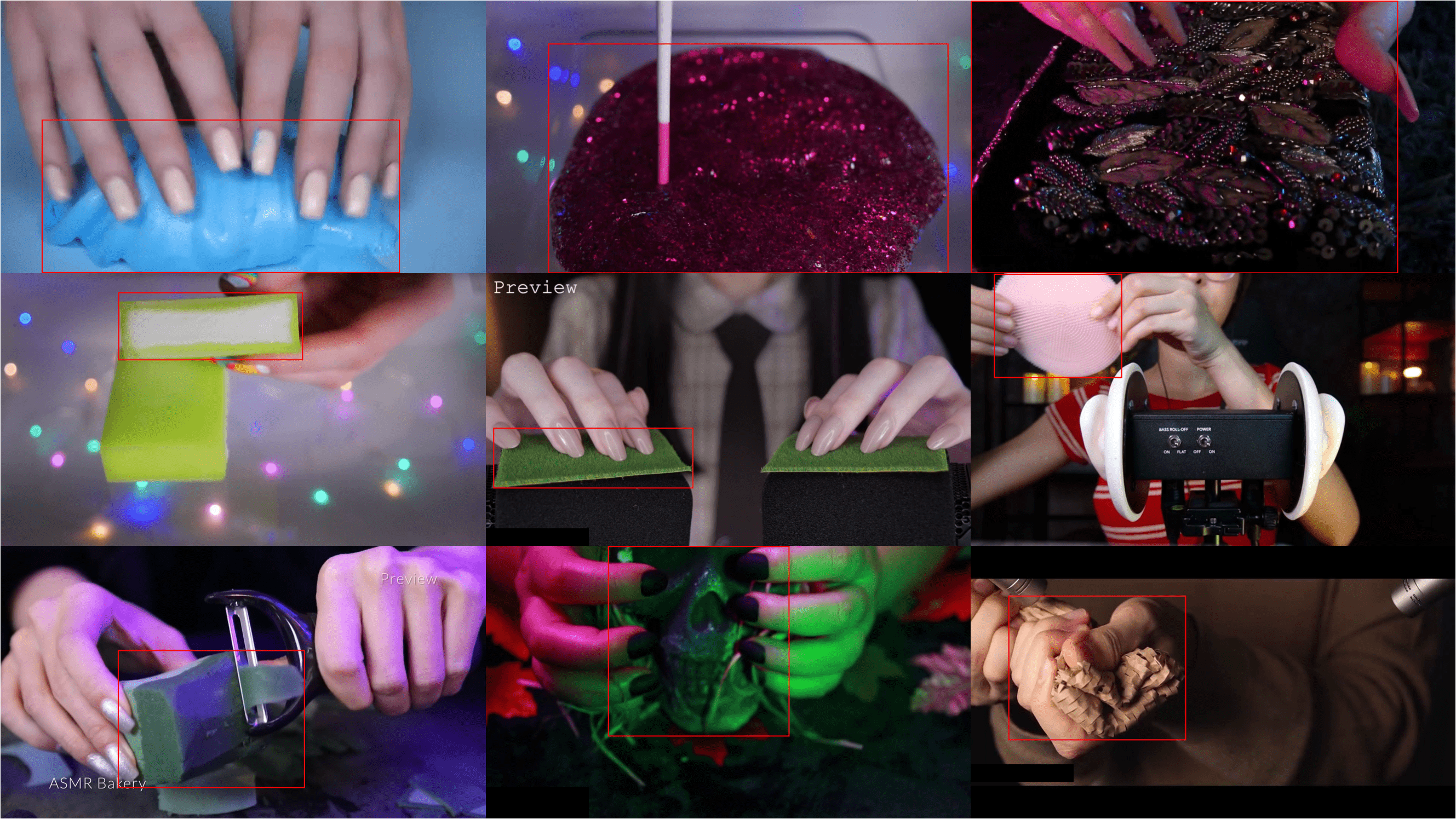}
  \caption{Example objects where CLIP without audio performed much worse than CLIP with audio. These examples show uncommon and likely out-of-domain objects, where audio can help the model infer various physical properties. }
  \label{fig:clip_uncommon}
\end{figure}

\clearpage

\subsection{Merlot Reserve}


\begin{figure}[t!]
    \centering
  \includegraphics[width=\textwidth]{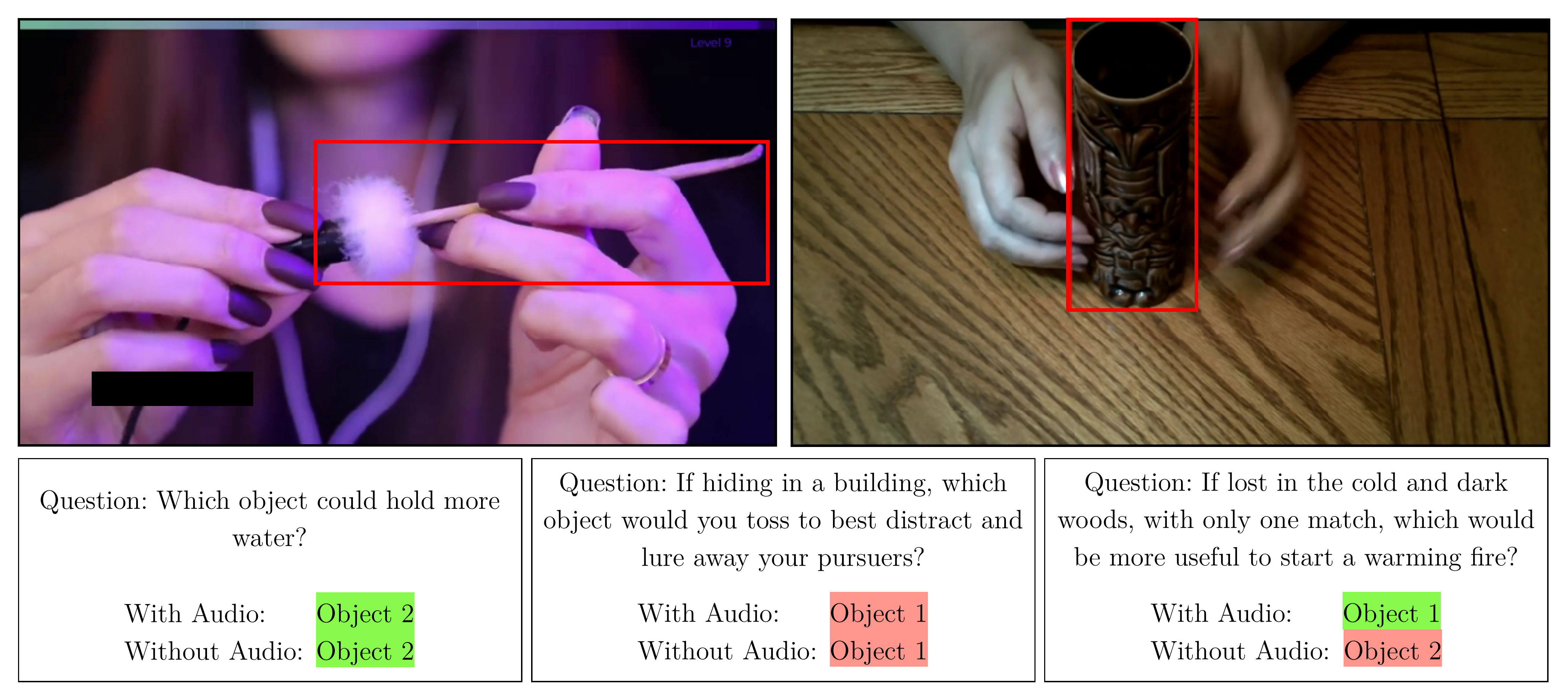}
  \vspace*{-7mm}
  \caption{In this example, the object on the right can be mistaken for painted wood or metal, rather than ceramic. As such, the model without audio mistakenly predicts that the object on the right is more useful to start a fire with. We also see that both models fail to answer the middlemost question, as it asks about physical properties such as shape, weight, and hardness in a less direct manner.}
  \label{fig:example2}
  \vspace{-2mm}
\end{figure}

\begin{figure}[t!]
    \centering
  \includegraphics[width=\textwidth]{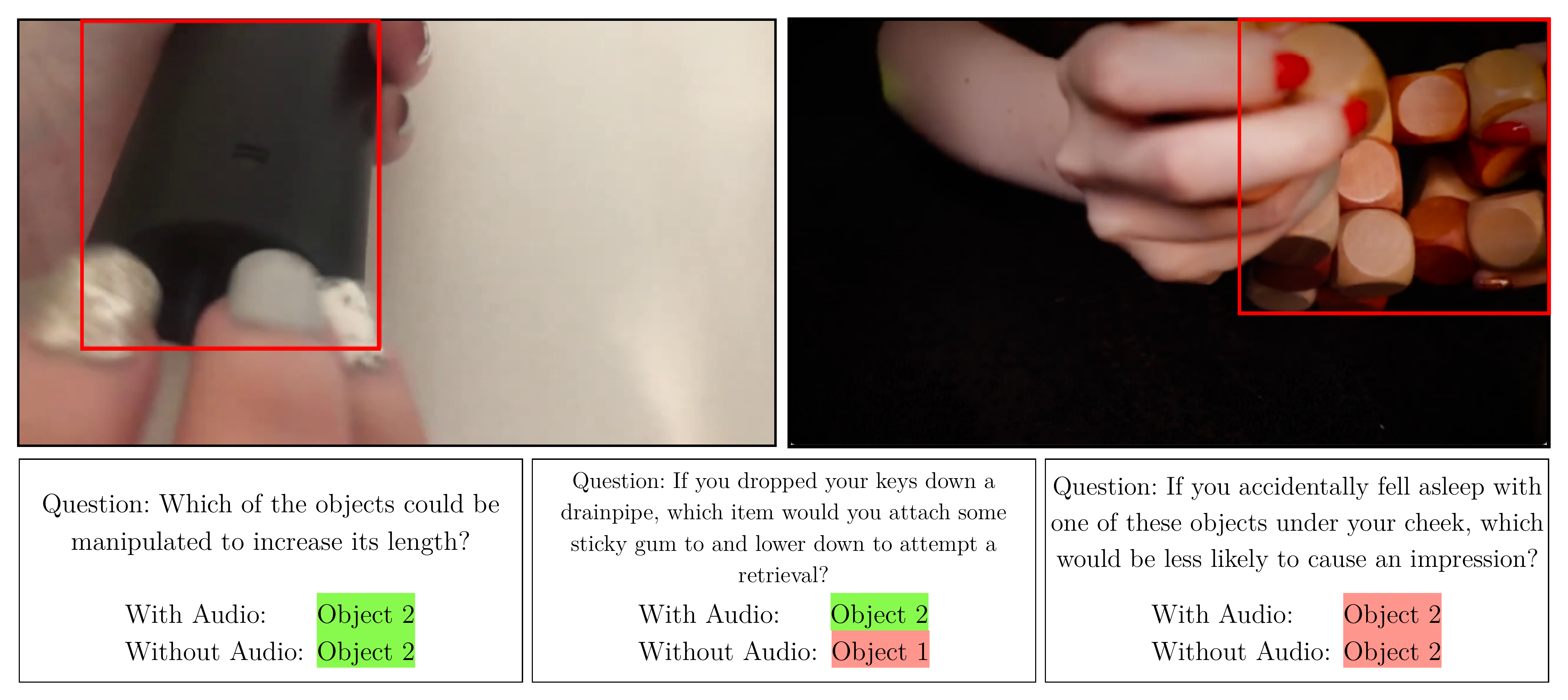}
  \vspace*{-7mm}
  \caption{ In this example, the left object is a plastic TV remote, and the right object is a wooden snake toy. As a common theme, both models struggle to answer longer questions with more ``human'' elements, such as the rightmost question in this example.}
  \label{fig:example7}
  \vspace{-5mm}
\end{figure}

\begin{figure}[t]
    \centering
  \includegraphics[width=\textwidth]{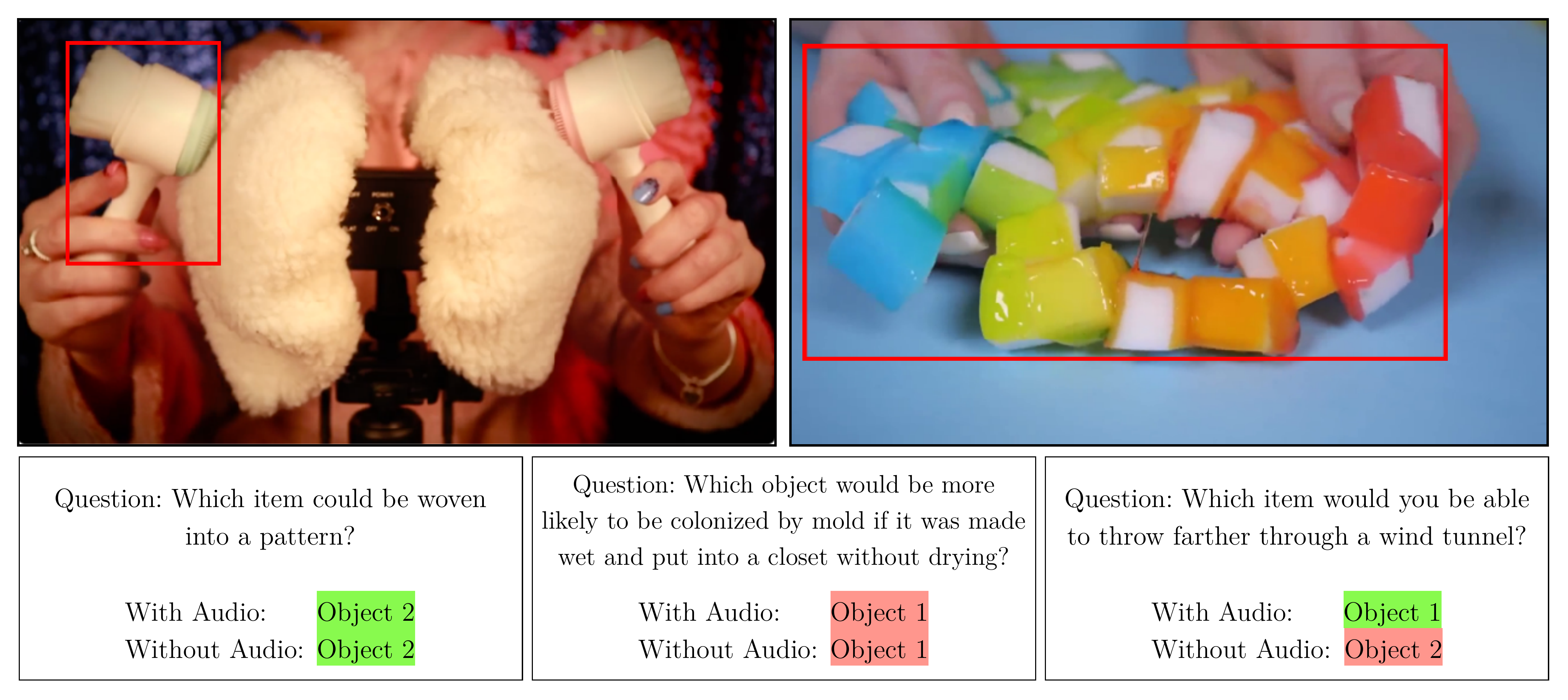}
  \vspace{-5mm}
  \caption{In this example, the object on the right is unusual, but through viewing the entire video, we can see that it is made of soft and sticky foam. Once again, both models fail to answer the more middlemost question, which is more complex than the others.}
  \label{fig:example3}
\end{figure}

\begin{figure}[t]
    \centering
  \includegraphics[width=\textwidth]{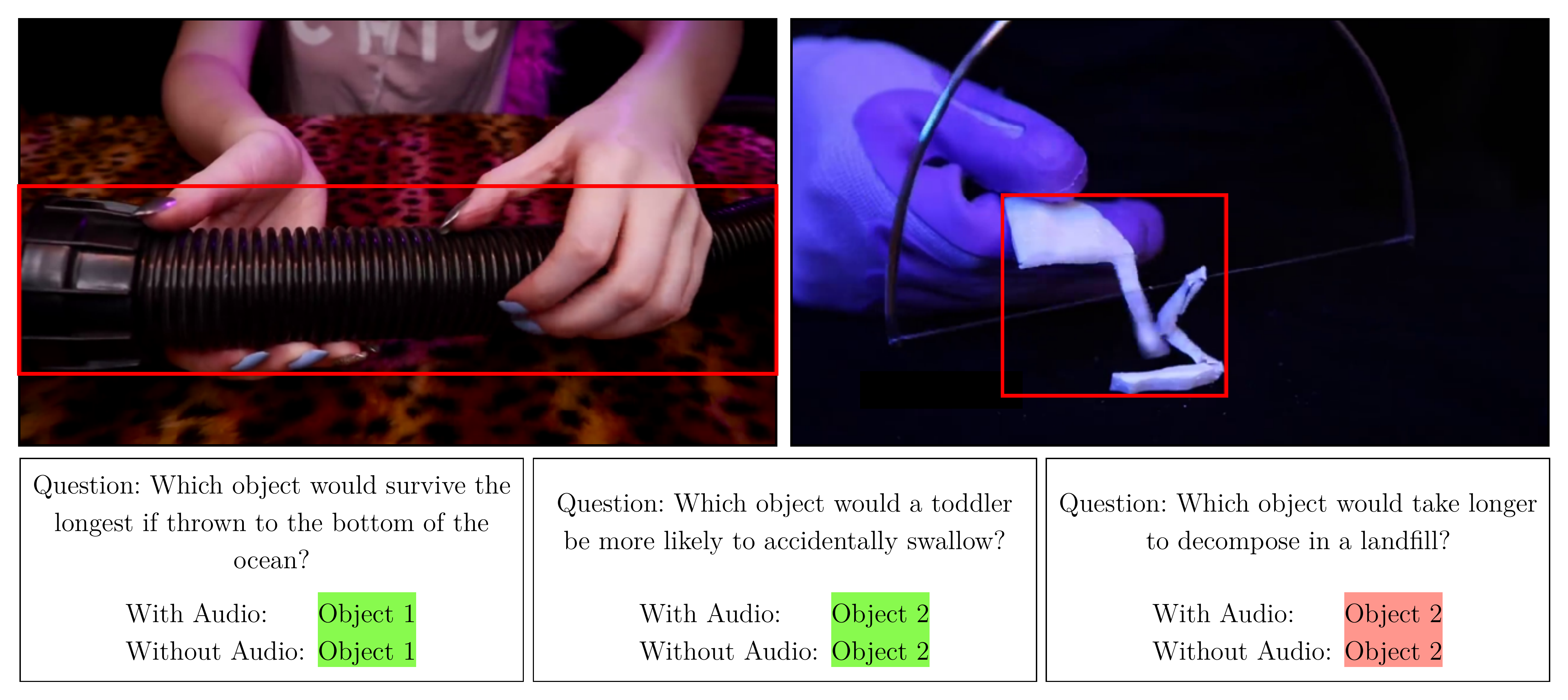}
  \vspace{-5mm}
  \caption{In this example, the left object is made of plastic, and the right object is a small piece of styrofoam, which may not be clear until provided audio. In this example, we see two questions that ask about very similar physical properties. While the two models correctly answer the leftmost question, when framed as ``which object would take longer to decompose'' in the rightmost question, the models answer incorrectly. This shows a possible difference between how humans and models reason about physical concepts. In a \textit{technical} sense, plastic and styrofoam both take roughly equally long to decompose. However, human annotators can confidently choose the smaller piece of styrofoam as what they believe to be the logical answer, but current models may lack the flexibility to answer in such a manner.}
  \label{fig:example4}
\end{figure}

\begin{figure}[t]
    \centering
  \includegraphics[width=\textwidth]{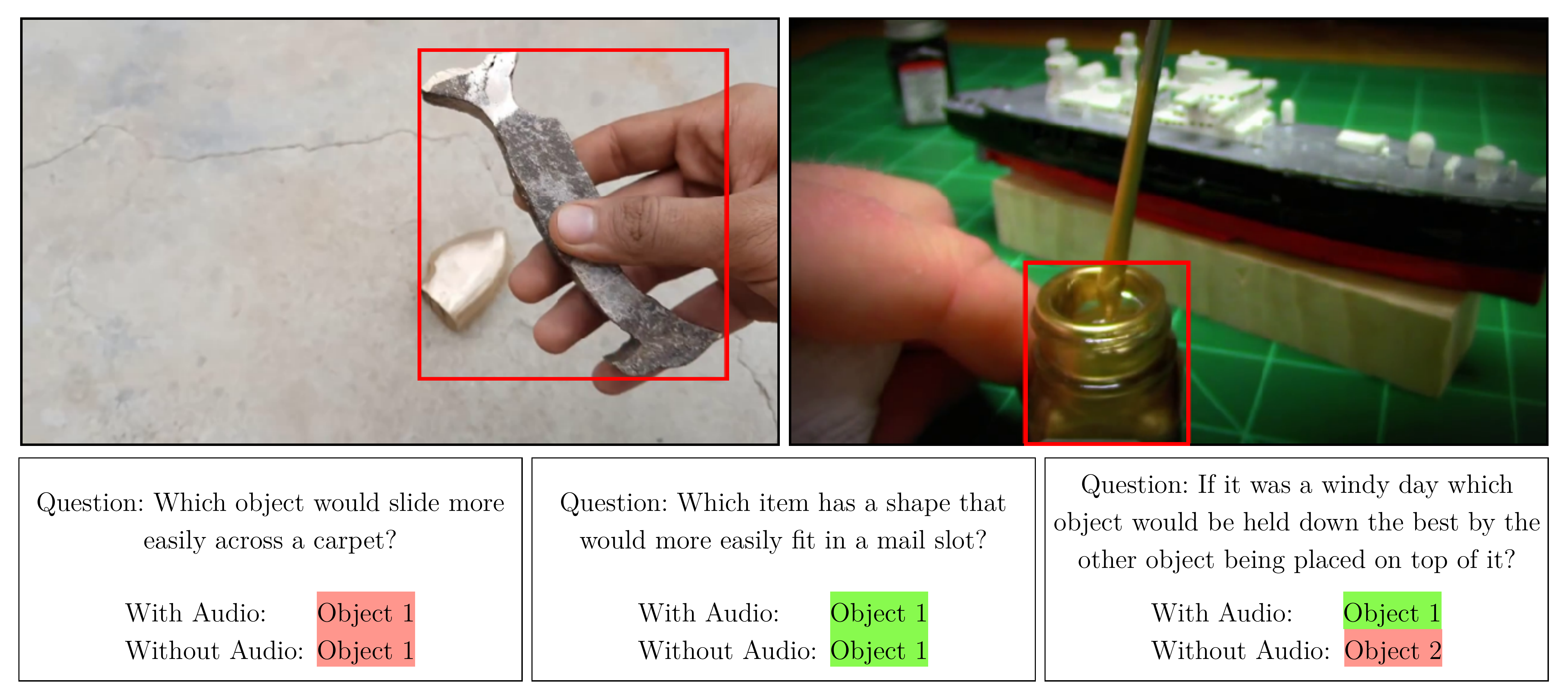}
  \vspace{-3mm}
  \caption{In this example, the left object is a piece of metal, and the right object is a small glass bottle. Surprisingly, both models fail to answer the leftmost question, which is a rather simple question asking about the shape and smoothness of the objects. One possibility for this failure is that the models rigidly associate flat objects with sliding and round objects with rolling, rather than developing a flexible understanding of how physical properties and words relate to each other. }
  \label{fig:example5}
\end{figure}

\begin{figure}[t]
    \centering
  \includegraphics[width=\textwidth]{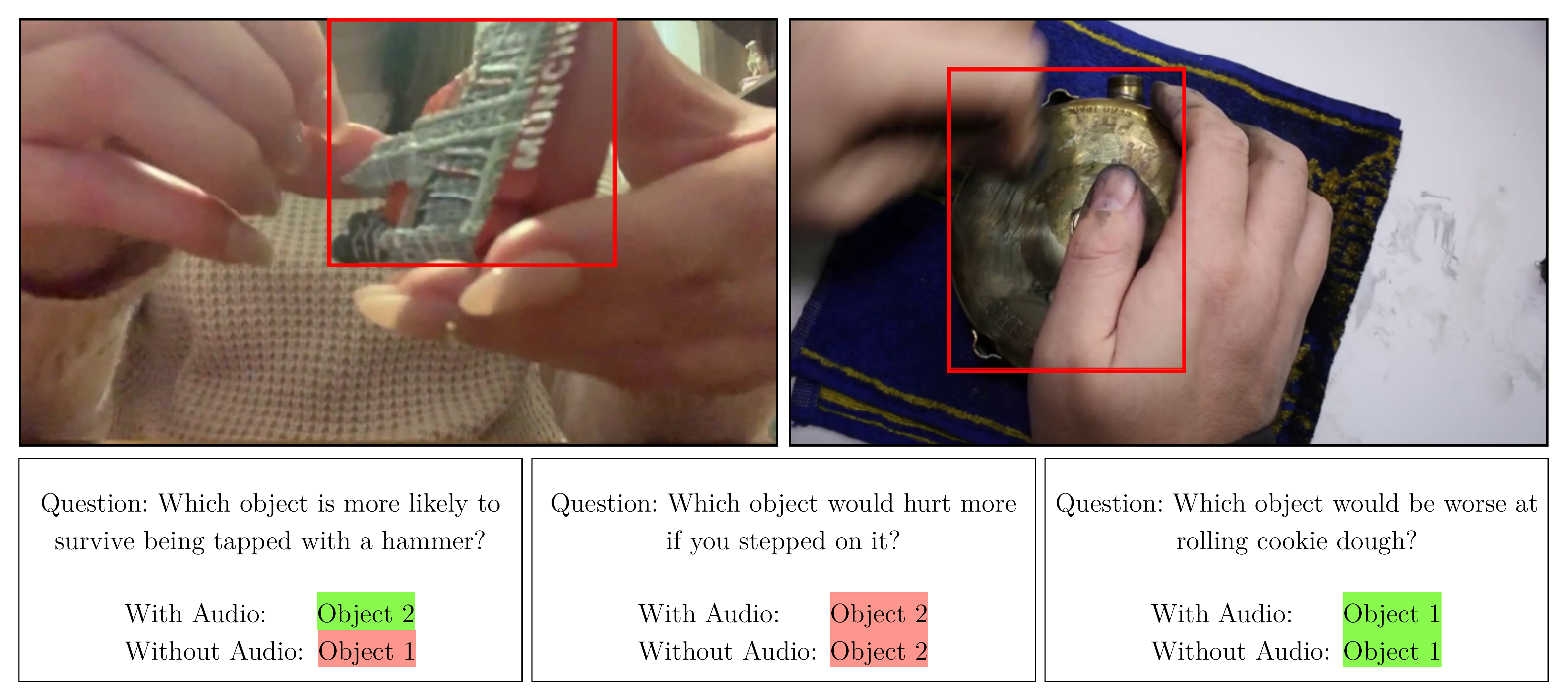}
  \vspace{-3mm}
  \caption{In this example, the left object is a ceramic figure, and the right object is made of metal. Such information may not be clear without audio, which could cause the model without audio to misunderstand the material, and thus the strength, of the left object. Both models fail to answer the middlemost question, possibly due to the models not entirely understanding the concept of pain.}
  \label{fig:example6}
\end{figure}

\begin{figure}[t]
    \centering
  \includegraphics[width=\textwidth]{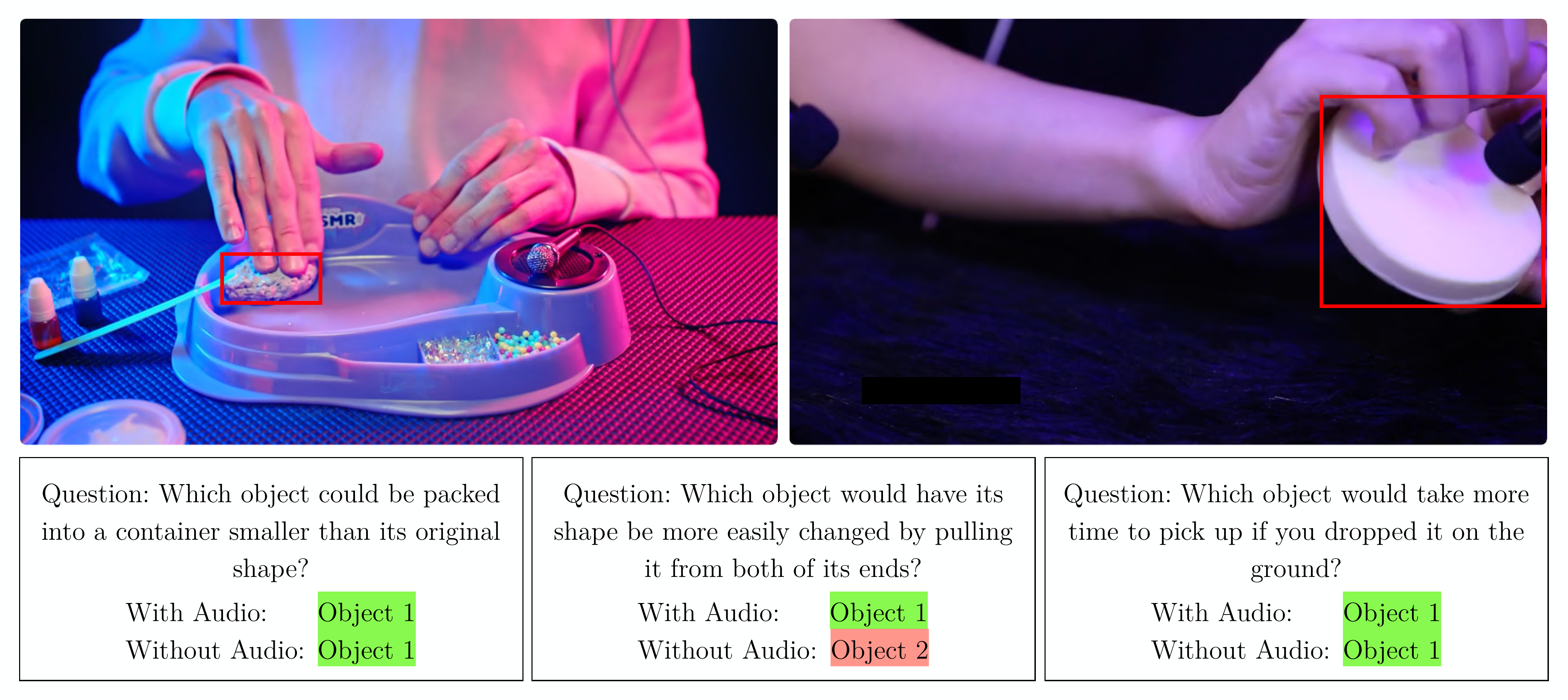}
  \vspace{-3mm}
  \caption{In this example, the first object is uncommon and the second object is visually ambiguous. Thus, without audio, the model struggles to understand how flexible or stretchy the second object is.}
  \label{fig:example8}
\end{figure}

\vspace{-2mm}
In this subsection, we provide additional sample outputs using the Merlot Reserve models trained with and without audio (see Figures~\ref{fig:example2}-~\ref{fig:example6}). From analyzing the qualitative examples, we see a couple of clear patterns. Firstly, the benefit of audio in understanding uncommon or visually ambiguous objects is reaffirmed, as shown in the rightmost question in Figure~\ref{fig:example2}, or the leftmost question in Figure~\ref{fig:example6}. Examples where both models fail to answer correctly commonly involve longer and more complex questions (e.g., the middlemost question in Figure~\ref{fig:example2} or the rightmost question in Figure~\ref{fig:example7}), or contain biases from human annotation (e.g., the middlemost question in Figure~\ref{fig:example6} or the rightmost question in Figure~\ref{fig:example4}). 

\clearpage

\section{\names\ Datasheet}\label{appendix:datasheet}

\subsection{Motivation} 

\begin{itemize}
    \item \textbf{For what purpose was the dataset created?} The \names\ dataset was created to bridge a gap in physical commonsense reasoning by introducing the acoustic and video modalities. 

\item \textbf{Who created the dataset (e.g., which team, research group) and on behalf of which entity (e.g., company, institution, organization)?} The dataset was created by Samuel Yu, Peter Wu, and Paul Pu Liang, as part of work done in the CMU Multicomp Lab. 

\item \textbf{Who funded the creation of the dataset?} This work is partially supported by the National Science Foundation (Awards \#1722822 and \#1750439) and National Institutes of Health (Awards \#R01MH125740, \#R01MH096951, and \#U01MH116925). 
\end{itemize}

\subsection{Composition}

\begin{itemize}
\item \textbf{What do the instances that comprise the dataset represent?} Each instance contains a physical commonsense question and two objects. Each object is represented by a short video, the corresponding audio, and a bounding box drawn around the middlemost image frame of the video. Finally, there is a binary label denoting which object is the correct answer to the question. 

\item \textbf{Are relationships between individual instances made explicit?} Yes, we provide an ID for each video and the question in every instance, and thus other instances using the same video or question can be determined. 

\item \textbf{How many instances are there?} There are \numprint{13400} instances in total, with \numprint{11044} training instances, \numprint{1192} validation instances, and \numprint{1164} testing instances. 

\item \textbf{Does the dataset contain all possible instances?} No, physical commonsense is a very broad topic, and we focus on a subset of objects, input modalities, and phsyical phenomena. 

\item \textbf{What data does each instance consist of?} The video clip and audio come from a longer YouTube video, and thus represented by a YouTube Video ID and frame range. The bounding box is represented by four image coordinates (Left, Top, Right, Bottom), and the specific frame it corresponds to in the YouTube video. Finally, the questions are represented by raw text, and there is a label denoting the correct answer. 

\item \textbf{Are there recommended data splits?} Yes, we provide training, validation, and testing sets. Users will be able to submit their predictions for the testing set online. The sets were randomly split, with measures to ensure that there is no overlap in the video clips used between the splits. 

\item \textbf{Are there any errors, sources of noise, or redundancies in the dataset?} Currently, none to our knowledge. 

\item \textbf{Is the dataset self-contained, or does it link to or otherwise rely on
external resources (e.g., websites, tweets, other datasets)?} We release YouTube video IDs and a script to download and trim the videos. It is possible for users to delete YouTube videos, but otherwise, our data is permanent. In cases where the video is deleted, we at minimum provide pre-extracted image features from the middlemost frame of the video to use. To our knowledge, users will not face any licensing issues when using clips from YouTube videos to train machine learning models~\cite{zellers2021merlot,zellers2022merlotreserve}. Beside the video data, our dataset is self-contained. 

\item \textbf{Does the dataset contain data that might be considered confidential?} No, all of the videos that we use are of publicly available data, following the Terms of Service that users agreed to when uploading to YouTube.

\item \textbf{Does the dataset contain data that, if viewed directly, might be offensive, insulting, threatening, or might otherwise cause anxiety?} To our knowledge, such videos do not exist in our dataset, as annotators were instructed to filter out inappropriate videos during our annotation process. 

\item \textbf{Does the dataset identify any subpopulations?} No

\item \textbf{Is it possible to identify individuals, either directly or indirectly from the dataset?} No explicit personal information is included in our dataset. Since our videos common from oftentimes popular channels, and some videos contain parts of human faces, it may be possible, though unlikely, to indirectly identify individuals. 

\item \textbf{Does the dataset contain data that might be considered sensitive in
any way?} To our knowledge, sensitive data was filtered out. 

\end{itemize}

\subsection{Collection Process} 
\begin{itemize}
    \item \textbf{How was the data associated with each instance acquired?} The videos were gathered from YouTube. The bounding boxes for each object, questions, and labels for each datapoint were gathered using annotators. For more details, see section~\ref{appendix:data}.

\item \textbf{What mechanisms or procedures were used to collect the data?} We used the YouTube API and the PyTube library to gather the videos. Then, we used Amazon Mechanical Turk for additional annotation steps. 

\item \textbf{If the dataset is a sample from a larger set, what was the sampling
strategy?} We gathered the top results from YouTube's search algorithm. The dataset splits were created randomly, and object pairs within each split were also created randomly while guaranteeing each object was paired 3 times. 

\item \textbf{Who was involved in the data collection process and how were they compensated?} We used two in-house annotators, and additional Amazon MTurk workers. Details about wages were presented in section~\ref{appendix:data}. 

\item \textbf{Over what timeframe was the data collected?} The YouTube videos were collected and filtered from July-September 2021, the questions were created from September-October 2021, and all other annotations were collected from October-November 2021. 

\item \textbf{Did you collect the data from the individuals in question directly,
or obtain it via third parties or other sources?} The video data was collected from YouTube, and annotation data were collected directly from the annotators via Amazon Mechanical Turk.

\item \textbf{Were the individuals in question notified about the data collection?} The Amazon MTurk workers were informed that ``all data submitted in this set of HITs will be used in a dataset for artificial intelligence research''. However, due to the difficulty of informing all involved YouTube channels, we did not explicitly inform them. We presume that the channel owners were aware that their videos would be public. 

\item \textbf{Did the individuals in question consent to the collection and use
of their data?} MTurk Workers consented to have their responses used in our dataset through the Amazon Mechanical Turk Participation Agreement, and by agreeing to submit HITs after being informed of the purpose of their work. 

\item \textbf{If consent was obtained, were the consenting individuals provided
with a mechanism to revoke their consent in the future or for certain
uses?}  By only providing YouTube video IDs, users are able to revoke consent by deleting or setting their video to private. There is no mechanism for MTurk workers at this time.

\end{itemize}

\subsection{Preprocessing/cleaning/labeling}

\begin{itemize}
    \item \textbf{Was any preprocessing/cleaning/labeling of the data done?} We first split long YouTube videos into segments roughly 5-10 seconds long, and used several filtering steps to isolate a high-quality set of videos for our dataset (see section~\ref{appendix:data} for details). 
    
    \item \textbf{Was the “raw” data saved in addition to the preprocessed/cleaned/labeled data?} Yes, we have saved all the datapoints and videos that were filtered out in our annotation process. 
    
    \item \textbf{Is the software that was used to preprocess/clean/label the data available?} Yes, details of the software used and example annotation interfaces are provided in section~\ref{appendix:data}. 
\end{itemize}

\subsection{Uses}

\begin{itemize}
    \item \textbf{Has the dataset been used for any tasks already?} Currently, only our submission has used this dataset. 
    
    \item \textbf{Is there a repository that links to any or all papers or systems that
use the dataset?} The dataset download and evaluation code can be found at \href{https://github.com/samuelyu2002/PACS}{\email{https://github.com/samuelyu2002/PACS}}.

\item \textbf{What (other) tasks could the dataset be used for?} As demonstrated in our paper, our dataset can also be used for material classification. Another use-case could involve identifying an ``object of focus'' given an image frame. 

\item \textbf{Are there tasks for which the dataset should not be used?} Our dataset is in no ways exhaustive, and is designed only as a benchmark to be used strictly for research purposes. \names\ should not be used to train any systems that will be deployed in the real world.
\end{itemize}

\subsection{Distribution}

\begin{itemize}
\item \textbf{Will the dataset be distributed to third parties outside of the entity (e.g., company, institution, organization) on behalf of which the
dataset was created?} Yes, the dataset will be publicly available. 

    \item \textbf{How will the dataset will be distributed?} We distribute all the video, midframes, and other necessary data. We will also release JSON files containing all the datapoints and other relevant metadata (eg. relevant physical properties for each question). Finally, we will also release pre-processed versions of our dataset to train UNITER (pre-extracted image features stored in NumPy databases), and Merlot-Reserve (compressed tfrecord files). Currently, the data is hosted in Google Drive, and accessible through the project's GitHub repository.

\item \textbf{When will the dataset be distributed?}  The dataset is available at \href{https://github.com/samuelyu2002/PACS}{\email{https://github.com/samuelyu2002/PACS}}.

\item \textbf{Will the dataset be distributed under a copyright or other intellectual property (IP) license?} The copyright of the video clips used in our dataset belongs to YouTube and the channels that published the videos. However, we believe that with the additional annotation and processing of short clips taken from YouTube videos, our use of such videos constitutes ``Fair Use''. 
For all other data, we will use a permissible license for research-based use. 

\end{itemize}

\subsection{Maintenence} 

\begin{itemize}
    \item \textbf{Who will be supporting/hosting/maintaining the dataset?} The first author of this work. 
    \item \textbf{How can the owner/curator/manager of the dataset be contacted?} Samuel Yu can be contacted at \email{samuelyu@andrew.cmu.edu}.
    \item \textbf{Will the dataset be updated?} We will update the dataset if flaws are found, but otherwise other updates (e.g., expanding the dataset) are not planned.
    \item \textbf{If the dataset relates to people, are there applicable limits on the retention of the data associated with the instances?} No.
    \item \textbf{Will older versions of the dataset continue to be supported/hosted/maintained?} If the dataset is updated, older versions will be kept for consistency. 
    \item \textbf{If others want to extend/augment/build on/contribute to the
dataset, is there a mechanism for them to do so?} Other researchers are free to download and build upon our dataset and contact the original authors about incorporating fixes/extensions. 
\end{itemize}

\clearpage

\end{document}